\documentclass[11pt]{article}
\usepackage{fullpage}

\usepackage[utf8]{inputenc} 
\pdfoutput=1

\usepackage{titling}
\usepackage{natbib}
\usepackage{graphicx}
\usepackage{amsmath}
\usepackage{amssymb}
\usepackage{bm}
\usepackage{bbm}
\usepackage{algorithmic}
\usepackage{booktabs}
\usepackage{amsmath}
\usepackage{algorithm}
\usepackage{color}
\usepackage{verbatim}
\usepackage{makecell}
\usepackage{enumerate}
\usepackage{graphicx}
\usepackage{threeparttable}
\usepackage{multirow}
\usepackage{mathrsfs}
\usepackage{caption}
\usepackage{subfigure, epsfig}
\usepackage{enumerate}
\def\bs{\boldsymbol}
\usepackage{hyperref}
\usepackage[figuresright]{rotating}
\usepackage{tabularx}
\usepackage{longtable}
\def\bs{\boldsymbol}

\newtheorem{theorem}{Theorem}
\newtheorem{remark}{Remark}

\DeclareMathOperator*{\argmax}{argmax}

\hypersetup{
	colorlinks=true,
	filecolor=blue,
	citecolor = blue,
	urlcolor=cyan,
}

\bibpunct[, ]{(}{)}{,}{a}{}{,}%
\def\BIBand{and}%

\title{Toward a Fairness-Aware Scoring System for Algorithmic Decision-Making}
\thanksmarkseries{alph}
\date{November, 2022}

\author{
	Yi Yang\thanks{Department of Intelligent Operations and Marketing, International Business School Suzhou,	Xi’an Jiaotong-Liverpool University} \\
	\and
	Ying Wu\thanks{Department of Information Systems and Intelligent Business,	School of Management, Xi'an Jiaotong University} \\
	\and
	Mei Li\thanks{Department of Marketing and Supply Chain Management, Price College of Business, University of Oklahoma} \\
\and
	Xiangyu Chang* \thanks{Department of Information Systems and Intelligent Business, School of Management, Xi’an Jiaotong University; E-mail: \texttt{xiangyuchang@xjtu.edu.cn}. } \\
	\and
	Yong Tan\thanks{Department of Information Systems and Operations Management, Michael G. Foster School of Business, University of Washington} \\
}

\begin{document}

	\maketitle

\begin{abstract}
		 Scoring systems, as a type of predictive model, have significant advantages in interpretability and transparency and facilitate quick decision-making. As such, scoring systems have been extensively used in a wide variety of industries such as healthcare and criminal justice. However, the fairness issues in these models have long been criticized, and the use of big data and machine learning algorithms in the construction of scoring systems heightens this concern. In this paper, we propose a general framework to create fairness-aware, data-driven scoring systems. First, we develop a social welfare function that incorporates both efficiency and group fairness. Then, we transform the social welfare maximization problem into the risk minimization task in machine learning, and derive a fairness-aware scoring system with the help of mixed integer programming. Lastly, several theoretical bounds are derived for providing parameter selection suggestions. Our proposed framework provides a suitable solution to address group fairness concerns in the development of scoring systems. It enables policymakers to set and customize their desired fairness requirements as well as other application-specific constraints. We test the proposed algorithm with several empirical data sets. Experimental evidence supports the effectiveness of the proposed scoring system in achieving the optimal welfare of stakeholders and in balancing the needs for interpretability, fairness, and efficiency.
\end{abstract}
	

\newpage
	
	\section{Introduction}
	
	Predictive models play an essential role in everyday decision-making \citep{bjarnadottir2018predicting}. A scoring system is a type of sparse linear predictive model whose coefficients are small integers \citep{UstunRu2016SLIM}. The coefficients derived from a scoring system can be directly transferred into point scores, which enable quick and easy calculation, interpretation, and decision-making \citep{souder1972scoring}. As such, scoring systems have been extensively adopted in a large variety of fields such as medical diagnosis \citep{Moreno2005saps,Wang2021instrumental}, criminal justice \citep{Hoffman1994twenty,Brayne2014surveillance}, consumer risk analysis \citep{capon1982credit,Karlan2011Microcredit}, marketing \citep{mcnamara1972present,zhang2015predicting}, and humanitarian operations \citep{ndirangu2013hiv,Skoufias2020estimating}. For example, in the medical field, scoring systems such as SAPS I, II, and III \citep{LeGall1984simplified,LeGall1993,Moreno2005saps} are used to predict ICU mortality risk. 
	Table \ref{paper_review} of the online supplementary material provides examples of the wide applications of scoring systems. 
	
	Notwithstanding their extensive applications, a crucial concern has been raised regarding fairness in assessment scores derived from these scoring systems, especially along some important social identities (hereafter referred to as sensitive features or sensitive attributes) such as membership in a certain race, gender, or socio-economic group \citep{coffman2021role}. For example, in the healthcare industry, racial bias has been found in a widely adopted patient health risk scoring system, where among patients receiving the same calculated risk scores, black patients are, in fact, found to be considerably sicker than white patients \citep{Obermeyer2019dissecting}. 
	Similarly, research shows that scores derived from COMPAS, a judicial system adopted for making critical pretrial, parole, probation, and sentencing decision in the U.S., are biased against African-American defendants in that it is skewed toward labeling black defendants as having high risk of reoffending whereas white defendants are labeled as having low risk \citep{Chouldechova2017fair}. There are ample other examples of biases found in scoring systems in the contexts of healthcare \citep{Bierman2007,Chen2008gender} and criminal justice \citep{campbell2020validation}, as well as in other contexts such as credit limits \citep{Vigdor2019apple} and loan applications \citep{Alesina2013women}. 
	
	
	As assessment scores are extensively used to make decisions and/or predictions, the lack of group fairness in scoring systems negatively impacts the lives of disadvantaged groups. In the case of COMPAS, assessment scores are used, among other things, to make probation and even sentencing decisions \citep{Berk2021fairness}. The lack of fairness in this type of scoring systems can have life-changing consequences for the disadvantaged population. In the healthcare industry, assessment scores are widely adopted and drive important healthcare resource-allocation decisions for millions of people in the U.S. \citep{Obermeyer2019dissecting}. The lack of fairness leads to disparity of access to medical resources for many lives in the disadvantaged groups.
	Thus, there is a dire need to incorporate group fairness into the development of scoring systems.
	
	In this paper, we propose a fairness-aware framework that aims to incorporate group fairness into the construction of scoring systems. 
	First, we quantify group fairness, integrating machine learning perspectives. Specifically, we investigate \textit{disparate impact} \citep{lambrecht2019algorithmic} and \textit{disparate mistreatment} \citep{Zafar2017fairness}, two types of outcome fairness recognized by machine learning literature. Then, we develop a social welfare function that incorporates both prediction efficiency and group fairness. We transform the social welfare maximization problem into the empirical risk minimization task in machine learning. With the help of mixed integer programming techniques, a fairness-aware scoring system is developed. Furthermore, we derive theoretical bounds to provide insights regarding the model parameter selection in the proposed fairness-aware scoring system. We validate the proposed algorithm with several empirical data sets. Findings support the effectiveness of the proposed scoring system in achieving the optimal welfare of stakeholders and in balancing the needs of interpretability, fairness, and efficiency.
	
	Our research makes the following contributions. First, we propose a workable solution to social group fairness issues by developing a new framework for the construction of scoring systems that is fairness aware. Specifically, we incorporate learning from machine learning and develop an objective function that appends fairness to prediction efficiency in an optimization problem for scoring system development. {This contrasts significantly with past research that is operation-centric, i.e., seeking to maximize or minimize some operational outcomes such as profit, costs, or accuracy \citep{celis2019classification}}, and treats fairness only as a constraint to an optimization problem. As such, our framework informs decision-makers on the overall welfare of those affected, incorporating the level of group fairness. This lays a foundation for the development of scoring systems that maximize total social welfare.
	
	Second, our framework provides a proven way of regulating the trade-off between efficiency and group fairness, as it can accommodate the different combinations of fairness and efficiency levels. This flexibility allows decision-makers the ability to set and customize their desired fairness requirement for different group fairness notions. In addition, our approach also provides policymakers with a way to customize other requirements by directly imposing a variety of application-specific constraints. Thus, our method provides decision-makers with a general way to construct scoring systems that suit an organization's context and objectives. 
	
	Third, when constructing the fairness-aware framework, 
	{{we utilize 0-1 hard loss to encode the efficiency objective (data utility), and the fairness objective as well as the fairness constraints are also formulated directly by several indicator variables derived via the 0-1 hard loss. Afterwards, we produce a scoring system without the rounding procedure. This contrasts with past research that applies surrogate loss both for objectives and constraints to make optimization problems convex,}} which leads to sub-optimization due to approximations. As a subsequent step, we derive the theoretical justification that provides guidance for model parameter selection. These parameters are instrumental to the implementation of the scoring systems.
	
	Fourth, we apply our algorithm to several empirical data sets. We find evidence of group biases in all of the empirical data sets we investigated, which supports the salience of group fairness issues in scoring systems and further motivates our research. More importantly, we validate the effectiveness of our new framework in enhancing group fairness in scoring systems. Lastly, our validation with sepsis data set sheds light on some promising clinical insights. The scoring system derived based on our framework has captured several risk factors with meaningful cut-off values for sepsis mortality prediction. Some of these factors have not been explored before. As such, a by-product of our research may help reveal new clinical insights in medical diagnosis.
	
	\section{Related Literature}
	{In this section, we provide an expanded view of fairness issues addressed in works of literature. We first describe scoring systems and their construction, paying particular attention to the use of data-driven methods to construct scoring systems. {We then introduce the three common notions of group fairness for predictive models. Lastly, we provide an overview of how the fairness issues are addressed in machine learning.}}

	\subsection{Scoring Systems}
	
	The use of scoring systems for decision-making can be traced back to the seminal work by \citet{Burgess1928factors} on parole violation where a scoring system is applied to inmates for prediction of parole success and failure. The popularity of scoring systems continues, and they are actively engaged in decision support in today’s environment. Because the coefficients in scoring systems are integer values (i.e., point scores), 
	these tools allow users to make quick predictions by only adding/subtracting or multiplying a few small numbers, without the need for a sophisticated computer or calculator, or for extensive training \citep{zeng2017interpretable}. The ease of calculation and interpretation makes scoring systems excellent decision tools in situations requiring quick and accurate judgments. For example, scoring systems are extensively used in medical diagnosis. {In clinical practice, the model to be used should be simple, readily available and not expensive \citep{patel2016controversies}. Thus, the integer-based scoring systems, which are much simpler to use and less expensive to implement compared to the complex or cumbersome algorithms, have been prevalent \citep{konishi2022new}. } Often times, medical professionals need to make a quick assessment of a patient’s health state at the bedside of the sick patient, and scoring systems are then applied to patients’ clinical data such as vital signs. Medical professionals subsequently perform a few simple calculations (addition or subtraction), which enables them to make a quick evaluation of the patient's condition \citep{strand2008severity}. {Therefore, especially in the data-poor environments such as outpatient settings, where clinicians do not have guaranteed access to computers, the scoring systems are particularly useful \citep{bjarnadottir2018predicting}. In fact, there are many countries in the word have already mandated the use of scoring systems in medical diagnosis applications \citep{breslow2012severity}.}
	
	Scoring systems can be constructed via multiple approaches. In some cases, construction is based on the experience and domain knowledge of a panel of subject matter experts (e.g., the APACHE I by \citet{Knaus1981APACHE}). A prevalent method to construct a scoring system is data-driven \citep{struck2017association}, where scoring systems are usually derived using regression models followed by the rounding of coefficients to obtain integer-valued point scores (e.g., the SAPS II by \citet{LeGall1993}). In addition to traditional statistical approaches, more and more machine learning techniques are introduced to construct scoring systems based on big data \citep{Hurley2017credit,chen2021finance}. For example, \citet{dumitrescu2021machine} use information extracted from various short-depth decision trees to improve the performance of logistic regression when constructing a credit scoring system. A typical standard procedure to develop a data-driven scoring system is illustrated in Figure \ref{fig:flow_scoring}.

	\subsection{Fairness Issues in Predictive Models}
	
	In recent years, a critical issue has been raised concerning predictive models. Many predictive models are criticized for the lack of incorporation of fairness dimensions \citep{zou2018ai}. Scoring systems, as a type of predictive model, are no exception. The aforementioned examples highlight people’s concerns over group fairness and give rise to growing public scrutiny in the decision-making systems that impact resource access and allocation. 
	In response to these concerns, several laws and regulations have been established to ensure group fairness for some high-stake domains such as credit, housing, education, healthcare, and employment (i.e., the Equal Credit Opportunity Act, Equal Employment Opportunity Act, Fair Housing Act, the Affordable Care Act, and General Data Protection Regulation, etc.).
	
	There are three common types of group (un)fairness discussed in literature \citep{Zafar2019fairness}: \textit{disparate treatmen}t, \textit{disparate impact}, and \textit{disparate mistreatment}. 
	\textit{Disparate treatment} is concerned with procedural discrimination, whereas the other two types (\textit{disparate impact} and \textit{disparate mistreatment}) are concerned with outcome discrimination. 
	
	We note that legal actions/legislation, in general, prohibit\textit{ disparate treatment}, i.e., treating individuals differently based on membership in certain groups (e.g., race or gender), and intent to discriminate \citep{Kallus2021assessing}. As we mentioned earlier, \textit{disparate treatment} is only concerned with procedural discrimination \citep{fu2020artificial}, and there is no guarantee for the elimination of outcome discrimination \citep{kleinberg2017inherent}. For example, although the legislation prohibits any explicit use of certain sensitive features (race, gender, or religion, etc.) in the construction {and threshold-setting} of predictive models \citep{Barocas2016big}, the outcome could still reflect discrimination along these sensitive features. One of the reasons is that even if sensitive features are excluded from inputs, there usually exist other attributes that are highly correlated to the precluded features. Algorithms can be designed to incorporate these seemingly neutral attributes to serve as proxies of sensitive features, thereby circumventing existing non-discriminatory legislation and allowing the systematic denial of resource access to certain groups \citep{Hurley2017credit}.

	This paper is motivated by group fairness concerns in existing research and aims to address outcome (un)fairness. We focus on two major types of issues related to outcome (un)fairness: \textit{disparate impact} and \textit{disparate mistreatment}. \textit{Disparate impact} refers to cases where a system adversely affects the members from one group more than another, even if it appears to be neutral  \citep{lambrecht2019algorithmic}. \textit{Disparate mistreatment} refers to cases where the misclassification rates differ for groups of individuals with different memberships \citep{Zafar2017fairness}. Both of these two outcome fairness types have received much attention in the literature \citep{harris2019approximation,wick2019unlocking}. {By deriving fairness-aware scoring systems considering these fairness types, our research contributes to the solution of some salient societal problems such as racial or gender discrimination, which has severe negative implications for disadvantaged groups and the rights and opportunities of their members \citep{Dover2015does}. In the following section,} we review existing literature on how these fairness issues are addressed in machine learning. 
	
	\subsection{Fairness Research in Machine Learning}
	
	Fairness issues associated with machine learning algorithms have attracted increasing attention \citep{Mehrabi2021survey}. Much of the focus is on classification scenarios where a disadvantaged group suffers from discrimination through a classifier. In this regard, past work has been conducted to formalize the concept of algorithmic fairness, such as \textit{statistical parity} \citep{Dwork2012,Corbett2017}, \textit{conditional statistical parity} \citep{Corbett2017}, \textit{equality of opportunity}, and \textit{equalized odds} \citep{Hardt2016}, etc. Based on these concepts, various algorithmic interventions are designed to implement the fairness requirements. 
	
	These algorithmic interventions can be mainly categorized into three groups, depending on the stage when intervention is implemented:  pre-processing, in-processing, and post-processing \citep{Barocas2017fairnesstutorial}. Our proposed algorithm belongs to the in-processing group.
	
	The key focus of existing research investigating in-processing algorithmic interventions is to solve a constrained optimization problem, by imposing a constraint on the fairness level, while optimizing the learning objective such as accuracy \citep{celis2019classification}. However, because most fairness metrics are non-convex due to the use of the indicator function, it is difficult to solve the master optimization problem. A widely used strategy to achieve convexity is to adopt surrogate functions for both objectives and fairness constraints. Examples of this scheme include \citet{Woodworth2017learning,Zafar2017fairness,Zafar2017fairness2,Donini2018,yona2018probably,Zafar2019fairness,Hossain2020designing}. 
	Most of these studies are limited to a single notion of fairness or support only a single sensitive feature, which limits their  generality \citep{Kozodoi2021fairness}. Although several attempts have been made to develop a unified framework that can handle more than one fairness notion \citep{Quadrianto2017recycling,Zafar2017fairness,Zafar2019fairness}, they still utilize surrogate functions instead of hard loss to avoid non-convex optimization. This may lead to sub-par fairness and the sub-optimality of the produced classifier \citep{Lohaus2020too}.  In addition, previous research usually assumes the classifier coefficients are continuous and is hard to control the sparsity, which becomes an obstacle to creating scoring systems. Our proposed framework directly applies 0-1 hard loss without approximation, thereby overcoming the above flaws when constructing scoring systems.

	The remainder of this paper is organized as follows. In Section \ref{sec:2_MotEx}, we motivate our research using a real-life example in medical diagnosis, and provide detailed illustrations regarding \textit{disparate impact} and \textit{disparate mistreatment}, two common types of outcome unfairness. In Section \ref{Sec:3}, we develop a general framework to construct a fairness-aware scoring system with the help of a social welfare function and mixed integer programming. We present in Section \ref{sec_faircons} how different fairness measures are formulated and incorporated into our framework for different application scenarios. Section \ref{sec:theoretical} derives and describes several theoretical bounds for the proposed method. In Section \ref{sec:experiment}, the experimental study of our approach is carried out on several empirical data sets. Section \ref{sec:conc} concludes. All the technical proofs can be found in online supplementary material.
	
	\section{An Illustrative Example of \textit{Disparate Impact} and \textit{Disparate Mistreatment}}\label{sec:2_MotEx}
	
	In this section, we provide a detailed explanation of the two common types of outcome unfairness (i.e., \textit{disparate impact} and \textit{disparate mistreatment}) in the context of sepsis diagnosis. We first explain these two concepts using simplified, made-up examples. We then use an empirical data set to illustrate the existence of these disparities, 
	which further motivates our research.
	
	\subsection{Sepsis and Scoring Systems in Sepsis Mortality Prediction}
	Sepsis is a severe and widespread syndrome, a leading cause of mortality and morbidity globally \citep{Sweeney2018A}. Due to the high costs associated with treatment, sepsis has posed a significant challenge to healthcare systems worldwide \citep{angus2001epidemiology}. The early prediction of clinical outcomes, such as in-hospital mortality, can save lives, as medical professionals can then respond quickly to sepsis patients at greater risk \citep{mukherjee2017implementation}. Several scoring systems (e.g., SAPS II) have been adopted to assess the severity of sepsis patients' sickness. These tools consider patient vital signs, laboratory results, and demographic statistics as risk factors, and produce severity assessments as outputs. If the outputs of these scoring systems are biased against certain demographic groups, they can result in unfair allocation of medical resources among patients with different sensitive traits. Below we discuss and illustrate the two types of outcome (un)fairness we study: \textit{disparate impact} and \textit{disparate mistreatment}, in the context of the scoring systems used to assess the severity of sepsis. 
	
	\subsection{Disparate Impact and Disparate Mistreatment}
	
\begin{figure*}[htbp]
		\begin{centering}
		\includegraphics[width=370pt]{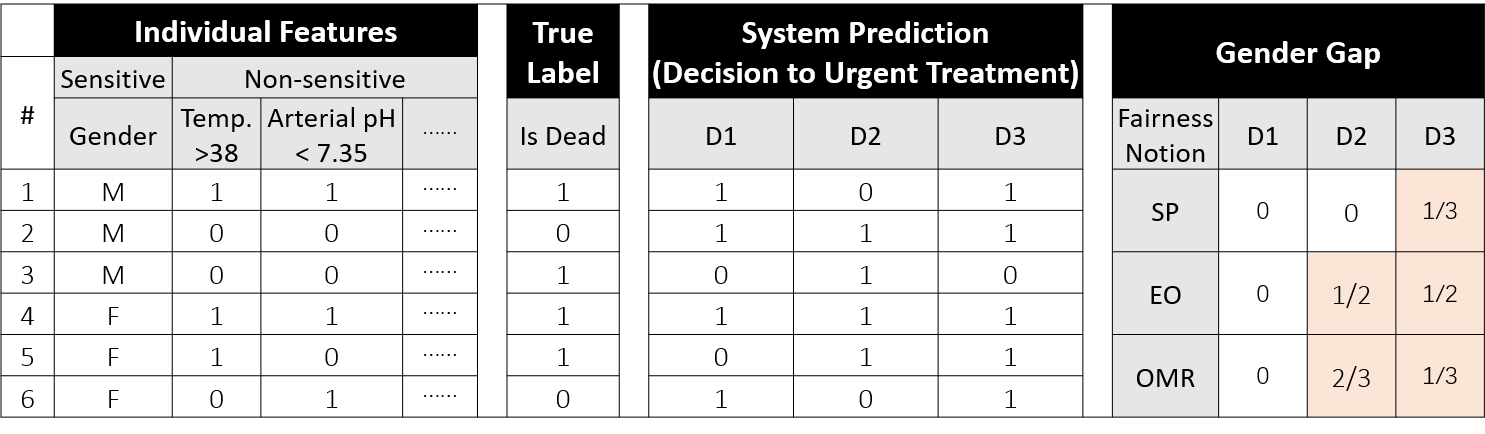}
		\caption{Decisions of three medical scoring systems (i.e., D1, D2, and D3) for sepsis mortality prediction.}
		\label{fig:moti_Ex}
		\end{centering}
		\footnotesize
		\textit{Note}: \textit{Statistical parity} is denoted by SP, \textit{equality of opportunity} by EO, and \textit{equal overall misclassification rate} by OMR.
	\end{figure*}

	Figure \ref{fig:moti_Ex} presents a simplified, made-up case, purely for the purpose of illustrating the concepts of \textit{disparate impact} and \textit{disparate mistreatment}. The made-up example shows the results from three medical scoring systems (denoted by D1, D2, and D3) on a data set containing six patients. This data set includes both non-sensitive features such as body temperature, as well as sensitive features such as gender. In addition, the data set includes a label indicating the in-hospital mortality status of a patient ($1=$ mortality, $0=$ otherwise). The goal of the scoring systems is to predict whether a patient is at a high mortality risk and needs urgent treatment. In what follows, we present different fairness issues through the performance of these scoring systems.
	
	\begin{quotation}
		\begin{description}
			\item[\bf  Disparate Impact:] As we described earlier, the \textit{disparate impact} problem arises if a decision-making system produces results that benefit (or hurt) a group of people with certain sensitive features more frequently than is the case for other groups \citep{Barocas2016big}. The elimination of \textit{disparate impact}  reflects the ability of a decision-making system to achieve \textit{statistical parity} (SP) \citep{Corbett2017}, also known as \textit{demographic parity} \citep{Dwork2012}.

			A sepsis patient benefits from a decision of urgent treatment since this indicates that s/he will be allocated more medical resources. In this case, we deem only system D3 to be unfair due to \textit{disparate impact}. As shown in Figure \ref{fig:moti_Ex}, the fraction of males and females that were predicted to be at high risk by D3 are different ($2/3$ and $1$, respectively). There exists a treatment rate gap of $1/3$ between the two gender groups and thus, D3 does not satisfy \textit{statistical parity}.\footnotemark[1] 
			\footnotetext[1]{Note that these examples are purely used to illustrate the concepts of \textit{disparate impact} and \textit{disparate mistreatment}. In real-life scenarios, decision-makers may have their own preferences regarding which fairness notion to adopt.}
			
			\item[\bf  Disparate Mistreatment:] Recall that \textit{disparate mistreatment} exists if a decision-making system achieves different misclassification error rates for groups of people with different values of sensitive features \citep{Zafar2019fairness}. In addition to the overall error rate, this has been extended to different misclassifications such as false negatives and false positives. Here, we consider two commonly used algorithmic fairness notions in this category, namely \textit{equal overall misclassification rate} (OMR) \citep{Zafar2017fairness} and \textit{equality of opportunity} (EO) \citep{Hardt2016}. The former notion eliminates \textit{disparate mistreatment} by ensuring the same error rate among different groups, while the latter ensures the same false negative rate.
			
			In Figure \ref{fig:moti_Ex}, only D1 is free from \textit{disparate mistreatment} because it has the same false negative and overall error rates for two groups. On contrast, both D2 and D3 are unfair due to \textit{disparate mistreatment} since their rates of erroneous decisions for males and females are different. D2 and D3 both achieve different false negative rates ($1/2$ and $0$) for males and females. D2 also has different error rates ($2/3$ and $0$) for males and females, whereas D3 has rates of $2/3$ and $1/3$.
		\end{description}
		\end{quotation}

		\subsection{Evidence of Disparate Impact and Disparate Mistreatment in Real-Life Sepsis Mortality Prediction}
		Next, we illustrate the existence of \textit{disparate impact} and \textit{disparate mistreatment} with a real-life example in the medical field.  We consider five commonly used scoring systems for sepsis mortality prediction: SAPS II, LODS, SOFA, qSOFA, and SIRS \citep{Sweeney2018A}. We evaluate their performance on (un)fairness through an empirical data set extracted from the Medical Information Mart for Intensive Care database (MIMIC-III)~\citep{johnson2016mimic}. Detailed information regarding the scoring systems and the data set can be found in Section \ref{Experiment_Sepsis}.
		
		Table \ref{Table_moti_Ex} displays the results of (un)fairness checks with respect to different notions of fairness. Columns (3) through (5) of Table \ref{Table_moti_Ex} give the absolute value of rate difference between the two genders, which is calculated based on different algorithmic fairness notions, including SP, EO, and OMR. It demonstrates that there indeed exist disparities between males and females. In particular,  \textit{disparate mistreatment} measured by EO is most evident. The gap in false negative rate between the two groups can reach $11.35\%$ (achieved by SAPS II). This indicates that over $10\%$ more male sepsis patients with a high mortality risk are ignored at the population level compared to female patients. In comparison, the \textit{disparate mistreatment} by OMR and \textit{disparate impact} by SP are less severe. However, the absolute values of rate difference are still up to $4.09\%$ (by SIRS) and $4.22\%$ (by qSOFA), respectively.

		\begin{table} [!htbp]
			\caption{Fairness checks of the existing medical scoring systems on \textit{Sepsis} data set.}
			\label{Table_moti_Ex}
			\scriptsize
			\centering
			\begin{threeparttable}
				\begin{tabular}{lcm{1.4cm}<{\centering}m{1.4cm}<{\centering}m{1.3cm}<{\centering}}
					\toprule
					\multirow{2}{*}{Scoring System}  &\multirow{2}{*}{Accuracy}  &\multicolumn{3}{c}{Fairness Level} \\
					\cmidrule(l){3-5} 
					&&SP&EO&OMR \\
					\midrule
					SAPS II&0.7442 & 0.0064 &0.1135 &0.0235\\
					LODS & 0.7363&0.0323&0.0772&0.0111 \\
					SOFA&0.7209&0.0329&0.0351&0.0081 \\
					qSOFA&0.6304&0.0422&0.0530&0.0052\\
					SIRS&0.5567&0.0011&0.1031&0.0409\\		
					\bottomrule
				\end{tabular}
				\begin{tablenotes}
					\item Note: \textit{Statistical parity} is denoted by SP, \textit{equality of opportunity} by EO, and \textit{equal overall misclassification rate} by OMR.
				\end{tablenotes}		
			\end{threeparttable}
		\end{table}	
		
		
		According to the discussions with subject matter experts in this field, currently, there is no clinical consensus that the gender factor significantly affects the severity and prognosis of this disease. 
		Therefore, many commonly used sepsis screening tools and diagnosis criteria, including the scoring systems investigated in our studies, do not use the gender factor. Besides, the widely used and regularly updated SSC international guidelines~\citep{evans2021surviving} do not suggest any gender-related management for sepsis patients. 

		{It is noteworthy that in this example, even though none of the scoring systems checked above uses the sensitive feature (i.e., gender) as an input, the  prediction results indicate unfairness against people with certain sensitive traits.} {In this application, the \textit{disparate mistreatment} (especially measured by EO) is evident in these scoring systems. This indicates that in the current diagnosis systems, there exists a disparity in misdiagnosis rates between male and female patients who are at high risk of death. Note that this result is obtained regardless of whether there are inherent differences in mortality between two groups, since the EO notion is conditional only on the patients who have high risks (i.e., with death label).} This provides empirical proof of the flaws associated with procedural fairness. A flawed scoring system leads to the unfair allocation of medical resources among patients from different groups. Our empirical findings provide living proof of the importance of considering outcome fairness, and the need to incorporate both fairness and efficiency in developing scoring systems.

	\section{Model Formulation}\label{Sec:3}

	To address the issues with \textit{disparate impact} and \textit{disparate mistreatment}, we develop a new fairness-aware framework. In this section, we first formalize a utilitarian social welfare function that captures these group fairness dimensions. After that, we translate a social welfare maximization task into an empirical risk minimization problem at the center of supervised learning, and then develop a scoring system achieving the optimal social welfare.
	
	\subsection{Model Construction}
	
	Suppose that $\mathcal D=\{(\bs{x}_i,y_i)\}_{i=1}^n$ denotes a data set with $n$ i.i.d. observations, where $\bs{x}_i\in\mathcal X\subseteq\mathbb{R}^{d+1}$ is the $i$th individual's feature vector with the form of $\bs{x}_i=\left[1, x_{i,1}, \dots, x_{i,d}\right]^{T}$ and $y_i\in\mathcal Y=\{-1,1\}$ is the $i$th's class label. 
	Moreover, $s_{i}$ represents the sensitive feature (e.g., race, gender) of individual $i$ with $s_{i}\in\mathcal{S} = \{a_{1}, a_{2},\cdots,a_{c}\}$. Thus, there are $c$ subgroups in the population according to the sensitive feature. 
	{In this paper, we focus on the group-based unfairness which considers the disparities among members of different social groups featured by $s_{i}$.}  We note that $\bs{x}_{i}$ may or may not contain the sensitive feature $s_{i}$ in the actual applications. 
	For simplicity, we assume here that $s_{i}$ is contained in the feature vector $\bs{x}_{i}$. Although a person might be coded with multiple sensitive features in some cases, we will consider only a single sensitive feature in this work. However, we will show that the proposed framework can be directly extended to the case where more than one sensitive feature needs to be considered. 
	In this work, we focus on the linear models of the form $\hat{y}=\text{sign}[\bs{w}^{T}\bs{x}]$, where $\bs{w}=\left[w_{0}, w_{1}, \dots, w_{d}\right]^{T}\in\mathcal{W}$ represents a vector of coefficients  and $w_{0}$ is the intercept term.

	Traditional economics has a long history of using models of \textit{homo economicus} \citep{henrich2001search}. That is, it assumes an individual is completely rational and only cares about his/her own payoffs, but is indifferent about the outcomes of others \citep{Cox2012Direct}. However, there is a large body of experimental and field evidence that contradicts that assumption. Existing research {in behavioral economics} supports that the majority of people are not purely self-interested \citep{babcock1996choosing,Charness2002,Bandiera2005,Benjamin2010}, they care about other people, and are driven by fairness considerations \citep{Fehr1999A,Dawes2007,Tricomi2010}. {The self-interest assumption must sometimes be appended to account for interdependent preferences such as fairness \citep{hamman2010self}.} Therefore, it is reasonable to assume that an individual’s utility function does not only depend on the system outcome regulated by $\bs{w}$, but also on its level of (un)fairness $\delta$,\footnotemark[2]	\footnotetext[2]{Here, $\delta$ actually represents the maximal level of unfairness of the system outcomes. Thus, a smaller value of $\delta$ implies a higher fairness level (i.e., smaller disparities among groups).}  which is represented as $U_{i}(\bs{w},\delta)$. Then, a decision-maker wishes to maximize the following social welfare function given as a weighted sum of individual utilities,

	\begin{align}\label{SWF}
	SWF(\bs{w}, \delta)=\sum_{i=1}^{n}\zeta_{i}U_{i}(\bs{w},\delta),
	\end{align}
	where $\zeta_{i}\in[0,1]$ is the social weight that represents the value placed by society on the $i$th individual's welfare and is normalized so that $\sum_{i=1}^{n}\zeta_{i}=1$.

	Social preference research captures such departures from narrow self-interest \citep{hamman2010self}. The distributional preferences model in social preference usually assumes that people are not only self-interested but also concerned over the inequity between their and others’ outcomes, and an individual’s utility function is a linear combination of these two parts \citep{Fehr1999A,Charness2002}. 
	Built on this, we consider a setting where the utility of an individual $i$ is additively separable into data utility and fairness utility:
	\begin{align}\label{SWF1}
	U_{i}(\bs{w},\delta)=u_{i}(\bs{w})+v_{i}(\delta),
	\end{align}
	where $u_{i}$ is data utility that reflects an individual $i$'s own payoff owing to system output, and $v_{i}$ is fairness utility that represents the influence of (un)fairness 
	on $i$. 
	For simplicity, let us consider the following case with linear utility functions:
	\begin{align}\label{SWF3}
	u_{i}(\bs{w})&=a_{i}-b_{i}\mathbbm{1}\left[y_{i} \bs{w}^{T} \bs{x}_{i} \leq 0\right],\\\label{SWF2}
	v_{i}(\delta)&=-\rho_{i}\delta,
	\end{align}
	where both $a_{i}\geq0$ and $b_{i}\geq 0$.\footnotemark[3] \footnotetext[3]{This study considers the simple case where the values of $a_{i}$, $b_{i}$, and $\rho_{i}$ are given in advance by decision-makers based on their knowledge or experience. Moreover, the value of $b_{i}$ can be set to be either cost-sensitive or cost-insensitive depending on the actual needs of the decision-makers.}
	
	Let us discuss the Eqs. \eqref{SWF3} and \eqref{SWF2} for a thorough understanding of this framework.

	\begin{quotation}
		\begin{description}
			\item {\bf Data Utility}: We first introduce data utility. For a person $i$, his/her data utility depends on whether the system classifies him/her correctly or not. Here, we assume that misclassification usually reduces a person’s data utility. Take the medical diagnosis as an example. Misdiagnosing a sick patient may lead to premature death due to a lack of appropriate treatment. On the other hand, incorrectly diagnosing a healthy person will also cost him/her time and money for unnecessary treatments, including possible invasive procedures, thus his/her utility will also decrease. Specifically, as shown in (\ref{SWF3}),  if $i$ is correctly classified, s/he will receive the positive data utility with $a_{i}$. If $i$ is misclassified, the data utility decreases to $a_i-b_{i}$. 
			
			\item {\bf Fairness Utility}: Next, we specify  fairness utility. In equation (\ref{SWF2}), $\rho_{i}$ is $i$'s preference weight placed on fairness. It reflects a person’s attitude towards unfairness {of outcomes} among groups. When $\rho_{i}=0$, it falls on the classical assumption of people as purely self-interested individuals. Next, we discuss the case where $\rho_{i}\neq 0$. When $i$ is a member of the advantaged group, if $\rho_{i}>0$, it is a weight that reflects his/her benevolence towards the less advantaged. In this case, $i$ is willing to sacrifice his/her own data utility  to help the disadvantaged group. On the other hand, $\rho_{i}<0$ represents $i$'s sense of ``competition.'' It means $i$ prefers to maintain a bigger gap among groups. In the case where $i$ belongs to the less  advantaged group, if $\rho_{i}>0$, it is a weight that reflects his/her ``hostility'' towards the advantaged. In this case, $i$ is unwilling to accept the inequity among outcomes of different groups, and hopes to reduce the gap among groups. If $\rho_{i}<0$, it shows the “friendliness” of $i$. That means even though $i$ is in a disadvantaged group, s/he is willing to see more benefits attributed to the advantaged group.
		\end{description}	
	\end{quotation}

	Afterwards, combining (\ref{SWF})-(\ref{SWF2}) leads to
	\begin{align}\label{eff-fair}
	SWF(\bs{w}, \delta)&=\sum_{i=1}^{n} u_{i}(\bs{w})+\sum_{i=1}^{n} v_{i}({\delta})\\\nonumber&=\sum_{i=1}^{n}\zeta_{i}\Big[a_{i}-b_{i}\mathbbm{1}\left[y_{i} \bs{w}^{T} \bs{x}_{i} \leq 0\right]-\rho_{i}\delta\Big]\\\nonumber&=\sum_{i=1}^{n}\zeta_{i}a_{i}-\sum_{i=1}^{n}\zeta_{i}b_{i}\mathbbm{1}\left[y_{i} \bs{w}^{T} \bs{x}_{i} \leq 0\right]-\delta\sum_{i=1}^{n}\zeta_{i}\rho_{i}.
	\end{align}
	Note that in Eq (\ref{eff-fair}), the first term on the right-hand side is the sum of data utilities over population, which reflects the efficiency of the system $\bs{w}$. The second term is the sum of fairness utilities. Further, we apply the utilitarian social welfare function in which all people are treated the same and social weights are equal across all individuals: $\zeta_{i}= \dfrac{1}{n}$ for all $i$.\footnotemark[4] \footnotetext[4]{For simplicity, we consider the case where the value of $\zeta_{i}$ is given in advance by decision-makers based on their experience or the requirements of applications. Here, we just simply utilize the equal weights for all individuals to mainly present the basic idea of the model construction. In practice, decision-makers could also apply other social weights in our framework based on their needs, and solve the model accordingly. The learning method proposed in this paper is still applicable on other weight choices.} Hence, finding a classifier to maximize social welfare in this case is equivalent to solving the following optimization problem:
	\begin{align}\label{framework1}
	\min_{\bs{w},\delta}& \quad\dfrac{1}{n}\sum_{i=1}^{n}b_{i}\mathbbm{1}\left[y_{i} \bs{w}^{T} \bs{x}_{i} \leq 0\right]+\dfrac{1}{n}\delta\sum_{i=1}^{n}\rho_{i},\\\label{fair_cons}
	\text{s.t.}& \quad g(\bs{w},\mathcal D)\leq \delta,\\\nonumber
	&   \quad \bs{w}\in\mathcal{W},
	\end{align} 
	where (\ref{fair_cons}) is the fairness constraint with $g(\bs{w},\mathcal D)$ encoding a specific fairness measure, $\delta$ is the maximal unfairness level (achieved by the system $\bs{w}$) that the users need to tolerate, and $\mathcal{W}$ encodes hard qualities that must be satisfied by the coefficients. Note that in problem (\ref{framework1}), the objective function consists of two parts. The first part is the average value of the weighted 0-1 loss which penalizes misclassification, and could be regarded as weighted error rate over data set. The second part reflects the ``penalty" for unfairness. Thus, the above optimization problem could be adapted into a regularized empirical risk minimization (ERM) framework in machine learning as follows:
	\begin{align}\label{framework_our}
	\min _{\bs{w},\delta} & \quad \dfrac{1}{n}\sum_{i=1}^{n}b_{i}\mathbbm{1}\left[y_{i} \bs{w}^{T} \bs{x}_{i} \leq 0\right]+\bar{\rho}\delta+\lambda_{0}\|\bs{w}\|_{0}+\epsilon\|\bs{w}\|_{1}, \\\nonumber\text{s.t.}& \quad g(\bs{w},\mathcal D)\leq \delta,\\\label{framework_consW}
	&  \quad \bs{w}\in\mathcal{W},
	\end{align}
	where $\bar{\rho}=\dfrac{\sum_{i=1}^{n}\rho_{i}}{n}$ is the average preference for fairness in the population, and $\lambda_0\geq 0$, $\epsilon\geq 0$ are the penalty parameters. Usually, the constraints (\ref{framework_consW}) restrict coefficients to a finite set of discrete values such as $\mathcal{W}=\{-10, \ldots, 10\}^{d+1}$ to output an integer score. In addition to the original objective in problem (\ref{framework1}), two more penalties are added into the problem (\ref{framework_our}). Specifically, $\ell_{0}$-penalty is applied to control the sparsity of the model where $\|\bs{w}\|_{0}=\sum_{j=1}^{d}\mathbbm{1}\left[w_{j}\neq0\right]$ is the number of non-zero coefficients. The classifier tends to include  less coefficients if its weight $\lambda_{0}$ becomes bigger. The $\ell_{1}$-penalty in the objective is used to obtain the coprime coefficients to reduce redundancy, and the $\ell_{1}$-penalty parameter $\epsilon$ should be set small enough to avoid $\ell_{1}$-regularization.

	Now, we have cast the social welfare maximization problem prevalent in economics as a regularized ERM task in machine learning to derive a fairness-aware scoring system. {In fact, the proposed framework could further degenerate to several commonly used classification approaches in the machine learning field with some choices of model parameters. More details about these simplified models are deferred to the supplementary material.}

	\subsection{Formulation of Mixed Integer Programming}\label{subsec_3.1}
	
	Unfortunately, solving the problem given by (\ref{framework_our}) is challenging. The indicator functions in (\ref{framework_our}) are non-continuous and non-convex functions of the classifier coefficient $\bs{w}$, therefore leading to non-convex formulations, which are difficult to solve directly. 
	To overcome this hurdle, we reformulate the problem (\ref{framework_our}) into the following mixed integer programming (MIP) task to recover the optimal fairness-aware scoring systems:
	\begin{subequations}
		\begin{align}\label{framework_IP}
		\min_{\bs{w}, \bs{\psi},\bs{\Phi}, \bs{\alpha}, \bs{\beta}, \delta}\  \dfrac{1}{n}\sum_{i=1}^{n}b_{i}\psi_{i}&+\sum_{j=1}^{d}\Phi_{j}+\bar{\rho}\delta,\\\label{0-1cons}
		\text{s.t.} \qquad\quad M_{i} \psi_{i} &\geq \gamma-\sum_{j=0}^{d} y_{i} w_{j} x_{i, j} \quad &i&=1, \ldots, N \qquad \qquad &  &\textit{0-1 loss,}\\\label{faircons}
		G(\psi , &\mathcal D_{p},\mathcal D_{q})\leq \delta&p&, q=1, \ldots, c \qquad\qquad &  &\textit{outcome fairness,}\\\label{proc_faircons}
		\alpha_{s}&=0 &s& \in \{1, \ldots, d\}  &  &\textit{procedural fairness,}\\
		\label{penalty}
		\Phi_{j}&=\lambda_{0} \alpha_{j}+\epsilon \beta_{j} &j&=1, \ldots, d  &  &\textit{coef.  penalty,}\\\label{l0cons}
		-\Omega_{j} \alpha_{j} &\leq w_{j} \leq \Omega_{j} \alpha_{j} & j&=1, \ldots, d  &  &\ell_{0}{\textit{-norm,}}\\\label{l1cons}
		-\beta_{j} &\leq w_{j} \leq \beta_{j} & j&=1, \ldots, d &  &\ell_{1}{\textit{-norm,}}\\\nonumber
		w_{j} &\in \mathcal{W}_{j} & j&=0, \ldots, d &&\textit{coefficient set,}\\\nonumber
		\psi_{i} &\in\{0,1\} & i&=1, \ldots, N & &\textit{loss variables,}\\\nonumber
		\Phi_{j} &\in \mathbb{R}_{+} & j&=1, \ldots, d & &\textit{penalty variables,}\\\label{l0_penalty}
		\alpha_{j} &\in\{0,1\} & j&=1, \ldots, d & &\ell_{0}{\textit{ variables,}}\\\nonumber
		\beta_{j} &\in \mathbb{R}_{+} & j&=1, \ldots, d& &\ell_{1}{\textit{ variables,}}\\\nonumber\delta&\in[0,1] &&&&\textit{(un)fairness level}.
		\end{align}
	\end{subequations}
	Here, $\mathcal D_{p}=\{(\bs{x}_i,y_i)\}_{s_{i}=a_{p}}$ and $\mathcal D_{q}=\{(\bs{x}_i,y_i)\}_{s_{i}=a_{q}}$ are individuals from any two different groups $p$ and $q$, and $\Omega_{j}=\max_{w_{j}\in\mathcal{W}_{j}}|w_{j}|$ is defined as the largest absolute value of $w_j$. 
	
	In this formulation, constraint set (\ref{0-1cons}) uses Big-M constraints for $0$-$1$ loss to set the loss variables $\psi_{i}=\mathbbm{1}\left[y_{i} \bs{w}^{T} \bs{x}_{i} \leq 0\right]$ to $1$ if the $i$th example is misclassified by the classifier $\bs{w}$. The Big-M constant \citep{Wolsey1998integer} $M_{i}$ can be set as $M_{i}=\max_{\bs{w}\in\mathcal{W}}(\gamma-y_{i}\bs{w}^{T}\bs{x}_{i})$, and its computation is simple since $\bs{w}$ is restricted to a finite set. The value of $\gamma$ could be set to a small positive number which is not greater than a lower bound on $|y_{i} \bs{w}^{T} \bs{x}_{i}|$ (i.e., $0<\gamma\leq\min_{i}|y_{i} \bs{w}^{T} \bs{x}_{i}|$). When the features are binary, $\gamma$ can be set to any value between $0$ and $1$ since the coefficients are integers. In other cases, $\gamma$ might be set arbitrarily based on an implicit assumption on the values of features \citep{zeng2017interpretable}. With this setting, if example $i$ is misclassified, the value of the right-hand side of the inequality (\ref{0-1cons}) is positive. Thus, $\psi_{i}$ has to be $1$ to satisfy the inequality. On the contrary, if $i$ is classified correctly, we have $\gamma-\sum_{j=0}^{d} y_{i} w_{j} x_{i, j}\leq 0$. In this case, the value of $\psi_{i}$ could be $0$ or $1$. However, since the bigger value of  $\psi_{i}$ results in more penalty in the objective,  $\psi_{i}$ will be forced to equal to $0$ in this case. Therefore, $\psi_{i}$ will work as an indicator to show whether the $i$ is misclassified or not. 
	
	To evaluate the unfairness level achieved by $\bs{w}$, we focus on several generally known fairness notions proposed via the machine learning community, which are evaluated over sub-population groups. Constraint set (\ref{faircons}) encodes the fairness assessment as inequalities among any two different groups $p$ and $q$ in society. Its explicit expressions $G(\cdot)$ depend on the given fairness notion and will be presented in detail in Section \ref{sec_faircons}. In addition, constraint (\ref{proc_faircons}) ensures the sensitive feature is not incorporated into the final classifier, and thus guarantees \textit{procedural fairness} at the same time.\footnotemark[5] \footnotetext[5]{In particular, constraint (\ref{proc_faircons}) could be removed in certain application scenarios where laws or regulations exceptionally allow the use of sensitive features for decision-making.} Constraint set (\ref{penalty}) represents the total penalty assigned to each coefficient, where $\alpha_{j}=\mathbbm{1}\left[w_{j}\neq0\right]$ defined by (\ref{l0cons}) encodes the $\ell_{0}$-penalty and $\beta_{j}=|w_{j}|$ defined by (\ref{l1cons}) encodes the $\ell_{1}$-penalty. 
	
	Although the proposed framework mainly focuses on the considerations regarding fairness, it also allows decision-makers to implement a variety of application-specific constraints into its MIP formulation. Remark \ref{remark_2} shows some examples of application-specific constraints that can be encoded into this method. Note that our framework could also handle multiple application-specific constraints at the same time. Thus, this framework provides decision-makers with great flexibility for their model customization in a simple way. 
	\begin{remark}\label{remark_2}
		The MIP formulation ensures that several types of application-specific constraints could be implemented. We specify here some common choices for different applications. 
		\begin{itemize}
			\item [1)] Model Size Control:
			we could limit the number of input features with the help of the indicator variables $\alpha_{j}$ by adding the constraint: $A_{l}\leq\sum_{j=1}^{d}\alpha_{j}\leq A_{u}$, where $ A_{l}$ is the lower bound and $ A_{u}$ is the upper bound of the model size, respectively.
			\item [2)] Logical Relationship:
			some logical structures such as ``if-then" constraints can be implemented. For example, to ensure that a classifier will only contain features $\alpha_{j}$ and $\alpha_{k}$ if it also contains the feature $\alpha_{l}$, we can encode this as $\alpha_{j}+\alpha_{k}\leq 2\alpha_{l}$.
			\item [3)]Domain Knowledge: some established relationships between input features and the outcome could be pre-specified with sign constraints in this model. For example, if the feature $j$ is a well-known factor for specific outcomes (e.g., excess body weight usually causes a higher risk of type 2 diabetes),  
			this positive or negative relationship could be set by adding $w_{j}>0$ or $w_{j}<0$, respectively.
			
			\item [4)] Preference for Feature Selection: practitioners may have  preferences between different features. For the hard preferences where practitioners insist on incorporating a feature $j$ into the final scorecard \citep{reilly2006translating}, we can set $\alpha_{j}=1$ to ensure the feature will be included. The soft preferences between features could be also realized by adjusting the value of $\lambda_{0}$ for different features. For example, if we prefer feature $j$ over feature $k$ to some extent, we can express this requirement as $\lambda_{0,k}=\lambda_{0,j}+\Lambda$, where $\Lambda>0$ shows the maximal additional social welfare loss we could tolerate for using feature $j$ instead of feature $k$. In this way, the feature $k$ will be used only if it brings additional welfare gain greater than $\Lambda$. This approach can also be used to deal with the problem of missing values in data set \citep{UstunRu2016SLIM}.

		\end{itemize}  
	\end{remark}

	\section{Fairness Constraints}\label{sec_faircons}

	In Section \ref{sec:2_MotEx}, we have provided intuitive illustrations of \textit{disparate impact} and \textit{disparate mistreatment}, in the context of sepsis morality prediction. In this section, we will show how these two concepts are modelled mathematically, and derive fairness metrics based on them. Then we present how to incorporate these fairness metrics into the proposed framework for deriving various fairness-aware scoring systems.

	\subsection{Elimination of Disparate Impact}
	
	As discussed previously, in the algorithmic decision-making context, even though laws strictly prohibit procedural discrimination, algorithms can still produce biased outcomes across groups of different sensitive feature values, thus resulting in disparate impact \citep{Fu2021crowds}. In response, \textit{statistical parity}, one of the fairness notions suggested in the machine learning field, is developed. \textit{Statistical parity}  simply requires the independence of the sensitive feature $s$ and the decision $\hat{y}$. In other words, the system decisions should achieve the same distributions across all demographic groups. Thinking of the event $\hat{y}=1$ as ``acceptance'' in the binary  classification scenario, this notion requires the acceptance rate to be identical for all groups, i.e.,
	$$P(\hat{y}=1 \mid s=a_{p})=P(\hat{y}=1 \mid s=a_{q})$$ for any two different groups $p$ and $q$. Then, a property that a decision system satisfies \textit{statistical parity} between two groups $p$ and $q$ up to bias $\delta$ could be expressed as:
	$$\Big|P(\hat{y}=1 \mid s=a_{p})-P(\hat{y}=1 \mid s=a_{q})\Big|\leq\delta$$for any $p, q=1, \ldots, c$. 
	
	Representing this inequality empirically leads to
	\begin{align}\label{SP_em}
	\left|\dfrac{1}{N_{a_{p}}}\right.&\left[\sum_{i \in I_{a_{p}}^{-}} \mathbbm{1}\left[y_{i} \bs{w}^{T} \bs{x}_{i} \leq 0\right]+N_{a_{p}}^{+}-\sum_{i \in I_{a_{p}}^{+}} \mathbbm{1}\left[y_{i} \bs{w}^{T} \bs{x}_{i} \leq 0\right]\right]\\\nonumber&\left.-\dfrac{1}{N_{a_{q}}}\left[\sum_{i \in I_{a_{q}}^{-}} \mathbbm{1}\left[y_{i} \bs{w}^{T} \bs{x}_{i} \leq 0\right]+N_{a_{q}}^{+}-\sum_{i \in I_{a_{q}}^{+}} \mathbbm{1}\left[y_{i} \bs{w}^{T} \bs{x}_{i} \leq 0\right]\right]\right|\leq \delta,
	\end{align}
	where $I_{a_{p}}^{+}=\big\{i\in\{1,2,\dots,n\}\left|s_{i}=a_{p}, y_{i}=1\right.\big\}$, $I_{a_{p}}^{-}=\big\{i\in\{1,2,\dots,n\}\left|s_{i}=a_{p}, y_{i}=-1\right.\big\}$,  $N_{a_{p}}=|I_{a_{p}}^{+}\cup I_{a_{p}}^{-}|$ and $N_{a_{p}}^{+}=|I_{a_{p}}^{+}|$ for any $p=1, \ldots, c$.
	
	For any two different groups $p, q=1, \ldots, c$, the left-hand side of (\ref{SP_em}) could be re-expressed by the indicator variables $\psi_{i}$ in Section \ref{subsec_3.1} as follows:
	$$G_{SP}=\left|\left(\frac{N_{a_{p}}^{+}}{N_{a_{p}}}-\frac{N_{a_{q}}^{+}}{N_{a_{q}}}\right)+\frac{1}{N_{a_{p}}}\left[\sum_{i \in I_{a_{p}}^{-}} \psi_{i}-\sum_{i \in I_{a_{p}}^{+} } \psi_{i}\right]-\frac{1}{N_{a_{q}}}\left[\sum_{i \in I_{a_{q}}^{-}} \psi_{i}-\sum_{i \in I_{a_{q}}^{+}} \psi_{i}\right]\right|.$$ Then, we can rewrite the fairness constraint (\ref{SP_em}) for \textit{statistical parity} as $G_{SP}\leq\delta$. Incorporating this inequality into (\ref{faircons}), we can derive a fairness-aware scoring system based on this notion.
	
	Note that \textit{statistical parity} is well-suited to contexts such as employment or school admissions, where it may be desirable or required by laws or regulations for diversity or affirmative action \citep{Chouldechova2017fair,Lohaus2020too}. In these situations, selecting individuals proportionally across racial, gender, or geographical groups might be necessary. Moreover, because this notion is independent of the target value $y$, it is also appealing in applications where there does not exist the ground-truth information for decisions, or where the historical decisions used for training are biased themselves and thus cannot be trusted \citep{Zafar2019fairness}. Implementing \textit{statistical parity} will aid the prevention of discrimination based on redundant encoding \citep{Dwork2012}. It may also help to level the playing field and benefit the disadvantaged group in the long run \citep{Hu2018short}. {However, this fairness notion might be inadequate in some cases. When disproportionality is truly present and independent from a sensitive feature, enforcing \textit{statistical parity} requires us to reject qualified candidates from one group and/or approve unqualified candidates from the other group. This risk introduces reverse discrimination against qualified individuals \citep{Zafar2017fairness}. In addition, since this notion ignores any possible correlation between $y$ and $s$, it may reject the optimal classifier $\hat{y}=y$ when base rates are different (i.e., $P(y=1|s=a_{p})\neq P(y=1|s=a_{q})$).} 

	\subsection{Elimination of Disparate Mistreatment}
	
	Disparate mistreatment has been extended to different types of misclassification such as false negatives and false positives, in addition to the general misclassification rate. Here we consider two frequently used fairness notions in the machine learning community: \textit{equal overall misclassification rate} (OMR) \citep{Zafar2017fairness} and \textit{equality of opportunity} (EO) \citep{Hardt2016}.

	\subsubsection{Equal Overall Misclassification Rate}
	This notion is also known as \textit{accuracy parity} \citep{Zhao2019inherent}, which requires the error rate to be same among all groups. It can be expressed as
	$$P(\hat{y} \neq y \mid s=a_{p})=P(\hat{y} \neq y \mid s=a_{q})$$
	for any $p, q=1, \ldots, c$.
	
	Then, an algorithmic decision-making system that satisfies equal OMR between any two groups $p$ and $q$ up to bias $\delta$ could be expressed as follows:

	$$\Big|P(\hat{y} \neq y \mid s=a_{p})-P(\hat{y} \neq y \mid s=a_{q})\Big|\leq\delta.$$
	Rewriting this inequality with the indicator variables $\psi_{i}$ used in the MIP formulation gives the fairness constraint for equal OMR as 
	\begin{align}\label{EOMR_cons}
	G_{OMR}=\left| \frac{1}{N_{a_{p}}} \sum_{i \in I_{a_{p}}} \psi_{i}-\frac{1}{N_{a_{q}}} \sum_{i \in I_{a_{q}}} \psi_{i}\right|\leq \delta,
	\end{align}
	where $I_{a_{p}}=\{i\in\{1,2,\dots,n\}\left|s_{i}=a_{p}\right.\}$ for any $p=1, \ldots, c$. 
	
	\subsubsection{Equality of Opportunity}
	The second fairness notion EO aims to ensure that the true positive rate of each group is identical, i.e., TP$_{p}=P\left(\hat{y}=1|s=a_{p}, y=1\right)$ is the same for $\forall p\in\{1,\cdots, c\}$. Note that this is equivalent to requiring that the false negative rate of each group is the same.   
	We also relax this equality requirement and instead require the maximal unfairness level to be tolerated (i.e., $\delta$). We re-express it with $\psi_{i}$ in a way similar to the previous notions. Then, the fairness constraint for EO is given by
	$$G_{EO}=\left| \frac{1}{N_{a_{p}}^{+}} \sum_{i \in I^{+}_{a_{p}}} \psi_{i}-\frac{1}{N_{a_{q}}^{+}} \sum_{i \in I^{+}_{a_{q}}} \psi_{i}\right|\leq\delta$$
	for any $p, q=1, \ldots, c.$
	
	\begin{remark}
		Some other notions of fairness related to the elimination of disparate mistreatment, such as \textit{predictive equality} \citep{Corbett2017} and \textit{equalized odds} \citep{Hardt2016}, could also be implemented directly by constructing corresponding fairness constraint function $G(\cdot)$ with the help of $\psi_{i}$. In this paper, we mainly focus on the aforementioned fairness measurements (equal OMR and EO). Other notions of fairness can be easily tailored for interested readers.  
	\end{remark}
	
	{Since disparate mistreatment slightly relaxes the requirement that $\hat{y}$ is independent of $s$}, it will not rule out the perfect predictor $\hat{y}=y$ even when the base rates differ across groups. In scenarios where ground truth information for decisions is accessible and reliable, it would be possible to distinguish disproportionality in decision outcomes among groups that results from candidates' qualifications and from discrimination (or bias) against certain groups. {Thus, compared to disparate impact, disparate mistreatment would be more applicable to the case in which an outcome to be predicated is correlated with the sensitive feature or there are inherent differences in qualifications between groups.} Disparate mistreatment will effectively avoid reverse discrimination and is widely discussed in healthcare \citep{Rajkomar2018}, criminal justice \citep{Chouldechova2017fair}, and credit fields \citep{Lohaus2020too}. However, it may also be insufficient under certain contexts. For example, \citet{Berk2021fairness} argue that in settings where a cost-weighted approach is required, \textit{equal overall misclassification rate} might be inadequate. \citet{Zhang2019nips} also suggest that enforcing \textit{equality of opportunity} can make the outcomes seem fairer in the short term but can lead to undesirable results in the long run.
	
	At this point, we have derived the specific expressions of fairness constraints for different fairness notions and discussed their proper application contexts. We would like to issue a caution that decision-makers need to match the selection of fairness notion with decision contexts. When this is properly done, one can then apply the expressions of the selected fairness notion in the constraint set (\ref{faircons}) so as to develop the optimal scoring system that maximizes social welfare. 
	
	\begin{remark}
		After incorporating the specific expressions of fairness constraints into the proposed framework (\ref{framework_IP}), the corresponding MIP problems can be solved via commercial optimization solvers such as CPLEX, Gurobi, or CBC. Note that some of these solvers utilize heuristics to speed up the process. Thus, the solving process may be sensitive to specific parameter settings of the solvers, which may affect the convergence rate and the global optimality of the returned solutions. 
	\end{remark}
	
	It is also worth noting that even though the fairness constraint set (\ref{faircons}) is defined on one specific fairness notion, the proposed framework could be easily extended for satisfying multiple fairness notions simultaneously. In certain application scenarios, it might be desirable to evaluate the level of (un)fairness on more than one notion of fairness defined above (e.g., measure the (un)fairness on both disparate impact and disparate mistreatment). In this case, a desirable scoring system could be developed by including the corresponding constraints simultaneously. Furthermore, the proposed framework can also incorporate fairness with respect to multiple sensitive features (e.g., race, gender, religion, disability) simultaneously by including constraints for each sensitive feature. These extensions showcase the high flexibility of the proposed framework. 

	\section{Theoretical Analysis}\label{sec:theoretical}
	
	In this section, we present some theoretical analyses of the proposed scoring systems. First, we show that although a finite discrete set $\mathcal{W}$ is used to construct the scoring system, the total social welfare of the proposed method is not worse than a baseline classifier with real-valued coefficients $\bs{\theta} \in \mathbb{R}^{d+1}$. Then, we show the relationship between the maximum social welfare and the optimal (un)fairness level. Thus, for a target social welfare level, we can roughly estimate the corresponding fairness level. This enables a quick approach to roughly forecast a range of optimal fairness levels for each scoring system. Proofs of the theorems are deferred to the supplementary material.

	Theorem \ref{Th1} indicates that we can always generate a finite discrete coefficients set such that the social welfare of the proposed method with discrete coefficients is even better than the social welfare of a baseline linear classifier with real-valued coefficients. This conclusion has been validated by the experimental study (see Section \ref{sec:experiment}).
	
	\begin{theorem}\label{Th1}
		Let $\bs{\theta}=\left[\theta_{0},\theta_{1}, \dots,\theta_{d}\right]^{T} \in \mathbb{R}^{d+1}$ denote the real-valued coefficients of any linear classifier which is trained with a data set $\mathcal D=\{(\bs{x}_i,y_i)\}_{i=1}^n$ and achieves an (un)fairness level $\delta$ with respect to a given fairness requirement $G(\cdot)$. {The margin achieved by an example $\bs{x}$ is defined as $\dfrac{|\bs{\theta}^{T}\bs{x}|}{\|\bs{\theta}\|_{2}}$.} We also denote $\eta_{(k)}$ as the value of the $k^{th}$ smallest margin achieved by training examples, especially when $k=1$, $\eta_{(1)}=\min_{i} \dfrac{|\bs{\theta}^{T}\bs{x}_{i}|}{\|\bs{\theta}\|_{2}}$. Let $\mathcal{I}_{(k)}=\left\{i\in \{1,2,\dots,n\}\left|\dfrac{|\bs{\theta}^{T}\bs{x}_{i}|}{\|\bs{\theta}\|_{2}}<\eta_{(k)}\right.\right\}$ denote the set of training examples whose margin is smaller than $\eta_{(k)}$, and $X_{(k)}=\max_{i\notin\mathcal{I}_{(k)}}\|\bs{x}_{i}\|_{2}$ denote the largest magnitude of 
		training example $\bs{x}_{i}\in \mathcal D$ for $i\notin\mathcal{I}_{(k)}$.
		
		Fitting a linear classifier via the proposed framework \eqref{framework_our} and restricting coefficients to $\mathcal{W}=\{-\Omega,\dots,\Omega\}^{d+1}$, we suppose that the coefficients $\bs{w}^{*}=\left[w_{0}^{*}, w_{1}^{*}, \dots, w_{d}^{*}\right]^{T}$ are obtained. If the resolution parameter $\Omega$ satisfies
		$$\Omega>\dfrac{X_{(k)}\sqrt{d+1}}{2\eta_{(k)}},$$
		then the difference of total social welfare between the two classifiers is bounded as 
		\begin{align}\label{eq2}
		SWF(\bs{w}^{*})- SWF(\bs{\theta})\geq(1-k)\max_{i\in\mathcal{I}_{(k)}}b_{i}-N\bar{\rho}\Delta_{F}(k),
		\end{align}
		where $\Delta_{F}(k)$ is a function of $k$, whose concrete expression depends on the given fairness metric. 
	\end{theorem}

	Note that when $k=1$, $\Delta_{F}(k)$ equals zero for all three fairness metrics. In this case, according to the proof of Theorem \ref{Th1}, the coefficient set $\mathcal{W}$ contains a classifier with discrete coefficients $\bs{w}$ that achieves the same classification results (and thus the same social welfare) as the baseline classifier with real coefficients $\bs{\theta}$. Since our linear classifier with $\bs{w}^{*}$ is optimal over $\mathcal{W}$, it may attain a better social welfare than $\bs{w}$, and thus is better than the real-valued classifier $\bs{\theta}$. {It is also noteworthy that the Corollary 1 proposed in \citet{UstunRu2016SLIM}, which considers only the accuracy, could be viewed as a special case of Theorem \ref{Th1} by setting $\bar{\rho}=0$ and $b_{i}=1$ for $\forall i$. Thus, the Theorem \ref{Th1} we proposed extends the result from the welfare as well as the fairness perspectives.}

	Consequently, Theorem \ref{Th1} shows that if we select an appropriate coefficient set $\mathcal{W}$, i.e., choose a large enough resolution parameter $\Omega$, the proposed scoring system could achieve even larger total social welfare than real-valued classifiers, {even though its coefficients are restricted to integers.} This may provide guidelines for practitioners on setting model parameters in order to pursue large social welfare.
	
	In Theorem \ref{Th2}, we establish the relationship between the optimal fairness level of the proposed scoring system and the social welfare it can achieve.
	
	\begin{theorem}\label{Th2}
		Let $\bs{w}^{*}$ and $\delta^{*}$ denote the optimal classifier maximizing total social welfare and its corresponding fairness level, respectively. Then, it holds that
		\begin{align}\label{Th2ineq}
		SWF(\bs{w}^{*})>\Delta^{*}(\delta^{*})-\mathbbm{1}\left[\bar{\rho}>0\right]\bar{\rho},
		\end{align}
		where $\Delta^{*}$ is a function of $\delta^{*}$ and its detailed expression depends on the selected fairness metric.
	\end{theorem}
	
	Theorem \ref{Th2} gives the lower bound of the social welfare for the proposed scoring system, as a function of its optimal fairness level $\delta^{*}$. The proof of Theorem \ref{Th2} in the supplementary material presents the concrete expression of $\Delta^{*}$ for all three fairness metrics discussed previously. It is noteworthy that for these fairness metrics, $\Delta^{*}$ is a linear function of $\delta^{*}$. This enables rapid estimation of the optimal (un)fairness level for the scoring system. For example, if a practitioner wants to roughly estimate the maximal unfairness level corresponding to a certain value of social welfare (e.g., SWF$_{1}$), s/he can simply deduce the upper bound of the corresponding unfairness level $\delta^{*}$ by setting the left-hand side of inequality (\ref{Th2ineq}) equal to SWF$_{1}$. This will provide a quick overview of the system's fairness performance, {which may further allow the practitioner to make rapid adjustments to relevant policies or information publishing.}

	\section{Experimental Study}\label{sec:experiment}
	
	In this section, we conduct experiments to validate the proposed method, using several empirical data sets. Results show that our fairness-aware scoring systems are effective in maximizing total welfare. In addition, we highlight two advantages associated with our framework:  interpretability and flexibility.
	
	\subsection{Application of Fairness-Aware Scoring System to Sepsis Mortality Prediction} \label{Experiment_Sepsis}
	
	In Section \ref{sec:2_MotEx}, we discussed the existence of unfairness in scoring systems used to predict sepsis mortality. Below, we provide details on the empirical data set, the construction of fairness-aware scoring systems, and comparison with existing scorecards.

	\subsubsection{Data and Processing}
	
	The \textit{Sepsis} data set is extracted from the Medical Information Mart for Intensive Care database (MIMIC-III)~\citep{johnson2016mimic}. This data repository has been widely used for medical model development and validation \citep{henry2015targeted}.
	The data set includes 2,021 patients with 19  variables. These variables track the worst value of patients’ health indicators within 24hr of ICU admission, and include patients' demographic characteristics such as age and gender. We consider gender (Male/Female) as the sensitive feature.\footnotemark[6] \footnotetext[6]{In this experiment, the sensitive feature (gender) is not excluded from input data since it might be justifiable to use the sensitive feature for decision-making in medical diagnosis, i.e., the constraint (\ref{proc_faircons}) is removed when developing scoring systems on \textit{Sepsis} data.}The outcome variable is in-hospital death: $1$ indicates death and $0$ otherwise. 
	
	We convert the raw variables in the sepsis data set into rule-based binary-coded data using the Rulefit method \citep{friedman2008predictive}, where each column represents whether the attributes satisfy a specific rule. We directly refer to the 77 rules discovered in \citet{ying2021spesis} as final model inputs and use them to predict the outcome variable. More details regarding the \textit{Sepsis} data set can be found in the supplementary material.

	\subsubsection{Model Setting and Baselines}
	
	In this experiment, we train fairness-aware scoring systems (FASSs) for different fairness measures as mentioned in Section \ref{sec_faircons}. To further show the interpretability and flexibility of our approach, we also consider the FASSs model with an additional application-specific constraint that limits the model size (see Remark \ref{remark_2}). Usually, a person can only handle a few cognitive entities at once ($7 \pm 2$ according to \citet{Miller1956magical}). Thus, we set the model size to be at most 7 (denoted by FASS7) so it can be explained and understood by medical professionals in a short period of time.

	For all our methods, the coefficient set is chosen as $\mathcal{W}=\{-10, \ldots, 10\}^{d+1}$ and $a_{i}=b_{i}=1$ for $\forall i=1,\dots,n$ for simplicity. Note that with this setting, the data utility is reduced to accuracy. {In addition, we set $\lambda_{0}\in[7\times 10^{-5}, 9\times 10^{-4}]$ and $\epsilon=0.01$ so that the proposed method will sacrifice little welfare for sparsity.} The CPLEX 12.6.3 is employed to solve the final MIP problem. The experiments also compare our approaches to several commonly used medical scoring systems for in-hospital mortality prediction of sepsis patients in ICUs, as discussed in Section  \ref{sec:2_MotEx}. More specifically, we consider the following baseline scoring systems: SAPS II, LODS, SOFA, qSOFA, and SIRS. The details of the baselines can be found in the supplementary material.
	
	With this set-up, we randomly partition the \textit{Sepsis} data into a training set (70\%) and test set (30\%) and repeat the partition randomly five times to evaluate the average performance of all models, unless otherwise stated. Then, FASS scorecards based on the whole data are produced to show the interpretability.

	\subsubsection{Results}
	Below we discuss our experimental findings. We first address the benefits of our framework in social welfare maximization. We then explain the advantage of our framework in terms of interpretability and flexibility.

	\begin{itemize}
		\item \textbf{Social Welfare Maximization}

		We first compare the performance of our methods in terms of social welfare maximization with the baseline scoring systems mentioned above. In this experiment, the value of average preference for fairness (i.e., $\bar{\rho}$) is set arbitrarily for simplicity. Table \ref{Table_Sepsis_welfare} provides an overview of average values of social welfare for all scoring systems on the \textit{Sepsis} data set. This chart shows clearly that the proposed FASS model consistently achieves the optimal social welfare regardless of which fairness metric is utilized. For example, when OMR measures fairness, FASS can increase total social welfare gains by 10.12\% compared to  LODS (the best one among baselines). We obtained similar findings when fairness is measured by EO or SP. Note that the results show  that compared to FASS (i.e., without model size constraint), FASS7 achieves slightly lower but similar performance. FASS7 still outperforms all baseline models on all three fairness metrics, even though at times the model size of FASS7 is smaller than some baseline models.
		
		In summary, all these results indicate that the proposed scoring system performs effectively in achieving optimal social welfare with different fairness metrics. 
		\begin{table} [!htbp]
			\caption{The average values of total social welfare for all scoring systems on \textit{Sepsis} data set.}
			\label{Table_Sepsis_welfare}
			\scriptsize		
			\centering
			\begin{threeparttable}
				
				\begin{tabular}{cccccccccc}
					\toprule
					\multirow{2}{*}{Dataset}&Fairness&\multirow{2}{*}{$\bar{\rho}$}& \multicolumn{5}{c}{Baselines}  & \multicolumn{2}{c}{Ours} \\
					\cmidrule(l){4-8} \cmidrule(l){9-10} 
					&Notions& & SAPS II& LODS &SOFA&qSOFA&SIRS&  FASS& FASS7\\
					\midrule
					\multirow{3}{*}{\textit{Sepsis}}&SP  &5 & 0.6761&0.5368&0.5638&0.5089&0.4520 &\textbf{0.6916}&0.6869 \\
					&EO  &0.2 & 0.7102&0.7198&0.7096&0.6278&0.5373&\textbf{0.7550}&0.7472\\
					&OMR  &5 & 0.5768&0.6393&0.5787&0.5509&0.4022 &\textbf{0.7405}&0.7223\\ 
					\bottomrule
				\end{tabular}
				\begin{tablenotes}
					\item Note: \textit{Statistical parity} is denoted by SP, \textit{equality of opportunity} by EO, and \textit{equal overall misclassification rate} by OMR. The optimal values are highted in bold.
				\end{tablenotes}		
			\end{threeparttable}
		\end{table}		
	
		In addition, Figure \ref{fig:curves_Sepsis} displays the accuracy and (un)fairness level of all scoring systems. As can be seen from this figure, the proposed models obtain a lower unfairness level compared to the baselines with all fairness metrics. Furthermore, both FASS and FASS7 achieve better accuracy. This indicates that the derived scoring systems outperform the existing medical scoring systems in sepsis mortality prediction. 
		
		\begin{figure*}[!htbp]
			\begin{centering}
				\includegraphics[width=470pt]{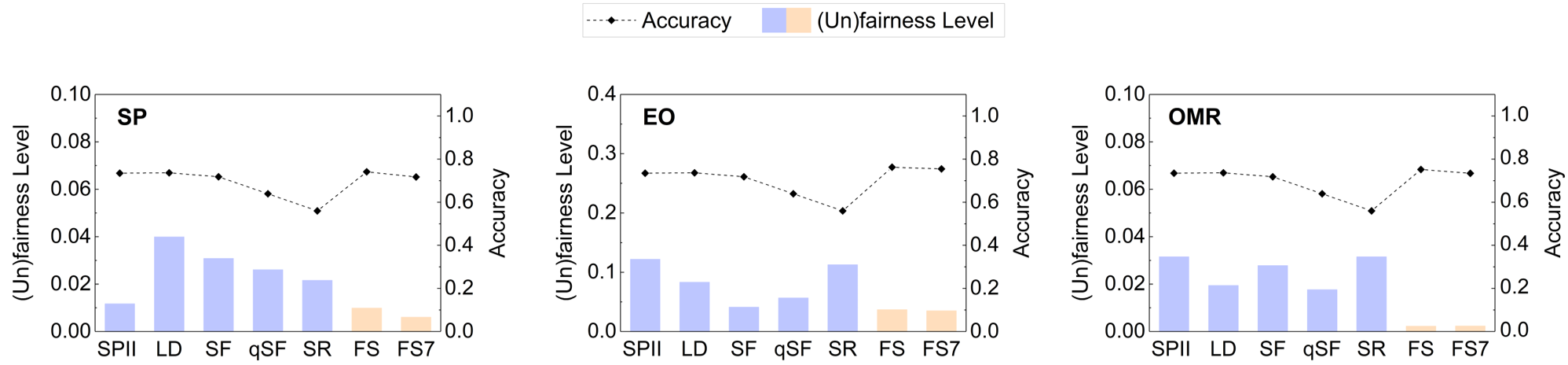}
				\caption{ Accuracy and (un)fairness level of scoring systems.}
				\label{fig:curves_Sepsis}
			\end{centering}
			\footnotesize
			\textit{Note}: SPII, LD, SF, qSF, and SR stand for SAPS II, LODS, SOFA, qSOFA, and SIRS, respectively.  FS and FS7 stand for FASS and FASS7, respectively.
		\end{figure*}

		\item \textbf{Interpretability and Flexibility}
		
		Interpretability is often a critical factor when machine learning models are utilized in a product or a decision process in practice \citep{molnar2020interpretable}. A good practical predictive model should be interpretable and also reasonable. {Especially in healthcare applications, \cite{bjarnadottir2018predicting} highlight that simpler and interpretable models are often preferred by a physician.} \cite{than2014development} also emphasize the importance of ``sensibility" of produced models in medical applications, where sensibility refers to whether a prediction rule is both clinically reasonable and easy to use.

		To show interpretability as well as sensibility, we focus on a FASS7 model. The choice of FASS7 model is to provide a scenario where we sacrifice some performance in order to improve interpretability and ease of use by restricting the model size. Figure \ref{fig:scorecard_Sepsis} shows the final scorecard produced by FASS7 on the whole \textit{Sepsis} data set. For brevity of illustration, we only represent the EO scorecard developed based on FASS7. We defer the SP and OMR scorecards  to the supplementary material.  As shown in Figure \ref{fig:scorecard_Sepsis}, the EO scorecard identifies two risk-increasing rules (rules with positive points) and five risk-decreasing rules (rules with negative points). Each rule consists of two or three conditions, and each condition consists of a variable and a related cut-off value (e.g., 7.2 for pH.art). If all the conditions are satisfied, a rule is endorsed, and we add or subtract the corresponding score. A higher total score indicates a greater risk of in-hospital death. We find that the risk tendency of the results is in line with the findings in \citet{ying2021spesis}.

		\begin{figure*}[!htbp]
			\centering
			\includegraphics[width=190pt]{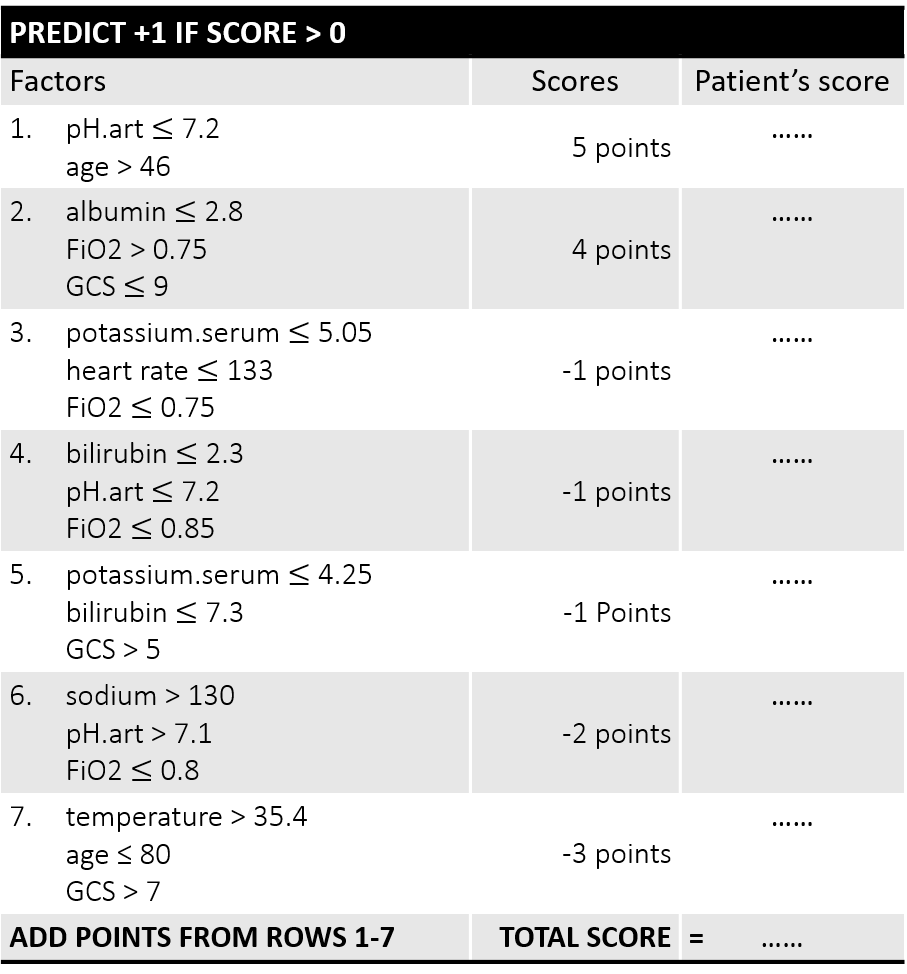}
			\caption{EO scorecard developed by FASS7 for sepsis mortality prediction.}
			\label{fig:scorecard_Sepsis}
		\end{figure*}

		Note that there are 77 rules that can be used to predict sepsis mortality. Our EO scorecard based on FASS7 is able to identify the seven most important predictors, providing  clinical sensibility. Specifically, the scorecard recognizes several indicators with cut-off values supported by medical research.  
		For example, the risk factors most involved in the scorecard are FiO2 (Fraction of inspired oxygen) and pH.art (arterial pH) with cut-off values at around 0.8 and 7.1, respectively. FiO2 is routinely measured in ICUs to assess patient pulmonary function and the presence or severity degree of sepsis-related respiratory dysfunction \citep{santana2013sao}. The scorecard indicates that a FiO2 level  higher than 0.8 may lead to greater risk, which is consistent with a piece of recent evidence in \citet{dahl2015variability}. Similarly, the identified 7.1 of arterial pH (below its normal range: 7.35-7.45) implies the potential presence of acidosis that leads to unfavorable outcomes for ICU patients. It is the recommended treatment threshold  of acute metabolic acidosis in severe sepsis and septic shock from the Survival Sepsis Campaign \citep{dellinger2013surviving}. Our scorecard also identifies several other risk factors that have been recognized by medical research, such as GCS (Glasgow Coma Scale) at 9, age at 80, and potassium at 4.25 \citep{solinger2013risks,martin2019risk}. Thus, the risk factors derived by our scorecard are consistent with those supported by medical research. This indicates that our prediction rules are sensible from the clinical perspective.
		
		An interesting observation is that the EO scorecard developed under our framework may potentially benefit clinical research. Besides factors that have been acknowledged, the proposed scorecard also identified new cut-off values for some important diagnostic indicators such as GCS at 5 and 7, bilirubin at 2.3 and 7.3, and pH.art at 7.2. This may help reveal possible and complicated interactions between these sepsis-related variables. Furthermore, the combination of rules developed under our framework is unique in the choice of indicators as well as cut-off values. Given that sepsis is a rather complex syndrome with unclear pathology and multiple comorbidities \citep{singer2016third}, a by-product of our research is that it reveals a holistic pattern of rules that are effective in medical diagnosis. We call for future examination of our research results by medical professionals. The detailed justification of the identified risk factors can be found in the supplementary material. 
		
		In addition, as shown in Figure \ref{fig:scorecard_Sepsis} and Figure \ref{figSM:scorecard_Sepsis} in the supplementary material, FASS7 scorecards have a model size not greater than $7$ for all fairness metrics. Therefore, the developed systems satisfy the additional constraint related to sparsity. This indicates that the proposed method could handle the application-specific constraints effectively. It provides decision-makers a high level of flexibility to customize their own requirements and develop an application-specific scoring system. 
		
		In conclusion, the proposed scoring systems can capture sensible risk-predictive rules. They help practitioners understand how the model works and how each input affects the final output. The transparency and interpretability facilitate the adoption of our proposed model in the real-life decision process. Moreover, sparsity and small integer coefficients in our models enable practitioners to make quick predictions by hand.  These are the advantages of our models for practical applications.

	\end{itemize}

	\subsection{Other Numerical Experiments}
	
	In this section, we conduct several numerical experiments to compare the performance of our scoring systems to other popular classification models in machine learning. We choose the contexts of income and credit prediction to showcase the generalizability of our framework beyond the healthcare industry.

	\subsubsection{Data Sets and Experimental Setup}\label{sec_CIbaseline}
	
	We conduct numerical experiments with two empirical data sets from the UCI Machine Learning Repository \citep{Dua2019}: the \textit{Adult} income data set and the \textit{German} credit data set. The original \textit{Adult} data set contains 48,842 observations with 14 features, including a binary class label which indicates whether an individual earns more than 50,000 dollars a year (coded as 1) or not (coded as 0). The original \textit{German} data set contains 1,000 observations with 20 features, including a binary class  label indicating whether a customer’s credit is good or not. We removed all data points with missing values, and processed each data set by binarizing all input features. Moreover, several sampling methods are used to create final balanced data set for model training. Detailed information regarding UCI data sets and data processing can be found in supplementary material.  %
	
	We consider gender (Male/Female) as the sensitive feature. In each experiment, we randomly partition the data into a training set ($70\%$) and test set ($30\%$) and repeat the partition randomly $5$ times to evaluate the average performance of models, unless otherwise stated. As a comparison, $6$ baseline scoring system and linear classifiers (Lasso, Ridge, Elastic Net, SVM, Huberized SVM, and SLIM) are also conducted in all the experiments, and the hyperparameters are selected via $5$-fold cross-validation. For the proposed fairness-aware scoring system, the coefficient set is chosen as $\mathcal{W}=\{-10, \ldots, 10\}^{d+1}$ and $a_{i}=b_{i}=1$ for all $i=1,\dots,n$ for simplicity. In this case, the data utility is reduced to the accuracy. {Furthermore, the cost-sensitive setting for model parameters is also considered, and the details are deferred to the supplementary material.} Besides, we set $\lambda_{0}<1/nd$ and $\epsilon=0.01$ so that our system will sacrifice little welfare for sparsity. The CPLEX 12.6.3 is employed to solve the final MIP problem.

	\subsubsection{Results} 
	
	\begin{itemize}
		
		\item {\textbf{Social Welfare Maximization}}
		
		We first verify the effectiveness of the proposed methods in achieving the optimal social welfare as defined in Section \ref{Sec:3}. In other words, the goal is to develop a scoring system maximizing the total social welfare that is defined as the sum of data utility and fairness utility over population. In this experiment, the value of average preference for fairness (i.e., $\bar{\rho}$) is set arbitrarily for simplicity.

		\begin{table} [!htbp]
			\caption{The average values of total social welfare for all methods on UCI data sets.}
			\label{ExResult_3}
			\scriptsize
			\centering
			\begin{threeparttable}
				\begin{tabular}{cccccccccc}
					\toprule
					\multirow{2}{*}{Dataset}&Fairness&\multirow{2}{*}{$\bar{\rho}$}& \multicolumn{6}{c}{Baselines}  & \multicolumn{1}{c}{Ours} \\
					\cmidrule(l){4-9} \cmidrule(l){10-10} 
					&Notions& &Ridge& Lasso&Elasticnet&SVM& Huberized SVM& SLIM& FASS\\
					\midrule
					
					\multirow{3}{*}{\textit{Adult}}	&SP  &0.2 & 0.7082&0.7070&0.7092&0.7030&0.6968&0.7020&\textbf{0.7468}\\
					&EO  &0.5 & 0.6698&0.6643&0.6682&0.6791&0.6479&0.6973&\textbf{0.7821}\\
					&OMR  &0.5 & 0.7499&0.7532&0.7521&0.7345&0.7412&0.7430&\textbf{0.7707}\\ 
					\midrule
					
					\multirow{3}{*}{\textit{German}}&SP  &0.2 & 0.7886&0.7938&0.7965&0.7872&0.7876&0.7801&\textbf{0.8089}\\
					&EO  &5& 0.5980&0.6004&0.6009&0.6061&0.5994&0.4537&\textbf{0.6659} \\
					&OMR  &5& 0.6389&0.6565&0.6318&0.6534&0.6350&0.6057&\textbf{0.7240}\\ 
					\bottomrule
				\end{tabular}
				\begin{tablenotes}
					\item Note: \textit{Statistical parity} is denoted by SP, \textit{equality of opportunity} by EO, and \textit{equal overall misclassification rate} by OMR. The optimal values are highted in bold. 
				\end{tablenotes}		
			\end{threeparttable}
		\end{table}

		Table \ref{ExResult_3} summarizes the results we obtained from applying baseline models and our approaches when incorporating different fairness measures. These results clearly show that the proposed method works well and yields the maximum total welfare in both data sets. Figure \ref{fig:2} further provides more insights regarding the data utility and fairness utility for welfare maximization on the \textit{Adult} data set. It can be seen from this figure that our framework attains competitive data utility compared to the baselines while gaining more fairness utility. Hence, our fairness-aware scoring system improves the overall social welfare significantly. {Due to the lack of space, we defer the graphs on \textit{German} data set and the results with the cost-sensitive setting to supplementary material, which also achieves similar trends.} Finally, the results from both data sets provide empirical validation for our theoretical analysis in Theorem \ref{Th1}.
		
		\begin{figure*}[!htbp]
			\centering
			\includegraphics[width=270pt]{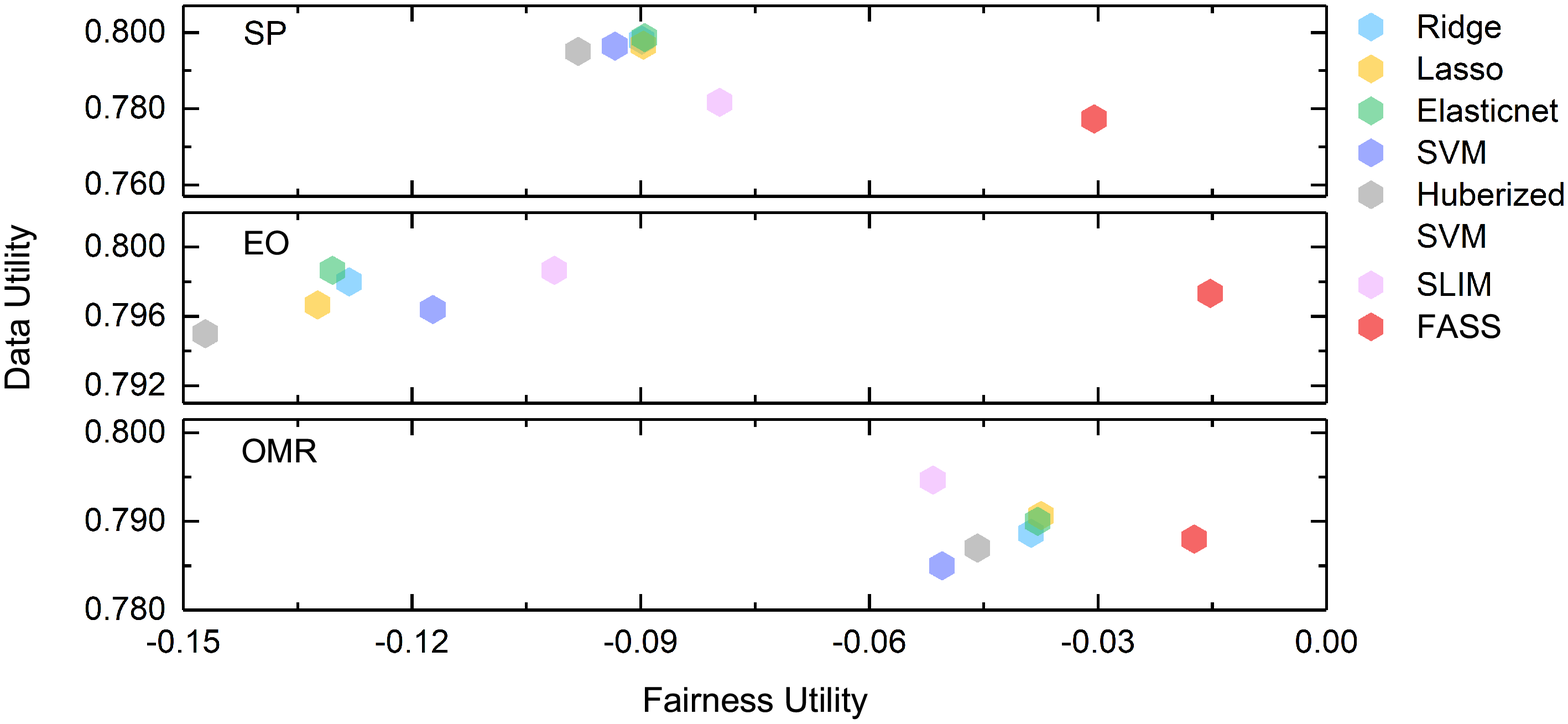}
			\caption{Data utility and fairness utility with different fairness measures on \textit{Adult} data set.}
			\label{fig:2}
		\end{figure*}

		\item {\textbf{Impact of Average Preference for Fairness}}

		In the previous study, we assume that the average value of fairness preference is set arbitrarily. In this experiment, we study the impact of the average preference for fairness on the performance of the proposed system. We start by varying the value of $\bar{\rho}$ associated with the fairness level. For each value of $\bar{\rho}$, we report the accuracy, which reflects data utility, and the (un)fairness level $\delta$ between two groups. The results are displayed in Figure \ref{fig:3}. It is apparent from this figure that both accuracy and (un)fairness level decrease as $\bar{\rho}$ increases in all fairness metrics. Note that $\bar{\rho}$ controls the trade-off between prediction efficiency and fairness. When $\bar{\rho}$ becomes larger, the scoring system produced by (\ref{framework1}) has the tendency to  sacrifice  more  classification accuracy to attain a lower unfairness level, since the latter will bring more benefits for the objective function. Thus, as $\bar{\rho}$ increases from $0 \to +\infty$, we transition from an unfair model with the best accuracy to a model with the best fairness level regardless of accuracy. The complete results on the \textit{German} data set, which show a tendency similar to the results in Figure \ref{fig:3}, are deferred to the supplementary material.

		\begin{figure*}[htbp]
			\centering
			\includegraphics[width=430pt]{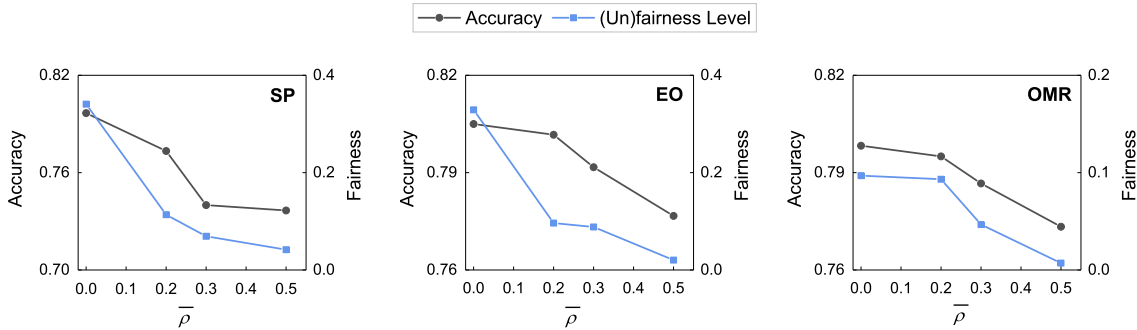}
			\caption{Trade-offs between accuracy and fairness with different fairness measures on \textit{Adult} data set in a randomly selected run.}
			\label{fig:3}
		\end{figure*}

	\end{itemize} 
	
	{{In addition, we also investigate the performance of the simplified models where the fairness level is specified in advance. Complete results are deferred to the supplementary material.}}
	
	
	\section{Conclusions}\label{sec:conc}
	Fairness has been a prevalent research topic that has garnered increasing interest in recent years \citep{Mccoy2014using,samorani2021overbooked}. In order to reduce systematic biases and improve transparency, operations management (OM) scholars have called for research work that integrates fairness into operational models  \citep{cohen2018big,rea2021unequal}.
	Our study answers this call by considering group fairness in the development of scoring systems, a type of predictive model that has been extensively adopted in a wide variety of industries. Past research attests that scores derived from many prevalent scoring systems are biased along with some important sensitive features such as gender, race, and socio-economic status \citep{Chouldechova2017fair,Obermeyer2019dissecting,campbell2020validation}. This has serious negative consequences for members of disadvantaged groups.
	
	{In this research, we propose a general framework for developing data-driven scoring systems that are fairness-centric. We first develop a social welfare optimization model that incorporates both fairness and efficiency components. Then, we cast the welfare maximization problem to the empirical risk minimization task in machine learning. By utilizing outcome fairness notions from machine learning, we develop metrics to quantify fairness so as to incorporate it into the objective function. Mixed integer programming techniques are utilized to derive a fairness-aware scoring system. In addition, several theoretical bounds are provided for model parameter selection.} Finally, experiments on empirical data sets verify the effectiveness of our approach.  
	
	
	
	
	Our research can be extended to many meaningful areas. First, our research question investigates algorithmic fairness notions in machine learning. Fairness notions in machine learning are largely inspired by the concepts of discrimination in social sciences and law \citep{grgic2016case}. As such, the key focus is the elimination of disparity in treatment, in impact, or in mistreatment. We take note that the meaning of fairness may differ in other research areas. For example, in economics and game theory, researchers also proposed several fairness notions based on the agent's preference, such as \textit{envy-freeness} \citep{Foley1967resource}. We call for future research to transfer these preference-based notions into classification tasks, formulate corresponding fairness metrics that can be recognized by the classification algorithm, and derive corresponding decision support systems.

	Second, we develop a fairness-aware scoring system with a fixed-size data set. Future research can take into consideration the speed of data accumulation, which presents a challenge to the constant maintenance and update of the existing scoring system. In this regard, we call for the exploration of methods such as online learning \citep{mcmahan2017survey} (i.e., a machine learning method in which data becomes available in a sequential order and is used to update the best predictor for future data \citep{hoi2014libol}) to meet the challenge of efficiently updating scoring systems.
	
	
	Lastly, in this study only a single sensitive feature is considered, but the proposed framework can be directly generalized to multiple sensitive features (e.g., gender and race). In order to avoid the ``fairness-gerrymandering" issue mentioned in \citet{kearns2018preventing}, when considering multiple sensitive features, one could construct all possible combinations of the sensitive feature values (e.g., black men, white women) and then add fairness constraints for each combination separately in the proposed framework. However, this may face considerable statistical and computational challenges, such as data scarcity at intersections of minorities and the rapidly growing number of subgroups. Therefore, future research could try to extend our method to the case of multiple sensitive features and develop mechanisms to overcome these hurdles.


	\bibliographystyle{apalike} 
	\bibliography{FairnessIP} 

\begin{thebibliography}{}

\bibitem[Alesina et~al., 2013]{Alesina2013women}
Alesina, A.~F., Lotti, F., and Mistrulli, P.~E. (2013).
\newblock {Do women pay more for credit? Evidence from Italy}.
\newblock {\em Journal of the European Economic Association}, 11(1):45--66.

\bibitem[Angus et~al., 2001]{angus2001epidemiology}
Angus, D.~C., Linde-Zwirble, W.~T., Lidicker, J., Clermont, G., Carcillo, J.,
  and Pinsky, M.~R. (2001).
\newblock {Epidemiology of severe sepsis in the United States: Analysis of
  incidence, outcome, and associated costs of care}.
\newblock {\em Critical Care Medicine}, 29(7):1303--1310.

\bibitem[Arabi et~al., 2003]{arabi2003assessment}
Arabi, Y., Al~Shirawi, N., Memish, Z., Venkatesh, S., and Al-Shimemeri, A.
  (2003).
\newblock Assessment of six mortality prediction models in patients admitted
  with severe sepsis and septic shock to the intensive care unit: A prospective
  cohort study.
\newblock {\em Critical Care}, 7(5):1--7.

\bibitem[Babcock et~al., 1996]{babcock1996choosing}
Babcock, L., Wang, X., and Loewenstein, G. (1996).
\newblock Choosing the wrong pond: Social comparisons in negotiations that
  reflect a self-serving bias.
\newblock {\em The Quarterly Journal of Economics}, 111(1):1--19.

\bibitem[Bandiera et~al., 2005]{Bandiera2005}
Bandiera, O., Barankay, I., and Rasul, I. ({2005}).
\newblock {Social preferences and the response to incentives: Evidence from
  personnel data}.
\newblock {\em {The Quarterly Journal of Economics}}, {120}({3}):{917--962}.

\bibitem[Barocas et~al., 2017]{Barocas2017fairnesstutorial}
Barocas, S., Hardt, M., and Narayanan, A. (2017).
\newblock Fairness in machine learning.
\newblock {\em Nips tutorial}, 1:2017.

\bibitem[Barocas and Selbst, 2016]{Barocas2016big}
Barocas, S. and Selbst, A.~D. (2016).
\newblock Big data's disparate impact.
\newblock {\em California Law Review}, 104:671--732.

\bibitem[Benjamin et~al., 2010]{Benjamin2010}
Benjamin, D.~J., Choi, J.~J., and Strickland, A.~J. (2010).
\newblock Social identity and preferences.
\newblock {\em American Economic Review}, 100(4):1913--28.

\bibitem[Berk et~al., 2021]{Berk2021fairness}
Berk, R., Heidari, H., Jabbari, S., Kearns, M., and Roth, A. (2021).
\newblock Fairness in criminal justice risk assessments: The state of the art.
\newblock {\em Sociological Methods \& Research}, 50(1):3--44.

\bibitem[Bierman, 2007]{Bierman2007}
Bierman, A.~S. (2007).
\newblock Sex matters: Gender disparities in quality and outcomes of care.
\newblock {\em Canadian Medical Association Journal}, 177(12):1520--1521.

\bibitem[Bjarnadottir et~al., 2018]{bjarnadottir2018predicting}
Bjarnadottir, M., Anderson, D., Zia, L., and Rhoads, K. (2018).
\newblock Predicting colorectal cancer mortality: Models to facilitate
  patient-physician conversations and inform operational decision making.
\newblock {\em Production and Operations Management}, 27(12):2162--2183.

\bibitem[Brayne, 2014]{Brayne2014surveillance}
Brayne, S. (2014).
\newblock Surveillance and system avoidance: Criminal justice contact and
  institutional attachment.
\newblock {\em American Sociological Review}, 79(3):367--391.

\bibitem[Breslow and Badawi, 2012]{breslow2012severity}
Breslow, M.~J. and Badawi, O. (2012).
\newblock Severity scoring in the critically iii: Part 2: Maximizing value from
  outcome prediction scoring systems.
\newblock {\em Chest}, 141(2):518--527.

\bibitem[Burgess, 1928]{Burgess1928factors}
Burgess, E.~W. (1928).
\newblock Factors determining success or failure on parole.
\newblock Technical report, Illinois Committee on Indeterminate-Sentence Law
  and Parole.

\bibitem[Campbell et~al., 2020]{campbell2020validation}
Campbell, C.~A., D’Amato, C., and Papp, J. (2020).
\newblock {Validation of the Ohio Youth Assessment System Dispositional Tool
  (OYAS-DIS): An examination of race and gender differences}.
\newblock {\em Youth Violence and Juvenile Justice}, 18(2):196--211.

\bibitem[Capon, 1982]{capon1982credit}
Capon, N. (1982).
\newblock Credit scoring systems: A critical analysis.
\newblock {\em Journal of Marketing}, 46(2):82--91.

\bibitem[Celis et~al., 2019]{celis2019classification}
Celis, L.~E., Huang, L., Keswani, V., and Vishnoi, N.~K. (2019).
\newblock Classification with fairness constraints: A meta-algorithm with
  provable guarantees.
\newblock In {\em Proceedings of the Conference on Fairness, Accountability,
  and Transparency}, page 319–328.

\bibitem[Charness and Rabin, 2002]{Charness2002}
Charness, G. and Rabin, M. (2002).
\newblock Understanding social preferences with simple tests.
\newblock {\em The Quarterly Journal of Economics}, 117(3):817--869.

\bibitem[Chen et~al., 2008]{Chen2008gender}
Chen, E.~H., Shofer, F.~S., Dean, A.~J., Hollander, J.~E., Baxt, W.~G., Robey,
  J.~L., Sease, K.~L., and Mills, A.~M. (2008).
\newblock Gender disparity in analgesic treatment of emergency department
  patients with acute abdominal pain.
\newblock {\em Academic Emergency Medicine}, 15(5):414--418.

\bibitem[Chen et~al., 2021]{chen2021finance}
Chen, T., Huang, Y., Lin, C., and Sheng, Z. (2021).
\newblock {Finance and Firm Volatility: Evidence from Small Business Lending in
  China}.
\newblock {\em Management Science}, page (forthcoming).

\bibitem[Chouldechova, 2017]{Chouldechova2017fair}
Chouldechova, A. (2017).
\newblock Fair prediction with disparate impact: A study of bias in recidivism
  prediction instruments.
\newblock {\em Big Data}, 5(2):153--163.

\bibitem[Coffman et~al., 2021]{coffman2021role}
Coffman, K.~B., Exley, C.~L., and Niederle, M. (2021).
\newblock The role of beliefs in driving gender discrimination.
\newblock {\em Management Science}, page (forthcoming).

\bibitem[Cohen, 2018]{cohen2018big}
Cohen, M.~C. (2018).
\newblock Big data and service operations.
\newblock {\em Production and Operations Management}, 27(9):1709--1723.

\bibitem[Corbett-Davies et~al., 2017]{Corbett2017}
Corbett-Davies, S., Pierson, E., Feller, A., Goel, S., and Huq, A. (2017).
\newblock Algorithmic decision making and the cost of fairness.
\newblock In {\em Proceedings of the 23rd ACM SIGKDD International Conference
  on Knowledge Discovery and Data Mining}, pages 797--806.

\bibitem[Cox and Sadiraj, 2012]{Cox2012Direct}
Cox, J.~C. and Sadiraj, V. (2012).
\newblock Direct tests of individual preferences for efficiency and equity.
\newblock {\em Economic Inquiry}, 50(4):920–931.

\bibitem[Dahl et~al., 2015]{dahl2015variability}
Dahl, R.~M., Gr{\o}nlykke, L., Haase, N., Holst, L.~B., Perner, A., Wetterslev,
  J., Rasmussen, B.~S., Meyhoff, C.~S., and { the 6S-Trial and TRISS Trial
  investigators} (2015).
\newblock Variability in targeted arterial oxygenation levels in patients with
  severe sepsis or septic shock.
\newblock {\em Acta Anaesthesiologica Scandinavica}, 59(7):859--869.

\bibitem[Dawes et~al., 2007]{Dawes2007}
Dawes, C.~T., Fowler, J.~H., Johnson, T., Mcelreath, R., and Smirnov, O.
  (2007).
\newblock Egalitarian motives in humans.
\newblock {\em Nature}, 446:794--796.

\bibitem[Dellinger et~al., 2013]{dellinger2013surviving}
Dellinger, R.~P., Levy, M.~M., Rhodes, A., Annane, D., Gerlach, H., Opal,
  S.~M., Sevransky, J.~E., Sprung, C.~L., Douglas, I.~S., Jaeschke, R., et~al.
  (2013).
\newblock {Surviving Sepsis Campaign: International guidelines for management
  of severe sepsis and septic shock, 2012}.
\newblock {\em Intensive Care Medicine}, 39(2):165--228.

\bibitem[Donini et~al., 2018]{Donini2018}
Donini, M., Oneto, L., Ben-David, S., Shawe-Taylor, J., and Pontil, M. (2018).
\newblock Empirical risk minimization under fairness constraints.
\newblock In {\em Proceedings of the 32nd International Conference on Neural
  Information Processing Systems}, page 2796–2806.

\bibitem[Dover et~al., 2015]{Dover2015does}
Dover, T.~L., Major, B., Kunstman, J.~W., and Sawyer, P.~J. (2015).
\newblock {Does unfairness feel different if it can be linked to group
  membership? Cognitive, affective, behavioral and physiological implications
  of discrimination and unfairness}.
\newblock {\em Journal of Experimental Social Psychology}, 56:96--103.

\bibitem[Dua and Graff, 2017]{Dua2019}
Dua, D. and Graff, C. (2017).
\newblock {UCI} machine learning repository.

\bibitem[Dumitrescu et~al., 2021]{dumitrescu2021machine}
Dumitrescu, E., Hue, S., Hurlin, C., and Tokpavi, S. (2021).
\newblock Machine learning for credit scoring: Improving logistic regression
  with non-linear decision-tree effects.
\newblock {\em European Journal of Operational Research}, page (forthcoming).

\bibitem[Dwork et~al., 2012]{Dwork2012}
Dwork, C., Hardt, M., Pitassi, T., Reingold, O., and Zemel, R. (2012).
\newblock Fairness through awareness.
\newblock In {\em Proceedings of the 3rd Innovations in Theoretical Computer
  Science Conference}, page 214–226.

\bibitem[Evans et~al., 2021]{evans2021surviving}
Evans, L., Rhodes, A., Alhazzani, W., Antonelli, M., Coopersmith, C.~M.,
  French, C., Machado, F.~R., Mcintyre, L., Ostermann, M., Prescott, H.~C.,
  et~al. (2021).
\newblock Surviving sepsis campaign: International guidelines for management of
  sepsis and septic shock 2021.
\newblock {\em Intensive Care Medicine}, 47(11):1181--1247.

\bibitem[Fehr and Schmidt, 1999]{Fehr1999A}
Fehr, E. and Schmidt, K.~M. (1999).
\newblock A theory of fairness, competition, and cooperation.
\newblock {\em The Quarterly Journal of Economics}, 114(3):817--868.

\bibitem[Foley, 1967]{Foley1967resource}
Foley, D.~K. (1967).
\newblock {\em Resource Allocation and the Public Sector}, volume~7.
\newblock Yale Economics Essays.

\bibitem[Friedman et~al., 2008]{friedman2008predictive}
Friedman, J.~H., Popescu, B.~E., et~al. (2008).
\newblock Predictive learning via rule ensembles.
\newblock {\em Annals of Applied Statistics}, 2(3):916--954.

\bibitem[Fu et~al., 2020]{fu2020artificial}
Fu, R., Huang, Y., and Singh, P.~V. (2020).
\newblock Artificial intelligence and algorithmic bias: Source, detection,
  mitigation, and implications.
\newblock In {\em Pushing the Boundaries: Frontiers in Impactful OR/OM
  Research}, pages 39--63. INFORMS.

\bibitem[Fu et~al., 2021]{Fu2021crowds}
Fu, R., Huang, Y., and Singh, P.~V. (2021).
\newblock Crowds, lending, machine, and bias.
\newblock {\em Information Systems Research}, 32(1):72--92.

\bibitem[Grgic-Hlaca et~al., 2016]{grgic2016case}
Grgic-Hlaca, N., Zafar, M.~B., Gummadi, K.~P., and Weller, A. (2016).
\newblock The case for process fairness in learning: Feature selection for fair
  decision making.
\newblock In {\em NIPS symposium on machine learning and the law}, volume~1,
  pages 2--11.

\bibitem[Hamman et~al., 2010]{hamman2010self}
Hamman, J.~R., Loewenstein, G., and Weber, R.~A. (2010).
\newblock Self-interest through delegation: An additional rationale for the
  principal-agent relationship.
\newblock {\em American Economic Review}, 100(4):1826--46.

\bibitem[Hardt et~al., 2016]{Hardt2016}
Hardt, M., Price, E., and Srebro, N. (2016).
\newblock Equality of opportunity in supervised learning.
\newblock In {\em Proceedings of the 30th International Conference on Neural
  Information Processing Systems}, page 3323–3331.

\bibitem[Harris et~al., 2019]{harris2019approximation}
Harris, D.~G., Li, S., Pensyl, T., Srinivasan, A., and Trinh, K. (2019).
\newblock Approximation algorithms for stochastic clustering.
\newblock {\em Journal of Machine Learning Research}, 20:1--33.

\bibitem[Henrich et~al., 2001]{henrich2001search}
Henrich, J., Boyd, R., Bowles, S., Camerer, C., Fehr, E., Gintis, H., and
  McElreath, R. (2001).
\newblock In search of homo economicus: Behavioral experiments in 15
  small-scale societies.
\newblock {\em American Economic Review}, 91(2):73--78.

\bibitem[Henry et~al., 2015]{henry2015targeted}
Henry, K.~E., Hager, D.~N., Pronovost, P.~J., and Saria, S. (2015).
\newblock {A targeted real-time early warning score (TREWScore) for septic
  shock}.
\newblock {\em Science Translational Medicine}, 7(299):299ra122.

\bibitem[Hoffman, 1994]{Hoffman1994twenty}
Hoffman, P.~B. (1994).
\newblock {Twenty years of operational use of a risk prediction instrument: The
  United States Parole Commission's Salient Factor Score}.
\newblock {\em Journal of Criminal Justice}, 22(6):477--494.

\bibitem[Hoi et~al., 2014]{hoi2014libol}
Hoi, S.~C., Wang, J., and Zhao, P. (2014).
\newblock Libol: A library for online learning algorithms.
\newblock {\em Journal of Machine Learning Research}, 15(1):495.

\bibitem[Hossain et~al., 2020]{Hossain2020designing}
Hossain, S., Mladenovic, A., and Shah, N. (2020).
\newblock Designing fairly fair classifiers via economic fairness notions.
\newblock In {\em Proceedings of The Web Conference 2020}, page 1559–1569.

\bibitem[Hu and Chen, 2018]{Hu2018short}
Hu, L. and Chen, Y. (2018).
\newblock A short-term intervention for long-term fairness in the labor market.
\newblock In {\em Proceedings of the 2018 World Wide Web Conference}, page
  1389–1398.

\bibitem[Hurley and Adebayo, 2016]{Hurley2017credit}
Hurley, M. and Adebayo, J. (2016).
\newblock Credit scoring in the era of big data.
\newblock {\em Yale Journal of Law and Technology}, 18:148--216.

\bibitem[Johnson et~al., 2016]{johnson2016mimic}
Johnson, A.~E., Pollard, T.~J., Shen, L., Li-Wei, H.~L., Feng, M., Ghassemi,
  M., Moody, B., Szolovits, P., Celi, L.~A., and Mark, R.~G. (2016).
\newblock {MIMIC-III, a freely accessible critical care database}.
\newblock {\em Scientific Data}, 3(1):1--9.

\bibitem[Kallus et~al., 2021]{Kallus2021assessing}
Kallus, N., Mao, X., and Zhou, A. (2021).
\newblock Assessing algorithmic fairness with unobserved protected class using
  data combination.
\newblock {\em Management Science}, page (forthcoming).

\bibitem[Karlan and Zinman, 2011]{Karlan2011Microcredit}
Karlan, D. and Zinman, J. (2011).
\newblock Microcredit in theory and practice: Using randomized credit scoring
  for impact evaluation.
\newblock {\em Science}, 332(6035):1278--84.

\bibitem[Kearns et~al., 2018]{kearns2018preventing}
Kearns, M., Neel, S., Roth, A., and Wu, Z.~S. (2018).
\newblock Preventing fairness gerrymandering: Auditing and learning for
  subgroup fairness.
\newblock In {\em Proceedings of the 35th International Conference on Machine
  Learning}, pages 2564--2572.

\bibitem[Kleinberg et~al., 2017]{kleinberg2017inherent}
Kleinberg, J., Mullainathan, S., and Raghavan, M. (2017).
\newblock Inherent trade-offs in the fair determination of risk scores.
\newblock In {\em The 8th Innovations in Theoretical Computer Science
  Conference}, volume~67, pages 43:1--43:23.

\bibitem[Knaus et~al., 1981]{Knaus1981APACHE}
Knaus, W.~A., Zimmerman, J.~E., Wagner, D.~P., Draper, E.~A., and Lawrence,
  D.~E. (1981).
\newblock {APACHE-acute physiology and chronic health evaluation: A
  physiologically based classification system}.
\newblock {\em Critical Care Medicine}, 9(8):591.

\bibitem[Konishi et~al., 2022]{konishi2022new}
Konishi, T., Goto, T., Fujiogi, M., Michihata, N., Kumazawa, R., Matsui, H.,
  Fushimi, K., Tanabe, M., Seto, Y., and Yasunaga, H. (2022).
\newblock New machine learning scoring system for predicting postoperative
  mortality in gastroduodenal ulcer perforation: a study using a japanese
  nationwide inpatient database.
\newblock {\em Surgery}, 171(4):1036--1042.

\bibitem[Kozodoi et~al., 2021]{Kozodoi2021fairness}
Kozodoi, N., Jacob, J., and Lessmann, S. (2021).
\newblock Fairness in credit scoring: Assessment, implementation and profit
  implications.
\newblock {\em European Journal of Operational Research}, page (forthcoming).

\bibitem[Lambrecht and Tucker, 2019]{lambrecht2019algorithmic}
Lambrecht, A. and Tucker, C. (2019).
\newblock {Algorithmic bias? An empirical study of apparent gender-based
  discrimination in the display of STEM career ads}.
\newblock {\em Management Science}, 65(7):2966--2981.

\bibitem[Le~Gall et~al., 1993]{LeGall1993}
Le~Gall, J.-R., Lemeshow, S., and Saulnier, F. (1993).
\newblock {A new simplified acute physiology score (SAPS II) based on a
  European/North American multicenter study}.
\newblock {\em JAMA}, 270(24):2957--2963.

\bibitem[Le~Gall et~al., 1984]{LeGall1984simplified}
Le~Gall, J.-R., Loirat, P., Alperovitch, A., Glaser, P., Granthil, C., Mathieu,
  D., Mercier, P., Thomas, R., and Villers, D. (1984).
\newblock {A simplified acute physiology score for ICU patients.}
\newblock {\em Critical Care Medicine}, 12(11):975--977.

\bibitem[Lohaus et~al., 2020]{Lohaus2020too}
Lohaus, M., Perrot, M., and Von~Luxburg, U. (2020).
\newblock Too relaxed to be fair.
\newblock In {\em International Conference on Machine Learning}, pages
  6360--6369.

\bibitem[Martin-Loeches et~al., 2019]{martin2019risk}
Martin-Loeches, I., Guia, M.~C., Vallecoccia, M.~S., Suarez, D., Ibarz, M.,
  Irazabal, M., Ferrer, R., and Artigas, A. (2019).
\newblock Risk factors for mortality in elderly and very elderly critically ill
  patients with sepsis: A prospective, observational, multicenter cohort study.
\newblock {\em Annals of Intensive Care}, 9(1):1--9.

\bibitem[McCoy and Lee, 2014]{Mccoy2014using}
McCoy, J.~H. and Lee, H.~L. (2014).
\newblock Using fairness models to improve equity in health delivery fleet
  management.
\newblock {\em Production and Operations Management}, 23(6):965--977.

\bibitem[McMahan, 2017]{mcmahan2017survey}
McMahan, H.~B. (2017).
\newblock A survey of algorithms and analysis for adaptive online learning.
\newblock {\em Journal of Machine Learning Research}, 18(1):3117--3166.

\bibitem[McNamara, 1972]{mcnamara1972present}
McNamara, C.~P. (1972).
\newblock The present status of the marketing concept.
\newblock {\em Journal of Marketing}, 36(1):50--57.

\bibitem[Mehrabi et~al., 2021]{Mehrabi2021survey}
Mehrabi, N., Morstatter, F., Saxena, N., Lerman, K., and Galstyan, A. (2021).
\newblock A survey on bias and fairness in machine learning.
\newblock {\em ACM Computing Surveys}, 54(6):1--35.

\bibitem[Miller, 1956]{Miller1956magical}
Miller, G.~A. (1956).
\newblock The magical number seven, plus or minus two: Some limits on our
  capacity for processing information.
\newblock {\em Psychological Review}, 63(2):81.

\bibitem[Molnar, 2020]{molnar2020interpretable}
Molnar, C. (2020).
\newblock {\em Interpretable machine learning}.
\newblock Lulu.com.

\bibitem[Moreno et~al., 2005]{Moreno2005saps}
Moreno, R.~P., Metnitz, P.~G., Almeida, E., Jordan, B., Bauer, P., Campos,
  R.~A., Iapichino, G., Edbrooke, D., Capuzzo, M., and Le~Gall, J.-R. (2005).
\newblock {SAPS 3-From evaluation of the patient to evaluation of the intensive
  care unit. Part 2: Development of a prognostic model for hospital mortality
  at ICU admission}.
\newblock {\em Intensive Care Medicine}, 31(10):1345--1355.

\bibitem[Mukherjee and Evans, 2017]{mukherjee2017implementation}
Mukherjee, V. and Evans, L. (2017).
\newblock Implementation of the surviving sepsis campaign guidelines.
\newblock {\em Current Opinion in Critical Care}, 23(5):412--416.

\bibitem[Ndirangu et~al., 2013]{ndirangu2013hiv}
Ndirangu, M., Sachs, S.~E., Palm, C., and Deckelbaum, R.~J. (2013).
\newblock {HIV affected households in Western Kenya experience greater food
  insecurity}.
\newblock {\em Food Policy}, 42:11--17.

\bibitem[Obermeyer et~al., 2019]{Obermeyer2019dissecting}
Obermeyer, Z., Powers, B., Vogeli, C., and Mullainathan, S. (2019).
\newblock Dissecting racial bias in an algorithm used to manage the health of
  populations.
\newblock {\em Science}, 366(6464):447--453.

\bibitem[Patel and Codner, 2016]{patel2016controversies}
Patel, J.~J. and Codner, P. (2016).
\newblock Controversies in critical care nutrition support.
\newblock {\em Critical Care Clinics}, 32(2):173--189.

\bibitem[Quadrianto and Sharmanska, 2017]{Quadrianto2017recycling}
Quadrianto, N. and Sharmanska, V. (2017).
\newblock Recycling privileged learning and distribution matching for fairness.
\newblock In {\em Advances in Neural Information Processing Systems},
  volume~30, pages 677--688.

\bibitem[Rajkomar et~al., 2018]{Rajkomar2018}
Rajkomar, A., Hardt, M., Howell, M.~D., Corrado, G., and Chin, M.~H. (2018).
\newblock Ensuring fairness in machine learning to advance health equity.
\newblock {\em Annals of Internal Medicine}, 169(12):866--872.

\bibitem[Rea et~al., 2021]{rea2021unequal}
Rea, D., Froehle, C., Masterson, S., Stettler, B., Fermann, G., and Pancioli,
  A. (2021).
\newblock Unequal but fair: Incorporating distributive justice in operational
  allocation models.
\newblock {\em Production and Operations Management}, page (forthcoming).

\bibitem[Reilly and Evans, 2006]{reilly2006translating}
Reilly, B.~M. and Evans, A.~T. (2006).
\newblock Translating clinical research into clinical practice: Impact of using
  prediction rules to make decisions.
\newblock {\em Annals of Internal Medicine}, 144(3):201--209.

\bibitem[Samorani et~al., 2021]{samorani2021overbooked}
Samorani, M., Harris, S.~L., Blount, L.~G., Lu, H., and Santoro, M.~A. (2021).
\newblock Overbooked and overlooked: Machine learning and racial bias in
  medical appointment scheduling.
\newblock {\em Manufacturing \& Service Operations Management}, page
  (forthcoming).

\bibitem[Santana et~al., 2013]{santana2013sao}
Santana, A.~R., de~Sousa, J.~L., Amorim, F.~F., Menezes, B.~M., Ara{\'u}jo, F.
  V.~B., Soares, F.~B., de~Carvalho~Santos, L.~C., de~Ara{\'u}jo, M. P.~B.,
  Rocha, P. H.~G., J{\'u}nior, P. N.~F., et~al. (2013).
\newblock {SaO2/FiO2 ratio as risk stratification for patients with sepsis}.
\newblock {\em Critical Care}, 17(4):1--59.

\bibitem[Singer et~al., 2016]{singer2016third}
Singer, M., Deutschman, C.~S., Seymour, C.~W., Shankar-Hari, M., Annane, D.,
  Bauer, M., Bellomo, R., Bernard, G.~R., Chiche, J.-D., Coopersmith, C.~M.,
  et~al. (2016).
\newblock {The third international consensus definitions for sepsis and septic
  shock (Sepsis-3)}.
\newblock {\em JAMA}, 315(8):801--810.

\bibitem[Skoufias et~al., 2020]{Skoufias2020estimating}
Skoufias, E., Diamond, A., Vinha, K., Gill, M., and Dellepiane, M.~R. (2020).
\newblock {Estimating poverty rates in subnational populations of interest: An
  assessment of the Simple Poverty Scorecard}.
\newblock {\em World Development}, 129:104887.

\bibitem[Solinger and Rothman, 2013]{solinger2013risks}
Solinger, A.~B. and Rothman, S.~I. (2013).
\newblock Risks of mortality associated with common laboratory tests: A novel,
  simple and meaningful way to set decision limits from data available in the
  electronic medical record.
\newblock {\em Clinical Chemistry and Laboratory Medicine}, 51(9):1803--1813.

\bibitem[Souder, 1972]{souder1972scoring}
Souder, W.~E. (1972).
\newblock A scoring methodology for assessing the suitability of management
  science models.
\newblock {\em Management Science}, 18(10):B526--B543.

\bibitem[Strand and Flaatten, 2008]{strand2008severity}
Strand, K. and Flaatten, H. (2008).
\newblock {Severity scoring in the ICU: a review}.
\newblock {\em Acta Anaesthesiologica Scandinavica}, 52(4):467--478.

\bibitem[Struck et~al., 2017]{struck2017association}
Struck, A.~F., Ustun, B., Ruiz, A.~R., Lee, J.~W., LaRoche, S.~M., Hirsch,
  L.~J., Gilmore, E.~J., Vlachy, J., Haider, H.~A., Rudin, C., et~al. (2017).
\newblock Association of an electroencephalography-based risk score with
  seizure probability in hospitalized patients.
\newblock {\em JAMA Neurology}, 74(12):1419--1424.

\bibitem[Sweeney et~al., 2018]{Sweeney2018A}
Sweeney, T.~E., Perumal, T.~M., Henao, R., Nichols, M., Howrylak, J.~A., Choi,
  A.~M., Bermejo-Martin, J.~F., Almansa, R., Tamayo, E., and Davenport, E.~E.
  (2018).
\newblock A community approach to mortality prediction in sepsis via gene
  expression analysis.
\newblock {\em Nature Communications}, 9(1):694.

\bibitem[Than et~al., 2014]{than2014development}
Than, M., Flaws, D., Sanders, S., Doust, J., Glasziou, P., Kline, J., Aldous,
  S., Troughton, R., Reid, C., Parsonage, W.~A., Frampton, C., Greenslade,
  J.~H., Deely, J.~M., Hess, E., Sadiq, A.~B., Singleton, R., Shopland, R.,
  Vercoe, L., Woolhouse-Williams, M., Ardagh, M., Bossuyt, P., Bannister, L.,
  and Cullen, L. (2014).
\newblock Development and validation of the emergency department assessment of
  chest pain score and 2h accelerated diagnostic protocol.
\newblock {\em Emergency Medicine Australasia}, 26(1):34--44.

\bibitem[Tricomi et~al., 2010]{Tricomi2010}
Tricomi, E., Rangel, A., Camerer, C.~F., and O'Doherty, J.~P. (2010).
\newblock Neural evidence for inequality-averse social preferences.
\newblock {\em Nature}, 463:1089--1091.

\bibitem[Ustun and Rudin, 2016]{UstunRu2016SLIM}
Ustun, B. and Rudin, C. (2016).
\newblock {Supersparse linear integer models for optimized medical scoring
  systems}.
\newblock {\em Machine Learning}, 102(3):349--391.

\bibitem[Vigdor, 2019]{Vigdor2019apple}
Vigdor, N. (2019).
\newblock Apple card investigated after gender discrimination complaints.
\newblock {\em The New York Times}.

\bibitem[Wang et~al., 2021]{Wang2021instrumental}
Wang, G., Li, J., and Hopp, W.~J. (2021).
\newblock An instrumental variable forest approach for detecting heterogeneous
  treatment effects in observational studies.
\newblock {\em Management Science}, page (forthcoming).

\bibitem[Wick et~al., 2019]{wick2019unlocking}
Wick, M., Panda, S., and Tristan, J.-B. (2019).
\newblock Unlocking fairness: A trade-off revisited.
\newblock In {\em Proceedings of the 33rd International Conference on Neural
  Information Processing Systems}, pages 8783--8792.

\bibitem[Wolsey, 1998]{Wolsey1998integer}
Wolsey, L.~A. (1998).
\newblock {\em Integer Programming}, volume~52.
\newblock John Wiley \& Sons.

\bibitem[Woodworth et~al., 2017]{Woodworth2017learning}
Woodworth, B., Gunasekar, S., Ohannessian, M.~I., and Srebro, N. (2017).
\newblock Learning non-discriminatory predictors.
\newblock In {\em Proceedings of the 2017 Conference on Learning Theory},
  volume~65, pages 1920--1953.

\bibitem[Wu et~al., 2021]{ying2021spesis}
Wu, Y., Huang, S., and Chang, X. (2021).
\newblock Understanding the complexity of sepsis mortality prediction via rule
  discovery and analysis: A pilot study.
\newblock {\em BMC Medical Informatics and Decision Making}, 21(1):1--15.

\bibitem[Yona and Rothblum, 2018]{yona2018probably}
Yona, G. and Rothblum, G. (2018).
\newblock Probably approximately metric-fair learning.
\newblock In {\em Proceedings of the 35th International Conference on Machine
  Learning}, volume~80, pages 5680--5688.

\bibitem[Zafar et~al., 2017a]{Zafar2017fairness}
Zafar, M.~B., Valera, I., Gomez~Rodriguez, M., and Gummadi, K.~P. (2017a).
\newblock Fairness beyond disparate treatment \& disparate impact: Learning
  classification without disparate mistreatment.
\newblock In {\em Proceedings of the 26th International Conference on World
  Wide Web}, page 1171–1180.

\bibitem[Zafar et~al., 2019]{Zafar2019fairness}
Zafar, M.~B., Valera, I., Gomez-Rodriguez, M., and Gummadi, K.~P. (2019).
\newblock Fairness constraints: A flexible approach for fair classification.
\newblock {\em Journal of Machine Learning Research}, 20(75):1--42.

\bibitem[Zafar et~al., 2017b]{Zafar2017fairness2}
Zafar, M.~B., Valera, I., Rogriguez, M.~G., and Gummadi, K.~P. (2017b).
\newblock Fairness constraints: Mechanisms for fair classification.
\newblock In {\em Proceedings of the 20th International Conference on
  Artificial Intelligence and Statistics}, volume~54, pages 962--970.

\bibitem[Zeng et~al., 2017]{zeng2017interpretable}
Zeng, J., Ustun, B., and Rudin, C. (2017).
\newblock Interpretable classification models for recidivism prediction.
\newblock {\em Journal of the Royal Statistical Society: Series A (Statistics
  in Society)}, 3(180):689--722.

\bibitem[Zhang et~al., 2019]{Zhang2019nips}
Zhang, X., Khalili, M.~M., Tekin, C., and Liu, M. (2019).
\newblock Group retention when using machine learning in sequential decision
  making: The interplay between user dynamics and fairness.
\newblock In {\em Proceedings of the 33rd International Conference on Neural
  Information Processing Systems}.

\bibitem[Zhang et~al., 2015]{zhang2015predicting}
Zhang, Y., Bradlow, E.~T., and Small, D.~S. (2015).
\newblock {Predicting customer value using clumpiness: From RFM to RFMC}.
\newblock {\em Marketing Science}, 34(2):195--208.

\bibitem[Zhao and Gordon, 2019]{Zhao2019inherent}
Zhao, H. and Gordon, G. (2019).
\newblock Inherent tradeoffs in learning fair representations.
\newblock In {\em Advances in Neural Information Processing Systems},
  volume~32, pages 15675--15685.

\bibitem[Zou and Schiebinger, 2018]{zou2018ai}
Zou, J. and Schiebinger, L. (2018).
\newblock {{AI can be sexist and racist—it’s time to make it fair}}.
\newblock {\em Nature}, 559:324–326.

\end{thebibliography}


\begin{thebibliography}{}
		\providecommand{\natexlab}[1]{#1}
		\providecommand{\url}[1]{\texttt{#1}}
		\providecommand{\urlprefix}{URL }
		
		\bibitem[{Antman et~al.(2000)Antman, Cohen, Bernink, McCabe, Horacek, Papuchis,
			Mautner, Corbalan, Radley, \protect\BIBand{} Braunwald}]{Antman20001}
		Antman EM, Cohen M, Bernink PJLM, McCabe CH, Horacek T, Papuchis G, Mautner B,
		Corbalan R, Radley D, Braunwald E (2000) The \text{TIMI Risk Score} for
		unstable angina/non–\text{ST} elevation \text{MI}: A method for
		prognostication and therapeutic decision making. \emph{JAMA} 284(7):835--842.
		
		\bibitem[{Benini et~al.(2003)Benini, Conley, Shdeed, Spurway, \protect\BIBand{}
			Yarmoshuk}]{benini2003integration1}
		Benini AA, Conley CE, Shdeed R, Spurway K, Yarmoshuk M (2003) Integration of
		different data bodies for humanitarian decision support: An example from mine
		action. \emph{Disasters} 27(4):288--304.
		
		\bibitem[{Bone et~al.(1992)Bone, Balk, Cerra, Dellinger, Fein, Knaus, Schein,
			\protect\BIBand{} Sibbald}]{bone1992definitions1}
		Bone RC, Balk RA, Cerra FB, Dellinger RP, Fein AM, Knaus WA, Schein RM, Sibbald
		WJ (1992) Definitions for sepsis and organ failure and guidelines for the use
		of innovative therapies in sepsis. \emph{Chest} 101(6):1644--1655.
		
		\bibitem[Brooks, 2011]{Brooks2011supporte}
		Brooks JP (2011) {Support vector machines with the ramp loss and the hard margin loss}.
		\emph{Operations Research}, 59(2):467--479.
		

		\bibitem[{Capon(1982)}]{capon1982credit1}
		Capon N (1982) Credit scoring systems: A critical analysis. \emph{Journal of
			Marketing} 46(2):82--91.
			
		\bibitem[{Chawla et~al.(2002)Chawla, Bowyer, Hall, \protect\BIBand{}
  Kegelmeyer}]{Chawla2002smote1}
Chawla NV, Bowyer KW, Hall LO, Kegelmeyer WP (2002) {SMOTE: Synthetic minority
  over-sampling technique}. \emph{Journal of Artificial Intelligence Research}
  16:321--357.
  
		\bibitem[{Chi \protect\BIBand{} Hsu(2012)}]{chi2012hybrid1}
		Chi BW, Hsu CC (2012) A hybrid approach to integrate genetic algorithm into
		dual scoring model in enhancing the performance of credit scoring model.
		\emph{Expert Systems with Applications} 39(3):2650--2661.
		
		\bibitem[{Dahl et~al.(2015)Dahl, Gr{\o}nlykke, Haase, Holst, Perner,
			Wetterslev, Rasmussen, Meyhoff, \protect\BIBand{} { the 6S-Trial and TRISS
				Trial investigators}}]{dahl2015variability1}
		Dahl RM, Gr{\o}nlykke L, Haase N, Holst LB, Perner A, Wetterslev J, Rasmussen
		BS, Meyhoff CS, { the 6S-Trial and TRISS Trial investigators} (2015)
		Variability in targeted arterial oxygenation levels in patients with severe
		sepsis or septic shock. \emph{Acta Anaesthesiologica Scandinavica}
		59(7):859--869.
		
		\bibitem[Donini et~al., 2018]{Donini2018e}
            Donini M, Oneto L, Ben-David S, Shawe-Taylor J, Pontil M (2018)
            {Empirical risk minimization under fairness constraints}.
        \emph{Proceedings of the 32nd International Conference on Neural Information Processing Systems}
        2796–2806.
		
		
		
		\bibitem[{Gogos et~al.(2003)Gogos, Lekkou, Papageorgiou, Siagris, Skoutelis,
			\protect\BIBand{} Bassaris}]{gogos2003clinical1}
		Gogos CA, Lekkou A, Papageorgiou O, Siagris D, Skoutelis A, Bassaris HP (2003)
		Clinical prognostic markers in patients with severe sepsis: A prospective
		analysis of 139 consecutive cases. \emph{Journal of Infection}
		47(4):300--306.
		
		\bibitem[{Harpviken et~al.(2003)Harpviken, Millard, Kjellman, \protect\BIBand{}
			Skara}]{harpviken2003measures1}
		Harpviken KB, Millard AS, Kjellman KE, Skara BA (2003) Measures for mines:
		Approaches to impact assessment in humanitarian mine action. \emph{Third
			World Quarterly} 24(5):889--908.
		
		\bibitem[{Hoffman(1983)}]{hoffman1983screening1}
		Hoffman PB (1983) Screening for risk: A revised salient factor score \text{(SFS
			81)}. \emph{Journal of Criminal Justice} 11(6):539--547.
		
		\bibitem[{Hoffman \protect\BIBand{} Beck(1997)}]{Hoffman1997CHS1}
		Hoffman PB, Beck JL (1997) The origin of the federal criminal history score.
		\emph{Federal Sentencing Reporter} 9(4):192--197.
		
		\bibitem[{Knaus et~al.(1985)Knaus, Draper, Wagner, \protect\BIBand{}
			Zimmerman}]{Knaus1985APACHE1}
		Knaus WA, Draper EA, Wagner DP, Zimmerman JE (1985) {APACHE II: A severity of
			disease classification system.} \emph{Critical Care Medicine}
		13(10):818--829.
		
		\bibitem[{Knaus et~al.(1991)Knaus, Wagner, Draper, Zimmerman, \protect\BIBand{}
			Damiano}]{Knaus1991The1}
		Knaus WA, Wagner DP, Draper EA, Zimmerman JE, Damiano AM (1991) {The APACHE III
			prognostic system: Risk prediction of hospital mortality for critically ill
			hospitalized adults.} \emph{Chest} 100(6):1619--1636.
		
		\bibitem[{Knaus et~al.(1981)Knaus, Zimmerman, Wagner, Draper, \protect\BIBand{}
			Lawrence}]{Knaus1981APACHE1}
		Knaus WA, Zimmerman JE, Wagner DP, Draper EA, Lawrence DE (1981) {APACHE-acute
			physiology and chronic health evaluation: A physiologically based
			classification system}. \emph{Critical Care Medicine} 9(8):591.
		
		\bibitem[{Kramer \protect\BIBand{} Scirica(1986)}]{Kramer1986complex1}
		Kramer JH, Scirica AJ (1986) {Complex policy choices: The Pennsylvania
			commission on sentencing}. \emph{Federal Probation} 50:15.
		
		\bibitem[{Kraut \protect\BIBand{} Madias(2010)}]{kraut2010metabolic1}
		Kraut JA, Madias NE (2010) Metabolic acidosis: Pathophysiology, diagnosis and
		management. \emph{Nature Reviews Nephrology} 6(5):274.
		
		\bibitem[{Kurowski et~al.(2016)Kurowski, Szarpak, Frass, Samarin,
			\protect\BIBand{} Czyzewski}]{Kurowski2016GCS1}
		Kurowski A, Szarpak {\L}, Frass M, Samarin S, Czyzewski {\L} (2016) \text{GCS}
		scale used as a prognostic factor in unconscious patients following cardiac
		arrest in prehospital situations: Preliminary data. \emph{American Journal of
			Emergency Medicine} 34(6):1178--1179.
		
		\bibitem[{Latessa et~al.(2009)Latessa, Smith, Lemke, Makarios,
			\protect\BIBand{} Lowenkamp}]{latessa2009creation1}
		Latessa E, Smith P, Lemke R, Makarios M, Lowenkamp C (2009) Creation and
		validation of the \text{Ohio} risk assessment system: Final report. Technical
		report, Center for Criminal Justice Research, School of Criminal Justice,
		University of Cincinnati, Cincinnati, OH,
		\urlprefix\url{http://www.ocjs.ohio.gov/ORAS\_FinalReport.pdf}.
		
		\bibitem[{Le~Gall et~al.(1996)Le~Gall, Klar, Lemeshow, Saulnier, Alberti,
			Artigas, \protect\BIBand{} Teres}]{le1996logistic1}
		Le~Gall JR, Klar J, Lemeshow S, Saulnier F, Alberti C, Artigas A, Teres D
		(1996) {The Logistic Organ Dysfunction System: A new way to assess organ
			dysfunction in the intensive care unit}. \emph{JAMA} 276(10):802--810.
		
		\bibitem[{Le~Gall et~al.(1993)Le~Gall, Lemeshow, \protect\BIBand{}
			Saulnier}]{LeGall19931}
		Le~Gall JR, Lemeshow S, Saulnier F (1993) {A new simplified acute physiology
			score (SAPS II) based on a European/North American multicenter study}.
		\emph{JAMA} 270(24):2957--2963.
		
		\bibitem[{Le~Gall et~al.(1984)Le~Gall, Loirat, Alperovitch, Glaser, Granthil,
			Mathieu, Mercier, Thomas, \protect\BIBand{} Villers}]{LeGall1984simplified1}
		Le~Gall JR, Loirat P, Alperovitch A, Glaser P, Granthil C, Mathieu D, Mercier
		P, Thomas R, Villers D (1984) {A simplified acute physiology score for ICU
			patients.} \emph{Critical Care Medicine} 12(11):975--977.
		
		\bibitem[{Lowder et~al.(2020)Lowder, Lawson, Grommon, \protect\BIBand{}
			Ray}]{lowder2020five1}
		Lowder EM, Lawson SG, Grommon E, Ray BR (2020) Five-county validation of the
		\text{Indiana Risk Assessment System--Pretrial Assessment Tool (IRAS-PAT)}
		using a local validation approach. \emph{Justice Quarterly} 37(7):1241--1260.
		
		\bibitem[{Marconi et~al.(2018)Marconi, Duncan, So-Armah, Re~3rd, Lim, Butt,
			Goetz, Rodriguez-Barradas, Alcorn, Lennox et~al.}]{marconi2018bilirubin1}
		Marconi VC, Duncan MS, So-Armah K, Re~3rd VL, Lim JK, Butt AA, Goetz MB,
		Rodriguez-Barradas MC, Alcorn CW, Lennox J, et~al. (2018) {Bilirubin is
			inversely associated with cardiovascular disease among HIV-positive and
			HIV-negative individuals in VACS (Veterans Aging Cohort Study)}.
		\emph{Journal of the American Heart Association} 7(10):e007792.
		
		\bibitem[{Martin-Loeches et~al.(2019)Martin-Loeches, Guia, Vallecoccia, Suarez,
			Ibarz, Irazabal, Ferrer, \protect\BIBand{} Artigas}]{martin2019risk1}
		Martin-Loeches I, Guia MC, Vallecoccia MS, Suarez D, Ibarz M, Irazabal M,
		Ferrer R, Artigas A (2019) Risk factors for mortality in elderly and very
		elderly critically ill patients with sepsis: A prospective, observational,
		multicenter cohort study. \emph{Annals of Intensive Care} 9(1):1--9.
		
		\bibitem[{Maxwell et~al.(2008)Maxwell, Caldwell, \protect\BIBand{}
			Langworthy}]{maxwell2008measuring1}
		Maxwell D, Caldwell R, Langworthy M (2008) Measuring food insecurity: Can an
		indicator based on localized coping behaviors be used to compare across
		contexts? \emph{Food Policy} 33(6):533--540.
		
		\bibitem[{McNamara(1972)}]{mcnamara1972present1}
		McNamara CP (1972) The present status of the marketing concept. \emph{Journal
			of Marketing} 36(1):50--57.
		
		\bibitem[{Moreno et~al.(2005)Moreno, Metnitz, Almeida, Jordan, Bauer, Campos,
			Iapichino, Edbrooke, Capuzzo, \protect\BIBand{} Le~Gall}]{Moreno2005saps1}
		Moreno RP, Metnitz PG, Almeida E, Jordan B, Bauer P, Campos RA, Iapichino G,
		Edbrooke D, Capuzzo M, Le~Gall JR (2005) {SAPS 3-From evaluation of the
			patient to evaluation of the intensive care unit. Part 2: Development of a
			prognostic model for hospital mortality at ICU admission}. \emph{Intensive
			Care Medicine} 31(10):1345--1355.
		
		\bibitem[{Ndirangu et~al.(2013)Ndirangu, Sachs, Palm, \protect\BIBand{}
			Deckelbaum}]{ndirangu2013hiv1}
		Ndirangu M, Sachs SE, Palm C, Deckelbaum RJ (2013) {HIV affected households in
			Western Kenya experience greater food insecurity}. \emph{Food Policy}
		42:11--17.
		
		\bibitem[{{Northpointe}(2015)}]{Northpointe2015practitioner1}
		{Northpointe} (2015) {Practitioner’s Guide to COMPAS Core}. Technical report,
		Northpointe Inc.,
		\urlprefix\url{http://www.northpointeinc.com/downloads/compas/Practitioners-Guide-COMPAS-Core-_031915.pdf}.
		
		\bibitem[{Sedlak \protect\BIBand{} Snyder(2004)}]{sedlak2004bilirubin1}
		Sedlak TW, Snyder SH (2004) Bilirubin benefits: Cellular protection by a
		biliverdin reductase antioxidant cycle. \emph{Pediatrics} 113(6):1776--1782.
		
		\bibitem[{Siddiqi(2012)}]{siddiqi2012credit1}
		Siddiqi N (2012) \emph{Credit Rsk Scorecards: Developing and Implementing
			Intelligent Credit Scoring}, volume~3 (John Wiley \& Sons).
		
		\bibitem[{Singer et~al.(2016)Singer, Deutschman, Seymour, Shankar-Hari, Annane,
			Bauer, Bellomo, Bernard, Chiche, Coopersmith et~al.}]{singer2016third1}
		Singer M, Deutschman CS, Seymour CW, Shankar-Hari M, Annane D, Bauer M, Bellomo
		R, Bernard GR, Chiche JD, Coopersmith CM, et~al. (2016) {The third
			international consensus definitions for sepsis and septic shock (Sepsis-3)}.
		\emph{JAMA} 315(8):801--810.
		
		\bibitem[{Six et~al.(2008)Six, Backus, \protect\BIBand{}
			Kelder}]{six2008chest1}
		Six A, Backus B, Kelder J (2008) {Chest pain in the emergency room: Value of
			the HEART score}. \emph{Netherlands Heart Journal} 16(6):191--196.
		
		\bibitem[{Skoufias et~al.(2020)Skoufias, Diamond, Vinha, Gill,
			\protect\BIBand{} Dellepiane}]{Skoufias2020estimating1}
		Skoufias E, Diamond A, Vinha K, Gill M, Dellepiane MR (2020) {Estimating
			poverty rates in subnational populations of interest: An assessment of the
			Simple Poverty Scorecard}. \emph{World Development} 129:104887.
		
		\bibitem[{Solinger \protect\BIBand{} Rothman(2013)}]{solinger2013risks1}
		Solinger AB, Rothman SI (2013) Risks of mortality associated with common
		laboratory tests: A novel, simple and meaningful way to set decision limits
		from data available in the electronic medical record. \emph{Clinical
			Chemistry and Laboratory Medicine} 51(9):1803--1813.
		
		\bibitem[{Stevenson(2018)}]{stevenson2018assessing1}
		Stevenson M (2018) Assessing risk assessment in action. \emph{Minnesota Law
			Review} 103:303.
		
		\bibitem[{Subbe et~al.(2001)Subbe, Kruger, Rutherford, \protect\BIBand{}
			Gemmel}]{Subbe20011}
		Subbe C, Kruger M, Rutherford P, Gemmel L (2001) {Validation of a modified
			Early Warning Score in medical admissions}. \emph{QJM: An International
			Journal of Medicine} 94(10):521--526.
		
		\bibitem[{Sveen et~al.(2016)Sveen, Bondjers, \protect\BIBand{}
			Willebrand}]{sveen2016psychometric1}
		Sveen J, Bondjers K, Willebrand M (2016) {Psychometric properties of the PTSD
			Checklist for DSM-5: A pilot study}. \emph{European Journal of
			Psychotraumatology} 7(1):30165.
		
		\bibitem[{Than et~al.(2014)Than, Flaws, Sanders, Doust, Glasziou, Kline,
			Aldous, Troughton, Reid, Parsonage, Frampton, Greenslade, Deely, Hess, Sadiq,
			Singleton, Shopland, Vercoe, Woolhouse-Williams, Ardagh, Bossuyt, Bannister,
			\protect\BIBand{} Cullen}]{than2014development1}
		Than M, Flaws D, Sanders S, Doust J, Glasziou P, Kline J, Aldous S, Troughton
		R, Reid C, Parsonage WA, Frampton C, Greenslade JH, Deely JM, Hess E, Sadiq
		AB, Singleton R, Shopland R, Vercoe L, Woolhouse-Williams M, Ardagh M,
		Bossuyt P, Bannister L, Cullen L (2014) Development and validation of the
		emergency department assessment of chest pain score and 2h accelerated
		diagnostic protocol. \emph{Emergency Medicine Australasia} 26(1):34--44.
		
		\bibitem[{Thomas et~al.(2017)Thomas, Crook, \protect\BIBand{}
        Edelman}]{Thomas2017credit1}
        Thomas L, Crook J, Edelman D (2017) \emph{Credit Scoring and Its Applications}
        (SIAM).
		
		\bibitem[{Ustun \protect\BIBand{} Rudin(2016)}]{UstunRu2016SLIM1}
		Ustun B, Rudin C (2016) {Supersparse linear integer models for optimized
			medical scoring systems}. \emph{Machine Learning} 102(3):349--391.
		
		\bibitem[{Vincent et~al.(1996)Vincent, Moreno, Takala, Willatts,
			De~Mendon{\c{c}}a, Bruining, Reinhart, Suter, \protect\BIBand{}
			Thijs}]{vincent1996sofa1}
		Vincent JL, Moreno R, Takala J, Willatts S, De~Mendon{\c{c}}a A, Bruining H,
		Reinhart C, Suter P, Thijs LG (1996) {The SOFA (Sepsis-related Organ Failure
			Assessment) score to describe organ dysfunction/failure}. \emph{Intensive
			Care Medicine} 22(7):707--710.
		
		\bibitem[{Wang et~al.(2020)Wang, He, \protect\BIBand{} Xu}]{wang2020serum1}
		Wang R, He M, Xu J (2020) Serum bilirubin level correlates with mortality in
		patients with traumatic brain injury. \emph{Medicine} 99(27):e21020.
		
		\bibitem[{Yang et~al.(2021)Yang, Huang, Huang, \protect\BIBand{}
			Chang}]{Yang2021}
		Yang Y, Huang S, Huang W, Chang X (2021) {Privacy-preserving cost-sensitive learning}. \emph{IEEE Transactions on Neural Networks and Learning Systems} 32(5):2105--2116.
		
		\bibitem[{Yeh et~al.(2009)Yeh, Yang, \protect\BIBand{}
			Ting}]{yeh2009knowledge1}
		Yeh IC, Yang KJ, Ting TM (2009) {Knowledge discovery on RFM model using
			Bernoulli sequence}. \emph{Expert Systems with Applications}
		36(3):5866--5871.
		
		
		\bibitem[Zafar et~al., 2019]{Zafar2019fairnesse}
		Zafar MB, Valera I, Gomez-Rodriguez M, Gummadi KP (2019) {Fairness constraints: A flexible approach for fair classification}.
		\emph{Journal of Machine Learning Research}, 20(75):1--42.
		
		\bibitem[{Zhang et~al.(2015)Zhang, Bradlow, \protect\BIBand{}
			Small}]{zhang2015predicting1}
		Zhang Y, Bradlow ET, Small DS (2015) {Predicting customer value using
			clumpiness: From RFM to RFMC}. \emph{Marketing Science} 34(2):195--208.
			


	
		
	\end{thebibliography}


\newpage
\appendix 
	\vspace{1cm}
\leftline{\LARGE\textbf{Supplementary Materials}}
	
	
	\section{Examples of the Wide Applications of Scoring Systems}
	
	Table \ref{paper_review} provides examples to show the wide applications of scoring systems in different areas.

	\begin{center}
		\begin{sidewaystable}[!htbp]
			\caption{Examples of the wide applications of scoring systems}
			\label{paper_review}
			\scriptsize
			\centering
			\begin{tabularx}{\textwidth}{clll}
				\toprule
				{\textbf{Areas}}&{\textbf{Papers}} & {\textbf{Scoring System Name}} & {\makecell[c]{\textbf{Application}}}\\
				\midrule
				\multirow{12}{*}{\textit{Healthcare} }& \citet{Knaus1981APACHE1}& APACHE I& \multirow{7}{*}{\makecell[l]{Assess the severity of disease and ICU mortality risk}}\\
				& \citet{Knaus1985APACHE1}& APACHE II & \\
				& \citet{Knaus1991The1}& APACHE III & \\
				& \citet{Subbe20011}&MEWS  & \\
				& \citet{LeGall1984simplified1}& SAPS I  & \\
				& \citet{LeGall19931}& SAPS II & \\
				& \citet{Moreno2005saps1}& SAPS III   & \\
				\cmidrule{2-4}
				& \citet{bone1992definitions1}& SIRS  &Detect system inflammatory response syndrome \\
				& \citet{Antman20001}& TIMI  & Assess the risk of death and ischemic events\\
				& \citet{six2008chest1}&HEART & Predict the outcome in chest pain patients in the emergency room\\
				& \citet{than2014development1} & EDACS & Assess the risk of a major adverse cardiac event for patients\\
				& \citet{sveen2016psychometric1}& PCL  & Screen for post-traumatic stress disorder\\
				\midrule
				\multirow{6}{*}{\textit{Criminal Justice}}& \citet{hoffman1983screening1}&Salient Factor Score  & Assess a federal prisoner's likelihood of recidivism after release\\
				& \citet{Kramer1986complex1}& Offense Gravity Score & Evaluate the seriousness of the offense\\
				& \citet{Hoffman1997CHS1}& Criminal History Score & Determine the applicable criminal history category for defendants\\
				& \citet{latessa2009creation1}& Ohio Risk Assessment System &Classify the risk level of offenders\\		 	
				& \citet{Northpointe2015practitioner1}& COMPAS &Assess the likelihood of a defendant becoming a recidivist \\
				\cmidrule{2-4}
				& \citet{stevenson2018assessing1}& Pretrial Risk Assessment Score& \multirow{2}{*}{Assess the risk of re-arrest and failure to appear after pretrial release} \\
				& \citet{lowder2020five1}& IRAS-PAT &  \\
				\midrule
				\multirow{3}{*}{\textit{Consumer Risk Analysis}} & \citet{capon1982credit1}  & \multirow{3}{*}{Scoring system without name} & \multirow{3}{*}{Assess the creditworthiness of applicants and the risk of default}\\	
				&\citet{siddiqi2012credit1}  &  &  \\	
				&\citet{chi2012hybrid1}  &  &  \\	
				\midrule
				\multirow{3}{*}{\textit{Marketing}} &\citet{yeh2009knowledge1}  &  \multirow{2}{*}{RFM Score}& \multirow{2}{*}{Measure the value of customers and predict their future behavior} \\	
				&\citet{zhang2015predicting1}  &  &  \\	
				\cmidrule{2-4}
				& \citet{mcnamara1972present1}  & Marketing Concept Score &  Quantify the extent to which the marketing concept is implemented\\	
				\midrule
				
				\multirow{5}{*}{\textit{Humanitarian Operations}} &\citet{harpviken2003measures1}   & \multirow{2}{*}{Mine Impact Score} & Assess the severity of socio-economic impacts of communities affected  \\
				& \citet{benini2003integration1}   &  & by landmines and/or unexploded ordnance \\
				\cmidrule{2-4}
				& \citet{maxwell2008measuring1}   & Coping Strategies Index& \multirow{2}{*}{Measure household food insecurity} \\
				& \citet{ndirangu2013hiv1}   &HFIAS Score&  \\
				\cmidrule{2-4}
				& \citet{Skoufias2020estimating1}   & Simple Poverty Scorecard  &  Monitor poverty rates and target services\\
				\bottomrule      
			\end{tabularx}
			
		\end{sidewaystable}
	\end{center}
	
\section{Development of Data-Driven Scoring Systems}

\begin{figure*}[htbp]
		\begin{center}
			\includegraphics[width=390pt]{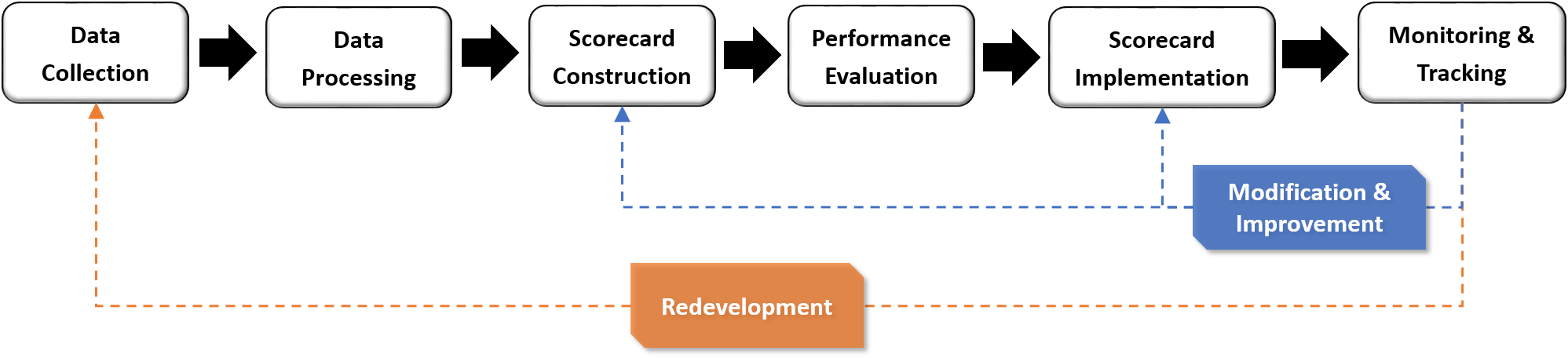}
			\caption{General steps in developing a data-driven scoring system.}
			\label{fig:flow_scoring}
		\end{center}
	\end{figure*}

A typical standard procedure to develop a data-driven scoring system is illustrated in the flowchart in Figure \ref{fig:flow_scoring}. There are commonly six steps. The first two are data construction steps, collecting and processing past data (e.g., data cleansing and feature engineering). The next two are scorecard development steps. The scorecard construction relies on the chosen algorithm and customized scenarios. It conducts model fitting with the training data set and scales the model into a scorecard. Then, the constructed scorecard will be evaluated on the test data set to provide an overview of its predicted performance. The last two are implementation and monitoring steps. Once the scorecard is validated, it is implemented in practice, and a monitoring and tracking procedure will be put in place to indicate the need for updates or redevelopments \citep{Thomas2017credit1}.
	
As indicated in Figure \ref{fig:flow_scoring}, data plays a vital role in the development of scoring systems and is the foundation of the whole process. The quality of the input data set primarily determines
the quality of system output. A scoring system built on biased data might produce biased predictions that lead to fairness issues.


	\section{Simplified Models}
	The proposed framework \eqref{framework_our} can further degenerate to several commonly used classification models in the machine learning field with some choices of model parameters. Here, we consider two types of simplified models.
	
	\subsection{Degeneration to Classification Model for Specified $\delta$}\label{degdelta}

	Now let us consider a case where the value of $\delta$ is pre-specified as $\delta^s$.	In this situation, the optimization problem (\ref{framework_our}) is equivalent to  
	\begin{align}\label{framework_deg2}
	\min _{\bs{w}} & \quad \frac{1}{N} \sum_{i=1}^{n}b_{i}\mathbbm{1}\left[y_{i} \bs{w}^{T} \bs{x}_{i} \leq 0\right]+\lambda_{0}\|\bs{w}\|_{0}+\epsilon\|\bs{w}\|_{1} \\\label{framework_deg2_fc}  \text{s.t.}& \quad g(\bs{w},\mathcal D)\leq \delta^{s},\\\nonumber
	&   \quad \bs{w}\in\mathcal{W}.
	\end{align}
	
	This will train a fairness-guaranteed  classifier following most of the existing algorithmic or in-processing approaches, which mainly aim at solving a constrained optimization problem by imposing a constraint on a given level of fairness while optimizing the accuracy \citep{Donini2018e,Zafar2019fairnesse}. Unlike the existing approaches, framework (\ref{framework_deg2}) directly optimizes the (weighted) error rate as well as the fairness by $0$-$1$ loss without the approximations that other methods utilize for scalability. As a result, it avoids sub-optimization and will normally achieve better classification performance while guaranteeing the given fairness level.

	\subsection{Degeneration to Classical Classification Model}

	Especially with a proper choice of the value of $\delta^{s}$ (e.g., $\delta^{s}=1$) such that the constraint (\ref{framework_deg2_fc}) ceases to bind, (\ref{framework_deg2}) could further degenerate to the ordinary classification model which only focuses on the (weighted) error rate (i.e., maximizing only the total data utility):
	\begin{align}\label{framework_deg1}
	\min _{\bs{w}} & \quad \frac{1}{N} \sum_{i=1}^{n}b_{i}\mathbbm{1}\left[y_{i} \bs{w}^{T} \bs{x}_{i} \leq 0\right]+\lambda_{0}\|\bs{w}\|_{0}+\epsilon\|\bs{w}\|_{1} \\\nonumber \text{s.t.}&  \quad \bs{w}\in\mathcal{W}.
	\end{align}
	
	Note that if the heterogeneity of data utility preference $b_{i}$ is further ignored, 
	then (\ref{framework_deg1}) degrades to the classic ERM framework as in \citet{UstunRu2016SLIM1}, which directly applies $0$-$1$ loss instead of convex surrogate functions. This will produce scoring systems that are robust to outliers and attain the learning-theoretic guarantee on predictive accuracy \citep{Brooks2011supporte}.
	

	\section{Omitted Proofs}
	\subsection{Proof of Theorem \ref{Th1}}

	\textbf{Proof:}
	Applying the Theorem 1 of \citet{UstunRu2016SLIM1}, it is easy to deduce that for a baseline classifier with real coefficients $\bs{\theta}$, there exists the coefficient set $\mathcal{W}$ (with a large enough value of $\Omega$) that contains a classifier with discrete coefficients $\bs{w}$ that assigns the exactly same label for any example $i$ as the baseline classifier with $\bs{\theta}$. I.e., for all $i\in\{1,2,\dots,n\}$, there exist $\bs{w}\in\mathcal{W}$ where $\mathcal{W}=\{-\Omega,\dots,\Omega\}$ with $\Omega>\dfrac{X_{1}\sqrt{d+1}}{2\eta_{(1)}}$, such that $\mathbbm{1}\left[y_{i} {\bs{w}}^{T} \bs{x}_{i} \leq 0\right]=\mathbbm{1}\left[y_{i} \bs{\rho}^{T} \bs{x}_{i} \leq 0\right]$. 
	
	{To further relate the resolution parameter $\Omega$ to the performance of scoring systems, we investigate the bounds on the welfare by considering progressively larger values of the margin. In other words, we develop the welfare bounds for scoring systems with the different choices of $\Omega$. 
	To this end, we apply the above results only to the reduced data set $D\backslash\mathcal{I}_{(k)}$}, and it follows that
	\begin{align}\nonumber
	&\sum_{i=1}^{n}b_{i}\mathbbm{1}\left[y_{i} \bs{w}^{T} \bs{x}_{i} \leq 0\right]-\sum_{i=1}^{n}b_{i}\mathbbm{1}\left[y_{i} \bs{\theta}^{T} \bs{x}_{i} \leq 0\right]\\\nonumber=&\sum_{i\in\mathcal{I}_{(k)}}b_{i}\left\{\mathbbm{1}\left[y_{i} \bs{w}^{T} \bs{x}_{i} \leq 0\right]-\mathbbm{1}\left[y_{i} \bs{\theta}^{T} \bs{x}_{i} \leq 0\right]\right\}+\sum_{i\notin\mathcal{I}_{(k)}}b_{i}\left\{\mathbbm{1}\left[y_{i} \bs{w}^{T} \bs{x}_{i} \leq 0\right]-\mathbbm{1}\left[y_{i} \bs{\theta}^{T} \bs{x}_{i} \leq 0\right]\right\}\\\label{eq1}=&\sum_{i\in\mathcal{I}_{(k)}}b_{i}\left\{\mathbbm{1}\left[y_{i} \bs{w}^{T} \bs{x}_{i} \leq 0\right]-\mathbbm{1}\left[y_{i} \bs{\theta}^{T} \bs{x}_{i} \leq 0\right]\right\}\\\label{ieq1}\leq&\sum_{i\in\mathcal{I}_{(k)}}b_{i}\\\label{ieq2}\leq&(k-1)\max_{i\in\mathcal{I}_{(k)}}b_{i}.
	\end{align}
	The equation (\ref{eq1}) is due to the fact that the classifier with $\bs{w}$ assigns the exact same label as the classifier with $\bs{\theta}$ for any example $i\in D\backslash\mathcal{I}_{(k)}$. The inequality in (\ref{ieq1}) results from the most extreme case where all examples in $\mathcal{I}_{(k)}$ are misclassified by $\bs{w}$ but correctly classified by $\bs{\theta}$.
	The inequality in (\ref{ieq2}) follows from the fact that there exist $k-1$ elements in $\mathcal{I}_{(k)}$.
	
	Next, we discuss the fairness level of the discrete linear classifier $\bs{w}$. Note that for convenience, we abuse notation somewhat and use $\psi_{i}(\bs{w})=\mathbbm{1}\left[y_{i} \bs{w}^{T} \bs{x}_{i} \leq 0\right]$ as in Section \ref{subsec_3.1} to indicate whether an example $i$ is misclassified or not by the classifier with coefficients $\bs{w}$. Now, we consider the following three fairness definitions mentioned in Section \ref{sec_faircons}. 
	
	\textbf{I. Equal Overall Misclassification Rate}
		
		We first consider the case where the \textit{equal overall misclassification rate} is used to measure fairness level. Note that $c$ is the number of different groups, and since the classifier with $\bs{\theta}$ satisfies a given fairness requirement $G(\cdot)$ and $\delta$, we have
		\begin{align}
		G_{OMR}(\bs{\theta})=\left| \frac{1}{N_{a_{p}}} \sum_{i \in I_{a_{p}}} \psi_{i}(\bs{\theta})-\frac{1}{N_{a_{q}}} \sum_{i \in I_{a_{q}}} \psi_{i}(\bs{\theta})\right|\leq\delta
		\end{align}
		for any $p, q=1, \ldots, c.$
		
		Recall that the classifier with $\bs{w}$ assigns the exact same label as the classifier with $\bs{\theta}$ for any example $i\in D\backslash\mathcal{I}_{(k)}$. Thus, $\psi_{i}(\bs{\theta})=\psi_{i}(\bs{w})$ for all $i\in D\backslash\mathcal{I}_{(k)}$. Besides, we denote $z_{p}=\sum_{i \in \mathcal{I}_{(k)}\cap I_{a_{p}}} \big(\psi_{i}(\bs{w})-\psi_{i}(\bs{\theta})\big)$ for $p=1,2,\dots,c$. Summing $|z_{p}|$ over $p$, it follows that
		\begin{align}\nonumber
		\sum_{p=1}^{c}\left|z_{p}\right|&=\sum_{p=1}^{c}\left|\sum_{i \in \mathcal{I}_{(k)}\cap I_{a_{p}}} \psi_{i}(\bs{w})-\psi_{i}(\bs{\theta})\right|\\\label{ieq3}&\leq\sum_{p=1}^{c}\sum_{i \in \mathcal{I}_{(k)}\cap I_{a_{p}}}\Big| \psi_{i}(\bs{w})-\psi_{i}(\bs{\theta})\Big|\\\label{ieq5}&\leq \left|\mathcal{I}_{(k)}\right|=k-1.
		\end{align}
		
		Since $\sum_{i \in \mathcal{I}_{(k)}\cap I_{a_{p}}} \psi_{i}(\bs{w})=z_{p}+\sum_{i \in \mathcal{I}_{(k)}\cap I_{a_{p}}} \psi_{i}(\bs{\theta})$ , we have
		\begin{align}\nonumber
		G_{OMR}(\bs{w})&=\left| \frac{1}{N_{a_{p}}} \sum_{i \in I_{a_{p}}} \psi_{i}(\bs{w})-\frac{1}{N_{a_{q}}} \sum_{i \in I_{a_{q}}} \psi_{i}(\bs{w})\right|\\\nonumber&=\left| \frac{1}{N_{a_{p}}} \left(\sum_{i \in \mathcal{I}_{(k)}\cap I_{a_{p}}} \psi_{i}(\bs{w})+ \sum_{i \in I_{a_{p}}-\mathcal{I}_{(k)}} \psi_{i}(\bs{w})\right)-\frac{1}{N_{a_{q}}} \left(\sum_{i \in \mathcal{I}_{(k)}\cap I_{a_{q}}} \psi_{i}(\bs{w})+\sum_{i \in I_{a_{q}}-\mathcal{I}_{(k)}} \psi_{i}(\bs{w})\right)\right|\\\label{ieq4}&=\left| \frac{1}{N_{a_{p}}} \left(\sum_{i \in \mathcal{I}_{(k)}\cap I_{a_{p}}} \psi_{i}(\bs{\theta})+z_{p}+ \sum_{i \in I_{a_{p}}-\mathcal{I}_{(k)}} \psi_{i}(\bs{\theta})\right)-\frac{1}{N_{a_{q}}} \left(\sum_{i \in \mathcal{I}_{(k)}\cap I_{a_{q}}} \psi_{i}(\bs{\theta})+z_{q}+\sum_{i \in I_{a_{q}}-\mathcal{I}_{(k)}} \psi_{i}(\bs{\theta})\right)\right|\\\nonumber&=\left| \frac{1}{N_{a_{p}}} \left(\sum_{i \in I_{a_{p}}} \psi_{i}(\bs{\theta})+z_{p}\right)-\frac{1}{N_{a_{q}}} \left(\sum_{i \in I_{a_{q}}} \psi_{i}(\bs{\theta})+z_{q}\right)\right|\\\nonumber&=\left| \frac{1}{N_{a_{p}}} \sum_{i \in I_{a_{p}}} \psi_{i}(\bs{\theta})-\frac{1}{N_{a_{q}}} \sum_{i \in I_{a_{q}}} \psi_{i}(\bs{\theta})+\frac{z_{p}}{N_{a_{p}}} -\frac{z_{q}}{N_{a_{q}}}\right|\\\nonumber&\leq\left| \frac{1}{N_{a_{p}}} \sum_{i \in I_{a_{p}}} \psi_{i}(\bs{\theta})-\frac{1}{N_{a_{q}}} \sum_{i \in I_{a_{q}}} \psi_{i}(\bs{\theta})\right|+\left|\frac{z_{p}}{N_{a_{p}}} -\frac{z_{q}}{N_{a_{q}}}\right|\\\nonumber&\leq\delta+\left|\frac{z_{p}}{N_{a_{p}}} -\frac{z_{q}}{N_{a_{q}}}\right|\\\nonumber&\leq\delta+(k-1)\max\left\{\dfrac{1}{N_{a_{p}}},\dfrac{1}{N_{a_{q}}}\right\},
		\end{align}
		for any $p, q=1, \ldots, c.$
		The equation (\ref{ieq4}) follows from the fact that $\psi_{i}(\bs{\theta})=\psi_{i}(\bs{w})$ for all $i\in D_{N}\backslash\mathcal{I}_{(k)}$, and the last inequality is due to the inequality (\ref{ieq5}).
		Hence, the maximal increment of the tolerance level of unfairness among all groups is $\Delta_{F}(k)=(k-1)\max\left\{\dfrac{1}{N_{a_{1}}},\dfrac{1}{N_{a_{2}}},\dots,\dfrac{1}{N_{a_{c}}}\right\}$ for $\bs{w}$.
		
		\textbf{II. Equality of Opportunity}
		
		Recall that if the \textit{equality of opportunity} is used, we have 
		$$G_{EO}(\bs{\theta})=\left| \frac{1}{N_{a_{p}}^{+}} \sum_{i \in I^{+}_{a_{p}}} \psi_{i}(\bs{\theta})-\frac{1}{N_{a_{q}}^{+}} \sum_{i \in I^{+}_{a_{q}}} \psi_{i}(\bs{\theta})\right|\leq\delta$$
		for any $p, q=1, \ldots, c.$
		We denote $z_{p}^{+}=\sum_{i \in \mathcal{I}_{(k)}\cap I^{+}_{a_{p}}} \big(\psi_{i}(\bs{w})-\psi_{i}(\bs{\theta})\big)$ for $p=1,2,\dots,c$. Similar in Case I, it follows that
		\begin{align}\nonumber
		\sum_{p=1}^{c}\left|z^{+}_{p}\right|&=\sum_{p=1}^{c}\left|\sum_{i \in \mathcal{I}_{(k)}\cap I^{+}_{a_{p}}} \psi_{i}(\bs{w})-\psi_{i}(\bs{\theta})\right|\\\nonumber&\leq\sum_{p=1}^{c}\sum_{i \in \mathcal{I}_{(k)}\cap I^{+}_{a_{p}}}\Big| \psi_{i}(\bs{w})-\psi_{i}(\bs{\theta})\Big|\\\nonumber&\leq\sum_{p=1}^{c}\sum_{i \in \mathcal{I}_{(k)}\cap I_{a_{p}}}\Big| \psi_{i}(\bs{w})-\psi_{i}(\bs{\theta})\Big|\\\nonumber&\leq \left|\mathcal{I}_{(k)}\right|=k-1.
		\end{align}
		Afterwards, for any two groups $p, q=1, \ldots, c$, we have
		\begin{align}\nonumber
		G_{EO}(\bs{w})&=\left| \frac{1}{N^{+}_{a_{p}}} \sum_{i \in I^{+}_{a_{p}}} \psi_{i}(\bs{w})-\frac{1}{N^{+}_{a_{q}}} \sum_{i \in I^{+}_{a_{q}}} \psi_{i}(\bs{w})\right|\\\nonumber&=\left| \frac{1}{N^{+}_{a_{p}}} \left(\sum_{i \in \mathcal{I}_{(k)}\cap I^{+}_{a_{p}}} \psi_{i}(\bs{w})+ \sum_{i \in I^{+}_{a_{p}}-\mathcal{I}_{(k)}} \psi_{i}(\bs{w})\right)-\frac{1}{N^{+}_{a_{q}}} \left(\sum_{i \in \mathcal{I}_{(k)}\cap I^{+}_{a_{q}}} \psi_{i}(\bs{w})+\sum_{i \in I^{+}_{a_{q}}-\mathcal{I}_{(k)}} \psi_{i}(\bs{w})\right)\right|\\\nonumber&=\left| \frac{1}{N^{+}_{a_{p}}} \left(\sum_{i \in \mathcal{I}_{(k)}\cap I^{+}_{a_{p}}} \psi_{i}(\bs{\theta})+z^{+}_{p}+ \sum_{i \in I^{+}_{a_{p}}-\mathcal{I}_{(k)}} \psi_{i}(\bs{\theta})\right)-\frac{1}{N^{+}_{a_{q}}} \left(\sum_{i \in \mathcal{I}_{(k)}\cap I^{+}_{a_{q}}} \psi_{i}(\bs{\theta})+z^{+}_{q}+\sum_{i \in I^{+}_{a_{q}}-\mathcal{I}_{(k)}} \psi_{i}(\bs{\theta})\right)\right|\\\nonumber&=\left| \frac{1}{N^{+}_{a_{p}}} \left(\sum_{i \in I^{+}_{a_{p}}} \psi_{i}(\bs{\theta})+z^{+}_{p}\right)-\frac{1}{N^{+}_{a_{q}}} \left(\sum_{i \in I^{+}_{a_{q}}} \psi_{i}(\bs{\theta})+z^{+}_{q}\right)\right|\\\nonumber&\leq\left| \frac{1}{N^{+}_{a_{p}}} \sum_{i \in I^{+}_{a_{p}}} \psi_{i}(\bs{\theta})-\frac{1}{N^{+}_{a_{q}}} \sum_{i \in I^{+}_{a_{q}}} \psi_{i}(\bs{\theta})\right|+\left|\frac{z^{+}_{p}}{N^{+}_{a_{p}}} -\frac{z^{+}_{q}}{N^{+}_{a_{q}}}\right|\\\nonumber&\leq\delta+\left|\frac{z^{+}_{p}}{N^{+}_{a_{p}}} -\frac{z^{+}_{q}}{N^{+}_{a_{q}}}\right|\\\nonumber&\leq\delta+(k-1)\max\left\{\dfrac{1}{N^{+}_{a_{p}}},\dfrac{1}{N^{+}_{a_{q}}}\right\}.
		\end{align}
		
		Hence, for $\bs{w}$ the maximal increment of the tolerance level of unfairness among all groups is $$\Delta_{F}(k)=(k-1)\max\left\{\dfrac{1}{N^{+}_{a_{1}}},\dfrac{1}{N^{+}_{a_{2}}},\dots,\dfrac{1}{N^{+}_{a_{c}}}\right\}.$$

	\textbf{III. Statistical Parity}
		
		Now, we consider a relatively complex case where the given fairness notion is \textit{statistical parity}. Recall that for any two groups $p, q=1, \ldots, c$, it follows that
		$$G_{SP}(\bs{\theta})=\left|\left(\frac{N_{a_{p}}^{+}}{N_{a_{p}}}-\frac{N_{a_{q}}^{+}}{N_{a_{q}}}\right)+\frac{1}{N_{a_{p}}}\left[\sum_{i \in I_{a_{p}}^{-}} \psi_{i}(\bs{\theta})-\sum_{i \in I_{a_{p}}^{+} } \psi_{i}(\bs{\theta})\right]-\frac{1}{N_{a_{q}}}\left[\sum_{i \in I_{a_{q}}^{-}} \psi_{i}(\bs{\theta})-\sum_{i \in I_{a_{q}}^{+}} \psi_{i}(\bs{\theta})\right]\right|\leq\delta.$$
		
		Again, we denote $z_{p}^{+}=\sum_{i \in \mathcal{I}_{(k)}\cap I^{+}_{a_{p}}} \big(\psi_{i}(\bs{w})-\psi_{i}(\bs{\theta})\big)$ and $z_{p}^{-}=\sum_{i \in \mathcal{I}_{(k)}\cap I^{-}_{a_{p}}} \big(\psi_{i}(\bs{w})-\psi_{i}(\bs{\theta})\big)$ for $p=1,2,\dots,c$. Then, it is easy to deduce that 
		\begin{align}\nonumber
		\sum_{p=1}^{c}\left|z^{+}_{p}\right|+\left|z^{-}_{p}\right|&=\sum_{p=1}^{c}\left(\left|\sum_{i \in \mathcal{I}_{(k)}\cap I^{+}_{a_{p}}} \big(\psi_{i}(\bs{w})-\psi_{i}(\bs{\theta})\big)\right|+\left|\sum_{i \in \mathcal{I}_{(k)}\cap I^{-}_{a_{p}}} \big(\psi_{i}(\bs{w})-\psi_{i}(\bs{\theta})\big)\right|\right)\\\nonumber&\leq\sum_{p=1}^{c}\left(\sum_{i \in \mathcal{I}_{(k)}\cap I^{+}_{a_{p}}}\left|\psi_{i}(\bs{w})-\psi_{i}(\bs{\theta})\right|+\sum_{i \in \mathcal{I}_{(k)}\cap I^{-}_{a_{p}}}\left|\psi_{i}(\bs{w})-\psi_{i}(\bs{\theta})\right|\right)\\\nonumber&\leq\sum_{p=1}^{c}\left(\sum_{i \in \mathcal{I}_{(k)}\cap I^{+}_{a_{p}}}1+\sum_{i \in \mathcal{I}_{(k)}\cap I^{-}_{a_{p}}}1\right)\\\nonumber&=\sum_{p=1}^{c}\sum_{i \in \mathcal{I}_{(k)}\cap I_{a_{p}}}1\\\label{ieq6}&=\left|\mathcal{I}_{(k)}\right|=k-1.
		\end{align}
		
		Similar to the above two cases, the value of $G_{SP}(\bs{w})$ is upper bounded
		as
		\begin{align}\nonumber
		G_{SP}(\bs{w})&=\left|\left(\frac{N_{a_{p}}^{+}}{N_{a_{p}}}-\frac{N_{a_{q}}^{+}}{N_{a_{q}}}\right)+\frac{1}{N_{a_{p}}}\left[\sum_{i \in I_{a_{p}}^{-}} \psi_{i}(\bs{w})-\sum_{i \in I_{a_{p}}^{+} } \psi_{i}(\bs{w})\right]-\frac{1}{N_{a_{q}}}\left[\sum_{i \in I_{a_{q}}^{-}} \psi_{i}(\bs{w})-\sum_{i \in I_{a_{q}}^{+}} \psi_{i}(\bs{w})\right]\right|\\\nonumber&=\left|\left(\frac{N_{a_{p}}^{+}}{N_{a_{p}}}-\frac{N_{a_{q}}^{+}}{N_{a_{q}}}\right)+\frac{1}{N_{a_{p}}}\left[\sum_{i \in I_{a_{p}}^{-}} \psi_{i}(\bs{\theta})+z^{-}_{p}-\sum_{i \in I_{a_{p}}^{+} } \psi_{i}(\bs{\theta})-z^{+}_{p}\right]\right.\\\nonumber&\qquad\left.-\frac{1}{N_{a_{q}}}\left[\sum_{i \in I_{a_{q}}^{-}} \psi_{i}(\bs{\theta})+z^{-}_{q}-\sum_{i \in I_{a_{q}}^{+}} \psi_{i}(\bs{w})-z^{+}_{q}\right]\right|\\\nonumber&=\left|\left(\frac{N_{a_{p}}^{+}}{N_{a_{p}}}-\frac{N_{a_{q}}^{+}}{N_{a_{q}}}\right)+\frac{1}{N_{a_{p}}}\left[\sum_{i \in I_{a_{p}}^{-}} \psi_{i}(\bs{\theta})-\sum_{i \in I_{a_{p}}^{+} } \psi_{i}(\bs{\theta})\right]-\frac{1}{N_{a_{q}}}\left[\sum_{i \in I_{a_{q}}^{-}} \psi_{i}(\bs{\theta})-\sum_{i \in I_{a_{q}}^{+}} \psi_{i}(\bs{\theta})\right]\right.\\\nonumber&\qquad\left.+\left(\dfrac{z^{-}_{p}-z^{+}_{p}}{N_{a_{p}}}-\dfrac{z^{-}_{q}-z^{+}_{q}}{N_{a_{q}}}\right)\right|\\\nonumber&\leq\delta+\left|\dfrac{z^{-}_{p}-z^{+}_{p}}{N_{a_{p}}}-\dfrac{z^{-}_{q}-z^{+}_{q}}{N_{a_{q}}}\right|\\\nonumber&\leq\delta+(k-1)\max\left\{\dfrac{1}{N_{a_{p}}},\dfrac{1}{N_{a_{q}}}\right\}
		\end{align}
		for any $p, q=1, \ldots, c.$ The last inequality follows from the inequality (\ref{ieq6}).
		Therefore, the maximal increment of the tolerance level of unfairness among all groups is $\Delta_{F}(k)=(k-1)\max\left\{\dfrac{1}{N_{a_{1}}},\dfrac{1}{N_{a_{2}}},\dots,\dfrac{1}{N_{a_{c}}}\right\}$  for $\bs{w}$.

	With $\Delta_{F}(k)$ in hand, we can then calculate the bound of social welfare difference between two classifiers. For the classifier with $\bs{w}$ and the classifier with $\bs{\theta}$, we have
	\begin{align}\nonumber
	SWF(\bs{w})&=\sum_{i=1}^{n}a_{i}-\sum_{i=1}^{n}b_{i}\mathbbm{1}\left[y_{i} \bs{w}^{T} \bs{x}_{i} \leq 0\right]-\delta_{\bs{w}}\sum_{i=1}^{n}\rho_{i}\\\nonumber SWF(\bs{\theta})&=\sum_{i=1}^{n}a_{i}-\sum_{i=1}^{n}b_{i}\mathbbm{1}\left[y_{i} \bs{\theta}^{T} \bs{x}_{i} \leq 0\right]-\delta_{\bs{\theta}}\sum_{i=1}^{n}\rho_{i},
	\end{align}
	where $\delta_{\bs{w}}$ (resp. $\delta_{\bs{\theta}}$) is the tolerance level of unfairness that the classifier with $\bs{w}$  (resp. $\bs{\theta}$) can achieve.  Note that $\delta_{\bs{\theta}}=\delta$ according to the assumption. Then, we have 
	\begin{align}\nonumber
	SWF(\bs{w})- SWF(\bs{\theta})&=-\left[\sum_{i=1}^{n}b_{i}\mathbbm{1}\left[y_{i} \bs{w}^{T} \bs{x}_{i} \leq 0\right]-\sum_{i=1}^{n}b_{i}\mathbbm{1}\left[y_{i} \bs{\theta}^{T} \bs{x}_{i} \leq 0\right]\right]-N\bar{\rho}\left(\delta_{\bs{w}}-\delta_{\bs{\theta}}\right)\\\nonumber&\geq(1-k)\max_{i\in\mathcal{I}_{(k)}}b_{i}-N\bar{\rho}(\delta+\Delta_{F}(k)-\delta)\\\label{ieq7}&=(1-k)\max_{i\in\mathcal{I}_{(k)}}b_{i}-N\bar{\rho}\Delta_{F}(k).
	\end{align}
	Recall that according to the definition of $\bs{w}^{*}$, it is the discrete classifier maximizing the social welfare. Based on the inequality (\ref{ieq7}), we have
	\begin{align}\nonumber
	SWF(\bs{w}^{*})- SWF(\bs{\theta})&\geq SWF(\bs{w})- SWF(\bs{\theta})\\
	&=(1-k)\max_{i\in\mathcal{I}_{(k)}}b_{i}-N\bar{\rho}\Delta_{F}(k).
	\end{align}

	\subsection{Proof of Theorem \ref{Th2}}
	
	\textbf{Proof:}
	Let $\mathcal{V}(\bs{w})=\sum_{i=1}^{n}v_{i}$ be the overall data utility of $\bs{w}$, and $\bs{w}^{data}=\argmax_{\bs{w}\in\mathcal{W}} \mathcal{V}(\bs{w})$ denote the classifier that only focuses on the data utility. We first prove
	
	\begin{align}\label{Th2_ieq}
	\mathcal{V}(\bs{w}^{data})\geq\max\left\{\Delta^{*}, \mathcal{V}(\bs{w}^{*})\right\},
	\end{align}
	where $\Delta^{*}$ is a parameter related to $\delta^{*}$ and the corresponding fairness definition. According to the definition of $\bs{w}^{data}$, we have $\mathcal{V}(\bs{w}^{data})\geq\mathcal{V}(\bs{w}^{*})$ directly.

\textbf{I. Equal Overall Misclassification Rate}
		
		First, we consider a classifier $\bs{w}_{uf}\in\mathcal{W}$ that does not satisfy the \textit{equal overall misclassification rate} requirement with $\delta^{*}$ for all groups. In other words,
		\begin{align}\label{Th2_pf1}
		G_{OMR}(\bs{w}_{uf})=\left| \frac{1}{N_{a_{p}}} \sum_{i \in I_{a_{p}}} \psi_{i}(\bs{w}_{uf})-\frac{1}{N_{a_{q}}} \sum_{i \in I_{a_{q}}} \psi_{i}(\bs{w}_{uf})\right|>\delta^{*}
		\end{align}
		for any $p, q=1, \ldots, c$ and $p\neq q$.	
		From the above equation, it follows that
		\begin{align}\label{Th2_pf2}
		\frac{1}{N_{a_{p}}} \sum_{i \in I_{a_{p}}} \psi_{i}(\bs{w}_{uf})-\frac{1}{N_{a_{q}}} \sum_{i \in I_{a_{q}}} \psi_{i}(\bs{w}_{uf})<-\delta^{*}
		\end{align}
		or
		\begin{align}\label{Th2_pf3}
		\frac{1}{N_{a_{q}}} \sum_{i \in I_{a_{q}}} \psi_{i}(\bs{w}_{uf})-\frac{1}{N_{a_{p}}} \sum_{i \in I_{a_{p}}} \psi_{i}(\bs{w}_{uf})<-\delta^{*}.
		\end{align}
		
		Multiplying (\ref{Th2_pf2}) by $N_{a_{p}}$, we can get
		\begin{align}\label{Th2_pf4}
		\sum_{i \in I_{a_{p}}} \psi_{i}(\bs{w}_{uf})-\frac{N_{a_{p}}}{N_{a_{q}}} \sum_{i \in I_{a_{q}}} \psi_{i}(\bs{w}_{uf})<-N_{a_{p}}\delta^{*}.
		\end{align}
		Thus, 
		\begin{align}\nonumber
		\sum_{i \in I_{a_{p}}} \psi_{i}(\bs{w}_{uf})+\sum_{i \in I_{a_{q}}} \psi_{i}(\bs{w}_{uf})&=\sum_{i \in I_{a_{p}}} \psi_{i}(\bs{w}_{uf})-\frac{N_{a_{p}}}{N_{a_{q}}} \sum_{i \in I_{a_{q}}} \psi_{i}(\bs{w}_{uf})+\frac{N_{a_{p}}+N_{a_{q}}}{N_{a_{q}}} \sum_{i \in I_{a_{q}}} \psi_{i}(\bs{w}_{uf})\\\label{Th2_pf5}&<N_{a_{p}}(1-\delta^{*})+N_{a_{q}}.
		\end{align}
		The inequality in (\ref{Th2_pf5}) results from (\ref{Th2_pf4}) and the fact that $\frac{1}{N_{a_{q}}} \sum_{i \in I_{a_{q}}} \psi_{i}(\bs{w}_{uf})\leq 1$. Then we multiply (\ref{Th2_pf3}) by $N_{a_{q}}$. Using the similar method, we have
		\begin{align}\nonumber
		\sum_{i \in I_{a_{p}}} \psi_{i}(\bs{w}_{uf})+\sum_{i \in I_{a_{q}}} \psi_{i}(\bs{w}_{uf})<N_{a_{q}}(1-\delta^{*})+N_{a_{p}}.
		\end{align}
		Combining this with (\ref{Th2_pf5}) gives $$\sum_{i \in I_{a_{p}}} \psi_{i}(\bs{w}_{uf})+\sum_{i \in I_{a_{q}}} \psi_{i}(\bs{w}_{uf})<\max\left\{N_{a_{p}}(1-\delta^{*})+N_{a_{q}},N_{a_{q}}(1-\delta^{*})+N_{a_{p}}\right\}$$
		for any $p, q=1, \ldots, c$ and $p\neq q$.	Summarizing this inequality over $c\choose 2$ unique pairs of $p, q$, it is easy to deduce that
		\begin{align}\nonumber
		\sum_{i=1}^{n}\psi_{i}(\bs{w}_{uf})=\sum_{p=1}^{c}\sum_{i \in I_{a_{p}}} \psi_{i}(\bs{w}_{uf})<\dfrac{1}{c-1}{\sum_{p\neq q}}\max\left\{N_{a_{p}}(1-\delta^{*})+N_{a_{q}},N_{a_{q}}(1-\delta^{*})+N_{a_{p}}\right\}.
		\end{align}
		
		Hence, the lower bound of $\mathcal{V}(\bs{w}_{uf})$ is given by
		\begin{align}\nonumber
		\mathcal{V}(\bs{w}_{uf})&=\sum_{i=1}^{n}a_{i}-\sum_{i=1}^{n}b_{i}\psi_{i}(\bs{w}_{uf})\\\nonumber&\geq\sum_{i=1}^{n}a_{i}-\max_{i}b_{i}\sum_{i=1}^{n}\psi_{i}(\bs{w}_{uf})\\\nonumber&>\sum_{i=1}^{n}a_{i}-\dfrac{\max_{i}b_{i}}{c-1}{\sum_{p\neq q}}\max\left\{N_{a_{p}}(1-\delta^{*})+N_{a_{q}},N_{a_{q}}(1-\delta^{*})+N_{a_{p}}\right\}.
		\end{align}
		Define $\Delta^{*}=\sum_{i=1}^{n}a_{i}-\dfrac{\max_{i}b_{i}}{c-1}{\sum_{p\neq q}}\max\left\{N_{a_{p}}(1-\delta^{*})+N_{a_{q}},N_{a_{q}}(1-\delta^{*})+N_{a_{p}}\right\}$. Since $\bs{w}^{data}=\argmax_{\bs{w}\in\mathcal{W}}\mathcal{V}(\bs{w})$ by definition, it follows that $\mathcal{V}(\bs{w}^{data})\geq \mathcal{V}(\bs{w}_{uf})>\Delta^{*}$.
		
\textbf{II. Equality of Opportunity}
		
		Similar to Case I, we first consider a classifier $\bs{w}_{uf}$ which does not satisfy the \textit{equality of opportunity} requirement with $\delta^{*}$ for all groups. That is,
		$$G_{EO}(\bs{w}_{uf})=\left| \frac{1}{N_{a_{p}}^{+}} \sum_{i \in I^{+}_{a_{p}}} \psi_{i}(\bs{w}_{uf})-\frac{1}{N_{a_{q}}^{+}} \sum_{i \in I^{+}_{a_{q}}} \psi_{i}(\bs{w}_{uf})\right|>\delta^{*}$$
		for any two groups $p, q=1, \ldots, c$ and $p\neq q$. Thus, for $\bs{w}_{uf}$, we have
		\begin{align}\label{Th2_c2_pf1}
		\frac{1}{N_{a_{p}}^{+}} \sum_{i \in I^{+}_{a_{p}}} \psi_{i}(\bs{w}_{uf})-\frac{1}{N_{a_{q}}^{+}} \sum_{i \in I^{+}_{a_{q}}} \psi_{i}(\bs{w}_{uf})<-\delta^{*}
		\end{align}
		or 
		\begin{align}\label{Th2_c2_pf2}
		\frac{1}{N_{a_{q}}^{+}} \sum_{i \in I^{+}_{a_{q}}} \psi_{i}(\bs{w}_{uf})-\frac{1}{N_{a_{p}}^{+}} \sum_{i \in I^{+}_{a_{p}}} \psi_{i}(\bs{w}_{uf})<-\delta^{*}.
		\end{align}
		Multiplying (\ref{Th2_c2_pf1}) by $N_{a_{p}}^{+}$ results in 
		\begin{align}\nonumber
		\sum_{i \in I^{+}_{a_{p}}} \psi_{i}(\bs{w}_{uf})-\frac{N_{a_{p}}^{+}}{N_{a_{q}}^{+}} \sum_{i \in I^{+}_{a_{q}}} \psi_{i}(\bs{w}_{uf})<-N_{a_{p}}^{+}\delta^{*}.
		\end{align}
		Then, we have 
		\begin{align}\nonumber
		\sum_{i \in I^{+}_{a_{p}}}\psi_{i}(\bs{w}_{uf})+\sum_{i \in I^{+}_{a_{q}}} \psi_{i}(\bs{w}_{uf})&=\sum_{i \in I^{+}_{a_{p}}}\psi_{i}(\bs{w}_{uf})-\frac{N_{a_{p}}^{+}}{N_{a_{q}}^{+}} \sum_{i \in I^{+}_{a_{q}}} \psi_{i}(\bs{w}_{uf})+\frac{N_{a_{p}}^{+}+N_{a_{q}}^{+}}{N_{a_{q}}^{+}} \sum_{i \in I_{a_{q}}^{+}} \psi_{i}(\bs{w}_{uf})\\\label{Th2_c2_pf3}&<N_{a_{p}}^{+}(1-\delta^{*})+N_{a_{q}}^{+}.
		\end{align}
		From the inequality (\ref{Th2_c2_pf2}), using similar methods leads to
		\begin{align}\label{Th2_c2_pf4}
		\sum_{i \in I^{+}_{a_{p}}}\psi_{i}(\bs{w}_{uf})+\sum_{i \in I^{+}_{a_{q}}} \psi_{i}(\bs{w}_{uf})<N_{a_{q}}^{+}(1-\delta^{*})+N_{a_{p}}^{+}.
		\end{align}
		Combining (\ref{Th2_c2_pf3}) and ((\ref{Th2_c2_pf4}), we obtain
		$$\sum_{i \in I^{+}_{a_{p}}}\psi_{i}(\bs{w}_{uf})+\sum_{i \in I^{+}_{a_{q}}} \psi_{i}(\bs{w}_{uf})<\max\left\{N_{a_{p}}^{+}(1-\delta^{*})+N_{a_{q}}^{+}, N_{a_{q}}^{+}(1-\delta^{*})+N_{a_{p}}^{+}\right\}$$
		for any $p, q=1, \ldots, c$ and $p\neq q$. Summarizing this inequality over $c\choose 2$ unique pairs of $p, q$, we can easily derive that
		\begin{align}\nonumber
		\sum_{i \in I^{+}}\psi_{i}(\bs{w}_{uf})<\dfrac{1}{c-1}{\sum_{p\neq q}}\max\left\{N_{a_{p}}^{+}(1-\delta^{*})+N_{a_{q}}^{+}, N_{a_{q}}^{+}(1-\delta^{*})+N_{a_{p}}^{+}\right\},
		\end{align}
		when $I^{+}=\big\{i\in\{1,2,\dots,n\}\left| y_{i}=1\right.\big\}$ is the set of individuals from positive class. Hence, 
		\begin{align}\nonumber
		\sum_{i=1}^{n}\psi_{i}(\bs{w}_{uf})&=\sum_{i \in I^{+}}\psi_{i}(\bs{w}_{uf})+\sum_{i \in I^{-}}\psi_{i}(\bs{w}_{uf})\\\nonumber&<\dfrac{1}{c-1}{\sum_{p\neq q}}\max\left\{N_{a_{p}}^{+}(1-\delta^{*})+N_{a_{q}}^{+}, N_{a_{q}}^{+}(1-\delta^{*})+N_{a_{p}}^{+}\right\}+N^{-},
		\end{align}
		when $I^{-}=\big\{i\in\{1,2,\dots,n\}\left| y_{i}=-1\right.\big\}$ is the set of individuals from negative class and $N^{-}=|I^{-}|$ is the size of $I^{-}$.  Then, the lower bound of $\mathcal{V}(\bs{w}_{uf})$ is 
		\begin{align}\nonumber
		\mathcal{V}(\bs{w}_{uf})&=\sum_{i=1}^{n}a_{i}-\sum_{i=1}^{n}b_{i}\psi_{i}(\bs{w}_{uf})\\\nonumber&\geq\sum_{i=1}^{n}a_{i}-\max_{i}b_{i}\sum_{i=1}^{n}\psi_{i}(\bs{w}_{uf})\\\nonumber&>\sum_{i=1}^{n}a_{i}-\dfrac{\max_{i}b_{i}}{c-1}{\sum_{p\neq q}}\max\left\{N_{a_{p}}(1-\delta^{*})+N_{a_{q}},N_{a_{q}}(1-\delta^{*})+N_{a_{p}}\right\}-\max_{i}b_{i}N^{-}.
		\end{align}
		Here, we define $\Delta^{*}=\sum_{i=1}^{n}a_{i}-\dfrac{\max_{i}b_{i}}{c-1}{\sum_{p\neq q}}\max\left\{N_{a_{p}}(1-\delta^{*})+N_{a_{q}},N_{a_{q}}(1-\delta^{*})+N_{a_{p}}\right\}-\max_{i}b_{i}N^{-}$. Recalling the definition of $\bs{w}^{data}$, it directly follows that $\mathcal{V}(\bs{w}^{data})\geq \mathcal{V}(\bs{w}_{uf})>\Delta^{*}$.
		
\textbf{III. Statistical Parity}
		
		When \textit{statistical parity} is used for the fairness requirement, we first consider a classifier $\bs{w}_{uf}$ which does not satisfy this fairness requirement with $\delta^{*}$ for all groups. This leads to
		\begin{align}\nonumber
		G_{SP}(\bs{w}_{uf})&=\left|\left(\frac{N_{a_{p}}^{+}}{N_{a_{p}}}-\frac{N_{a_{q}}^{+}}{N_{a_{q}}}\right)+\frac{1}{N_{a_{p}}}\left[\sum_{i \in I_{a_{p}}^{-}} \psi_{i}(\bs{w}_{uf})-\sum_{i \in I_{a_{p}}^{+} } \psi_{i}(\bs{w}_{uf})\right]-\frac{1}{N_{a_{q}}}\left[\sum_{i \in I_{a_{q}}^{-}} \psi_{i}(\bs{w}_{uf})-\sum_{i \in I_{a_{q}}^{+}} \psi_{i}(\bs{w}_{uf})\right]\right|\\\nonumber&>\delta^{*}
		\end{align}
		for any two groups $p, q=1, \ldots, c$ and $p\neq q$. For $\bs{w}_{uf}$, this implies that
		\begin{align}\label{Th2_c3_pf1}
		\left(\frac{N_{a_{p}}^{+}}{N_{a_{p}}}-\frac{N_{a_{q}}^{+}}{N_{a_{q}}}\right)+\frac{1}{N_{a_{p}}}\left[\sum_{i \in I_{a_{p}}^{-}} \psi_{i}(\bs{w}_{uf})-\sum_{i \in I_{a_{p}}^{+} } \psi_{i}(\bs{w}_{uf})\right]-\frac{1}{N_{a_{q}}}\left[\sum_{i \in I_{a_{q}}^{-}} \psi_{i}(\bs{w}_{uf})-\sum_{i \in I_{a_{q}}^{+}} \psi_{i}(\bs{w}_{uf})\right]<-\delta^{*}
		\end{align}
		or
		\begin{align}\label{Th2_c3_pf2}
		\left(\frac{N_{a_{q}}^{+}}{N_{a_{q}}}-\frac{N_{a_{p}}^{+}}{N_{a_{p}}}\right)+\frac{1}{N_{a_{q}}}\left[\sum_{i \in I_{a_{q}}^{-}} \psi_{i}(\bs{w}_{uf})-\sum_{i \in I_{a_{q}}^{+}} \psi_{i}(\bs{w}_{uf})\right]-\frac{1}{N_{a_{p}}}\left[\sum_{i \in I_{a_{p}}^{-}} \psi_{i}(\bs{w}_{uf})-\sum_{i \in I_{a_{p}}^{+} } \psi_{i}(\bs{w}_{uf})\right]<-\delta^{*}.
		\end{align}
		
		In the case where (\ref{Th2_c3_pf1}) holds, we multiply both sides of it by $N_{a_{p}}N_{a_{q}}$. Then, we have
		\begin{align}\nonumber
		N_{a_{q}}\left[\sum_{i \in I_{a_{p}}^{-}} \psi_{i}(\bs{w}_{uf})-\sum_{i \in I_{a_{p}}^{+} } \psi_{i}(\bs{w}_{uf})\right]-N_{a_{p}}\left[\sum_{i \in I_{a_{q}}^{-}} \psi_{i}(\bs{w}_{uf})-\sum_{i \in I_{a_{q}}^{+}} \psi_{i}(\bs{w}_{uf})\right]<N_{a_{p}}N_{a_{q}}^{+}-N_{a_{q}}N_{a_{p}}^{+}-N_{a_{p}}N_{a_{q}}\delta^{*}.
		\end{align}
		
		Rearranging this inequality, it is easy to derive that
		\begin{align}\nonumber
		&N_{a_{q}}\left[\sum_{i \in I_{a_{p}}} \psi_{i}(\bs{w}_{uf})+\sum_{i \in I_{a_{q}}} \psi_{i}(\bs{w}_{uf})\right]\\\nonumber=&
		N_{a_{q}}\left[\sum_{i \in I_{a_{p}}^{-}} \psi_{i}(\bs{w}_{uf})+\sum_{i \in I_{a_{p}}^{+} } \psi_{i}(\bs{w}_{uf})+\sum_{i \in I_{a_{q}}^{-}} \psi_{i}(\bs{w}_{uf})+\sum_{i \in I_{a_{q}}^{+}} \psi_{i}(\bs{w}_{uf})\right]\\\label{Th2_c3_pf3} <& -N_{a_{p}}N_{a_{q}}\delta^{*}+N_{a_{p}}N_{a_{q}}^{+}+N_{a_{q}}N_{a_{p}}^{+}+N_{a_{p}}N_{a_{q}}^{-}+N_{a_{q}}N_{a_{q}}^{-}+(N_{a_{q}}-N_{a_{p}})\sum_{i \in I_{a_{q}}^{+}} \psi_{i}(\bs{w}_{uf}).
		\end{align}
		$\bullet$ If $N_{a_{q}}>N_{a_{p}}$, based on (\ref{Th2_c3_pf3}) we have
		\begin{align}
		&N_{a_{q}}\left[\sum_{i \in I_{a_{p}}} \psi_{i}(\bs{w}_{uf})+\sum_{i \in I_{a_{q}}} \psi_{i}(\bs{w}_{uf})\right]\\\nonumber<&-N_{a_{p}}N_{a_{q}}\delta^{*}+N_{a_{p}}N_{a_{q}}^{+}+N_{a_{q}}N_{a_{p}}^{+}+N_{a_{p}}N_{a_{q}}^{-}+N_{a_{q}}N_{a_{q}}^{-}+(N_{a_{q}}-N_{a_{p}})N_{a_{q}}^{+}.
		\end{align}
		The last inequality is due to the fact that $\sum_{i \in I_{a_{q}}^{+}} \psi_{i}(\bs{w}_{uf})\leq N_{a_{q}}^{+}$. This finally leads to
		\begin{align}\label{Th2_c3_pf4}
		\sum_{i \in I_{a_{p}}} \psi_{i}(\bs{w}_{uf})+\sum_{i \in I_{a_{q}}} \psi_{i}(\bs{w}_{uf})< N_{a_{p}}\left(\dfrac{N_{a_{q}}^{-}}{N_{a_{q}}}-\delta^{*}\right)+N_{a_{p}}^{+}+N_{a_{q}}.
		\end{align}
		$\bullet$ If $N_{a_{q}}\leq N_{a_{p}}$, from  (\ref{Th2_c3_pf3}) we have
		\begin{align}\nonumber
		N_{a_{q}}\left[\sum_{i \in I_{a_{p}}} \psi_{i}(\bs{w}_{uf})+\sum_{i \in I_{a_{q}}} \psi_{i}(\bs{w}_{uf})\right]< -N_{a_{p}}N_{a_{q}}\delta^{*}+N_{a_{p}}N_{a_{q}}^{+}+N_{a_{q}}N_{a_{p}}^{+}+N_{a_{p}}N_{a_{q}}^{-}+N_{a_{q}}N_{a_{q}}^{-},
		\end{align}
		since $\sum_{i \in I_{a_{q}}^{+}} \psi_{i}(\bs{w}_{uf})\geq0$. Then,
		\begin{align}\label{Th2_c3_pf5}
		\sum_{i \in I_{a_{p}}} \psi_{i}(\bs{w}_{uf})+\sum_{i \in I_{a_{q}}} \psi_{i}(\bs{w}_{uf})<(1-\delta^{*})N_{a_{p}}+N_{a_{p}}^{+}+N_{a_{q}}^{-}.
		\end{align}
		
		Now, we consider the other case where (\ref{Th2_c3_pf2}) holds. We multiply both sides of (\ref{Th2_c3_pf2}) also by $N_{a_{p}}N_{a_{q}}$. Applying similar steps as above, we can derive that
		\begin{align}\nonumber
		&N_{a_{p}}\left[\sum_{i \in I_{a_{p}}} \psi_{i}(\bs{w}_{uf})+\sum_{i \in I_{a_{q}}} \psi_{i}(\bs{w}_{uf})\right]\\\nonumber <& -N_{a_{p}}N_{a_{q}}\delta^{*}+N_{a_{q}}N_{a_{p}}^{+}+N_{a_{p}}N_{a_{q}}^{+}+N_{a_{q}}N_{a_{p}}^{-}+N_{a_{p}}N_{a_{p}}^{-}+(N_{a_{p}}-N_{a_{q}})\sum_{i \in I_{a_{p}}^{+}} \psi_{i}(\bs{w}_{uf}).
		\end{align}
		$\bullet$ If $N_{a_{q}}\geq N_{a_{p}}$, we can obtain
		\begin{align}\label{Th2_c3_pf6}
		\sum_{i \in I_{a_{p}}} \psi_{i}(\bs{w}_{uf})+\sum_{i \in I_{a_{q}}} \psi_{i}(\bs{w}_{uf})<(1-\delta^{*})N_{a_{q}}+N_{a_{q}}^{+}+N_{a_{p}}^{-}.
		\end{align}
		$\bullet$ If $N_{a_{q}}< N_{a_{p}}$, we have
		\begin{align}\label{Th2_c3_pf7}
		\sum_{i \in I_{a_{p}}} \psi_{i}(\bs{w}_{uf})+\sum_{i \in I_{a_{q}}} \psi_{i}(\bs{w}_{uf})< N_{a_{q}}\left(\dfrac{N_{a_{p}}^{-}}{N_{a_{p}}}-\delta^{*}\right)+N_{a_{q}}^{+}+N_{a_{p}}.
		\end{align}
		Combining this with (\ref{Th2_c3_pf4}), (\ref{Th2_c3_pf5}), and (\ref{Th2_c3_pf6}) gives 
		\begin{align}\nonumber
		&\sum_{i \in I_{a_{p}}} \psi_{i}(\bs{w}_{uf})+\sum_{i \in I_{a_{q}}} \psi_{i}(\bs{w}_{uf})\\\nonumber<&\mathbbm{1}\left[N_{a_{q}}>N_{a_{p}}\right]\max\left\{(1-\delta^{*})N_{a_{q}}+N_{a_{q}}^{+}+N_{a_{p}}^{-},N_{a_{p}}\left(\dfrac{N_{a_{q}}^{-}}{N_{a_{q}}}-\delta^{*}\right)+N_{a_{p}}^{+}+N_{a_{q}}\right\}\\\nonumber&+\mathbbm{1}\left[N_{a_{q}}<N_{a_{p}}\right]\max\left\{(1-\delta^{*})N_{a_{p}}+N_{a_{p}}^{+}+N_{a_{q}}^{-},N_{a_{q}}\left(\dfrac{N_{a_{p}}^{-}}{N_{a_{p}}}-\delta^{*}\right)+N_{a_{q}}^{+}+N_{a_{p}}\right\}\\\nonumber&+\mathbbm{1}\left[N_{a_{q}}=N_{a_{p}}\right]\max\left\{(1-\delta^{*})N_{a_{q}}+N_{a_{q}}^{+}+N_{a_{p}}^{-}, (1-\delta^{*})N_{a_{p}}+N_{a_{p}}^{+}+N_{a_{q}}^{-}\right\}\\\label{Th2_c3_pf8}=&\mathcal{M}(p,q,\delta^{*}).
		\end{align}
		We summarize (\ref{Th2_c3_pf8}) over $c\choose 2$ unique pairs of $p, q$. Then, we can easily obtain
		\begin{align}\nonumber
		\sum_{i=1}^{n}\psi_{i}(\bs{w}_{uf})<\dfrac{1}{c-1}{\sum_{p\neq q}}\mathcal{M}(p,q,\delta^{*}).
		\end{align}
		Thus, 
		\begin{align}\nonumber
		\mathcal{V}(\bs{w}_{uf})&=\sum_{i=1}^{n}a_{i}-\sum_{i=1}^{n}b_{i}\psi_{i}(\bs{w}_{uf})\\\nonumber&\geq\sum_{i=1}^{n}a_{i}-\max_{i}b_{i}\sum_{i=1}^{n}\psi_{i}(\bs{w}_{uf})\\\nonumber&>\sum_{i=1}^{n}a_{i}-\dfrac{\max_{i}b_{i}}{c-1}{\sum_{p\neq q}}\mathcal{M}(p,q,\delta^{*}).
		\end{align}
		We define $\Delta^{*}=\sum_{i=1}^{n}a_{i}-\dfrac{\max_{i}b_{i}}{c-1}{\sum_{p\neq q}}\mathcal{M}(p,q,\delta^{*})$. Afterwards, it directly follows that $\mathcal{V}(\bs{w}^{data})\geq \mathcal{V}(\bs{w}_{uf})>\Delta^{*}$ with the \textit{statistical parity} notion.
	
	Consequently, there all exists $\Delta^{*}$ related to $\delta^{*}$ for three different fairness notions such that  $\mathcal{V}(\bs{w}^{data})>\Delta^{*}$. Because of the fact that $\mathcal{V}(\bs{w}^{data})\geq\mathcal{V}(\bs{w}^{*})$, we can conduct that $\mathcal{V}(\bs{w}^{data})\geq\max\left\{\Delta^{*}, \mathcal{V}(\bs{w}^{*})\right\}$.
	
	Now, we have shown that in these three fairness notions,  $\mathcal{V}(\bs{w}^{data})>\Delta^{*}$ and $\mathcal{V}(\bs{w}^{data})\geq \mathcal{V}(\bs{w}^{*})$. Due to the definition of $\bs{w}^{*}$, we have $SWF(\bs{w}^{*})\geq SWF(\bs{w}^{data})$ which leads to
	\begin{align}\label{Th2_final}
	\mathcal{V}(\bs{w}^{*})-\bar{\rho}\delta^{*}\geq\mathcal{V}(\bs{w}^{data})-\bar{\rho}\delta^{data}
	\end{align}
	where $\delta^{data}\in[0,1]$ is the (un)fairness level of $\bs{w}^{data}$. Combining (\ref{Th2_final}) and $\mathcal{V}(\bs{w}^{data})>\Delta^{*}$, we can obtain
	\begin{align}\label{Th2_final2}
	SWF(\bs{w}^{*})=	\mathcal{V}(\bs{w}^{*})-\bar{\rho}\delta^{*}>\Delta^{*}-\bar{\rho}\delta^{data}.
	\end{align}
	If $\bar{\rho}>0$, the right-hand side of the inequality in (\ref{Th2_final2}) is greater than $\Delta^{*}-\bar{\rho}$. If $\bar{\rho}\leq0$, it is greater than $\Delta^{*}$. This yields the statement of the theorem. 
	
	\section{Supplementary Information for Experimental Study}
	
	\subsection{Sepsis Mortality Prediction}
	
	\subsubsection{Summary Statistics of Data Set}
	
	The MIMIC-III database contains information related to patients admitted to ICU at a large tertiary care hospital between 2001 and 2012 in the U.S. This database is publicly accessible. It includes patients' records and information like their demographics, vital sign measurements, laboratory test results, procedures, medications, etc. The \textit{Sepsis} data set is extracted from MIMIC-III. It involves a subset of MIMIC-III patients who are diagnosed with sepsis, severe sepsis, or septic shock based on the ICD-9 code suggested in \citet{singer2016third1}. 
	
	Table \ref{sepsisdata_table} displays the summary statistics of the final \textit{Sepsis} data set.
	
	\begin{table} [!htbp]
		\caption{Summary statistics of the sepsis data set}
		\label{sepsisdata_table}
		\footnotesize
		\centering
		\begin{tabular}{cccccccc}
			\toprule
			Data set&{$N$} & {$d_{original}$} &{$d$}&{Sensitive feature}& Positive$\%$ & Male$\%$\\
			\midrule
			\textit{Sepsis} & $2021$ & $19$ & $77$ & Gender (binary) &$33.40\%$&$53.74\%$\\
			\bottomrule      
		\end{tabular}
	\end{table}
	
	\subsubsection{Baseline Scoring Systems}
	In this experiment, we consider the following $5$ baseline scoring systems that are widely applied for mortality prediction in medical practice:
	\begin{quotation}
		\begin{description}
			\item
			\item[\bf SAPS II:] Simplified Acute Physiology Score \citep{LeGall19931} is developed based on medical data from 137 ICUs of 12 countries in Europe/North America. The scoring system measures 17 variables of ICU patients and assigns different points to different variables. A regression-based model is provided to convert a total score to a mortality probability. 
			
			\item[\bf LODS:] Logistic Organ Dysfunction System \citep{le1996logistic1} uses the same database as SAPS II for model development but aims to assess the dysfunction levels of 6 human organ systems among ICU patients. The total dysfunction score ranges from 0 to 22 and can also be converted to a mortality probability by a logistic regression model.
			
			\item[\bf SOFA:] Developed through expert consensus, Sepsis-related Organ Failure Assessment \citep{vincent1996sofa1} measures the degree of organ failure and evaluates morbidity of septic patients. The total score ranges from 0 to 24, with 0 to 4 for each of the six organ systems. Although the score is not initially designed to predict patient survival, it has become a basic variable in many mortality prediction models due to a high correlation between organ failure and survival.
			
			\item[\bf qSOFA:] The quick SOFA \citep{singer2016third1} is a simplified and quick version of SOFA, often used at the bedside to identify patients with suspected infection who are at greater risk of bad clinical outcomes outside the ICU. It only consists of three criteria (1 point for each) about blood pressure, respiratory rate, and central nervous system status. A score $\ge 2$ is usually considered a sepsis case and is associated with at least a threefold increase in in-hospital mortality.
			
			\item[\bf SIRS:] Systemic Inflammatory Response Syndrome \citep{bone1992definitions1} is actually not a scoring system specifically designed for sepsis mortality, but we include it in our experiments as it forms an essential part of the initial definition of sepsis: a host's systemic inflammatory response syndrome to infection. The manifestation by two or more of the four conditions of SIRS is considered to be a SIRS case. We manually assign 1 point to each condition and assume a score of $>$ 3 to be a positive case. 
			
		\end{description}
	\end{quotation}
	
		The above scoring systems have been widely used by medical centers and researchers worldwide \citep{arabi2003assessment}. All the systems can be used at ICU  admission or 24 hrs after admission for evaluation of the severity of sickness and for mortality prediction. In general, higher scores indicate more severe health conditions and hence a higher risk of mortality. We assume a risk probability greater than 0.5 to be at a high mortality risk for prediction models such as SAPS II and LODS. Similarly, we consider a cutoff score greater than 12, 2, and 3 to be at a high mortality risk for SOFA, qSOFA, and SIRS, respectively. 

	\subsubsection{Social Welfare Maximization}
	
	Table \ref{TableSM_Sepsis_welfare} represents the complete results of scoring systems both on training and test sets. 
	All results on these sets suggest that our methods can achieve the optimal social welfare compared to baseline scoring systems. 
	
	\begin{table} [!htbp]
		\caption{The average values of total social welfare for all scoring systems on \textit{Sepsis} data set.}
		\label{TableSM_Sepsis_welfare}
		\footnotesize
		\centering
		\begin{threeparttable}
			\begin{tabular}{cccccccccc}
				\toprule
				\multirow{2}{*}{Dataset}&Fairness&\multirow{2}{*}{$\bar{\rho}$}& \multicolumn{5}{c}{Baselines}  & \multicolumn{2}{c}{Ours} \\
				\cmidrule(l){4-8} \cmidrule(l){9-10} 
				&Notions& & SAPS II& LODS &SOFA&qSOFA&SIRS&  FASS& FASS7\\
				\midrule
				\multirow{8}{*}{\textit{Sepsis}}&\multicolumn{8}{l}{Training set}\\\cmidrule(l){2-10} 
				&SP & 5 &0.7024&0.5904&0.5540&0.3805&0.5015 &\textbf{0.7124} &0.7060 \\
				&EO  & 0.2 & 0.7269&0.7217&0.7158&0.6167&0.5355&\textbf{0.7665}&0.7507\\
				&OMR  & 5 &0.6298&0.6601&0.6400&0.5871&0.3281&\textbf{0.7442}&0.7339\\
				\cmidrule(l){2-10} 
				&\multicolumn{8}{l}{Test set}\\\cmidrule(l){2-10} 
				
				&SP  &5 & 0.6761&0.5368&0.5638&0.5089&0.4520 &\textbf{0.6916}&0.6869 \\
				&EO  &0.2 & 0.7102&0.7198&0.7096&0.6278&0.5373&\textbf{0.7550}&0.7472\\
				&OMR  &5 & 0.5768&0.6393&0.5787&0.5509&0.4022 &\textbf{0.7405}&0.7223\\ 		
				\bottomrule
			\end{tabular}
			\begin{tablenotes}
				\item Note: \textit{Statistical parity} is denoted by SP, \textit{equality of opportunity} by EO, and \textit{equal overall misclassification rate} by OMR. The optimal values are highted in bold.
			\end{tablenotes}		
		\end{threeparttable}
	\end{table}
	
	\subsubsection{Interpretability of Scorecard}
	
	In medical applications, for example, the following statement of \cite{than2014development1} describes the importance of sensibility\footnotemark[3] of produced models: 
		\footnotetext[3]{As mentioned in \cite{than2014development1}, ``sensibility" refers to whether a prediction rule is both clinically reasonable and easy to use.}
		\begin{quote}
			\textit{``An important consideration during development is the clinical sensibility of the resulting prediction rule [...] Evaluation of sensibility requires judgment rather than statistical methods. A sensible rule is easy to use, and has content and face validities. Prediction rules are unlikely to be applied in practice if they are not considered sensible by the end-user, even if they are accurate.”}
		\end{quote}
	
	Thus, it is necessary for the scorecard to be interpretable and reasonable for medical applications. As shown in the main manuscript, the developed EO scorecard identified several risk factors that have been recognized by medical research, which provides clinical sensibility. For example, it identifies FiO2 (Fraction of inspired oxygen) with cut-off values at around $0.8$. This is consistent with the findings in \citet{dahl2015variability1} that the relative risk of mortality is $2.1$ in patients with an average FiO2 $ \geq 0.80$ as compared to patients with an average FiO2 $\leq 0.40$. Moreover, it also points out the risk factors, like GCS (Glasgow Coma Scale) at $9$. GCS reflects a patient's degree of disturbance of consciousness. A GCS below $9$ is well-acknowledged as severe disturbance and is associated with higher death risk \citep{Kurowski2016GCS1}. Other risk factors like age at 80, and potassium at 4.25 also play roles in the scorecard, and their effect on the prediction of ICU mortality has been demonstrated in \citet{gogos2003clinical1,solinger2013risks1}, and \citet{martin2019risk1}.
	
	In addition, the EO scorecard identifies multiple cut-off values associated with variables, which are not yet recognized in the literature. Examples include GCS at 5 and 7, bilirubin at 2.3 and 7.3, and pH.art at 7.2. This may help reveal possible and complicated interactions between these sepsis-related variables. The interaction between GCS and bilirubin, as suggested in the fifth rule in our EO scorecard, may affect the risk thresholds of both variables. Recent evidence like \citet{wang2020serum1} argues that the level of bilirubin correlates with mortality in patients with traumatic brain injury who always have lower GCS. Further, \citet{sedlak2004bilirubin1} and \citet{marconi2018bilirubin1} point out that a high bilirubin level sometimes confers various health benefits due to its antioxidant activity. For pH.art, the study of \citet{kraut2010metabolic1} suggests that metabolic acidosis might be beneficial for oxygen delivery and metabolism, and thus a slight upward adjustment of pH cut-off from 7.1 to 7.2 may not always be harmful, which is also shown in the fourth rule of the EO scorecard.

\begin{figure} [!htbp]
		\begin{centering}
			\subfigure[Statistical Parity]{ 
				\includegraphics[width=190pt]{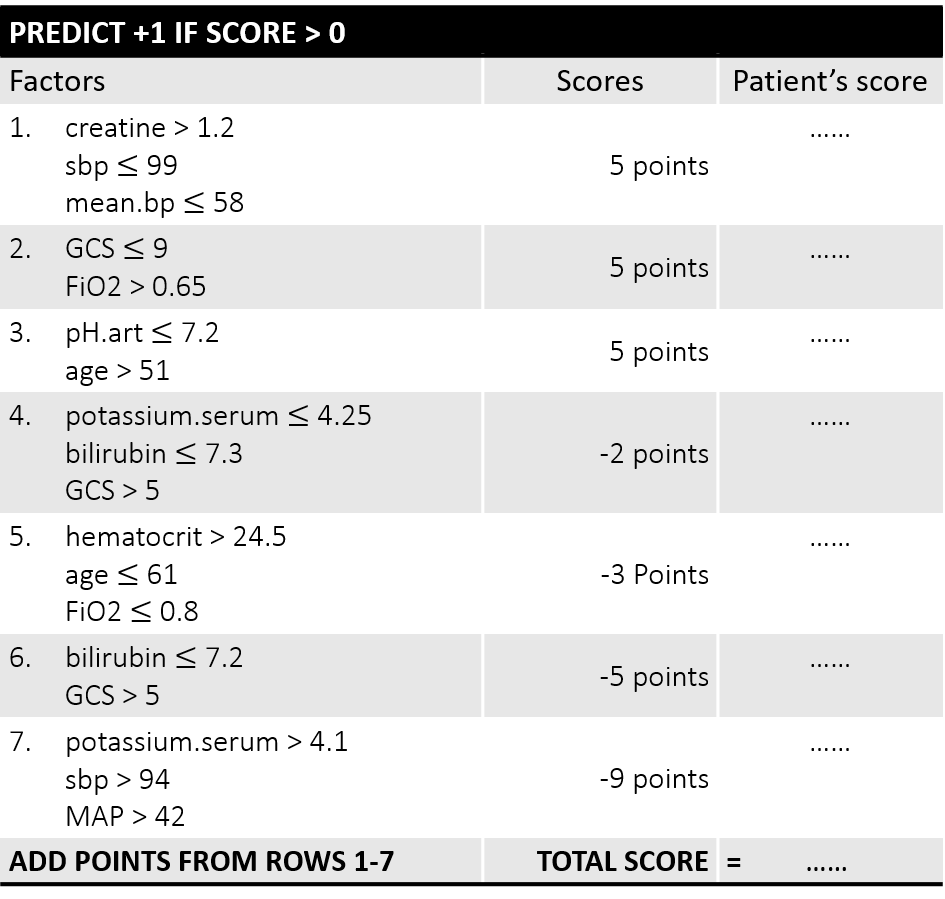}} 
			\hspace{25pt}
			\subfigure[Equality of Opportunity]{ 
				\includegraphics[width=190pt]{Fig//Scorecard_EO.png}} 
			\hspace{1in} \subfigure[Equal Overall Misclassification Rate]{ 
				\includegraphics[width=190pt]{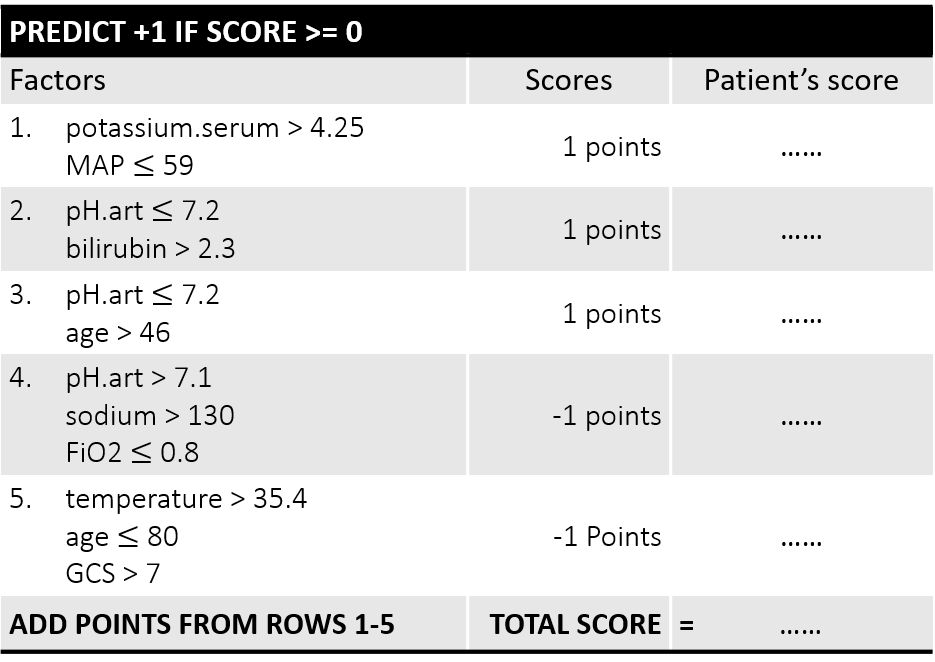}} 
			\caption{Scorecards developed by FASS7 for Sepsis prediction.}
			\label{figSM:scorecard_Sepsis}
		\end{centering}
		\footnotesize
		\textit{Note:} The training welfare of the SP scorecard and EO scorecard  is $0.6896$ and $0.7466$, respectively. The scorecard derived for equal OMR achieves a training welfare of $0.7277$.
	\end{figure}

	The complete scorecards for all fairness metrics are displayed in Figure \ref{figSM:scorecard_Sepsis}. Similar to the EO scorecard, the SP and OMR scorecards also identify several sensible risk factors. All of these results indicate that the developed scorecards can capture sensible risk-predictive rules. Furthermore, they help in revealing the complexity of sepsis by discovering promising interactions between these variables, which may give insights to further medical research and decision-making.
	
	\subsection{UCI Numerical Experiments}
	
	This section shows the detailed information and complete experimental results on UCI data sets.
	\subsubsection{Data Processing and Summary Statistics}
	
	Because \textit{Adult} and \textit{German} data sets are imbalanced (the positive rate in the raw data is $24\%$ for the \textit{Adult} data set and $30\%$ for the \textit{German} data set), some sampling methods in imbalanced learning are applied to eliminate an impact on the results. We used random undersampling on the \textit{Adult} data set and SMOTENC \citep{Chawla2002smote1} on the \textit{German} data set to create balanced data sets for the purpose of comparing the relative performance between classifiers. The summary statistics of final data sets are shown in Table \ref{UCIdata_table}, where gender (with feature values: male and female) is used as a sensitive feature.
	
	\begin{table} [!htbp]
		\caption{Summary of real UCI data sets}
		\label{UCIdata_table}
		\scriptsize
		\centering
		\begin{tabular}{lccccccccc}
			\toprule
			{Data set}&{$N_{original}$} & {$d_{original}$} & {$N$}&{d}&Male$\%$ &{Sensitive feature}\\
			\midrule
			\textit{Adult} & $48,842$ & $14$ & $2,000$ & $36$ & $72.8\%$&  \multirow{2}{*}{Gender (binary)}\\
			\textit{German} & $1,000$ & $20$ & $3,000$ &$65$&$68.1\%$ &  \\		
			\bottomrule      
		\end{tabular}
	\end{table}

	\subsubsection{Social Welfare Maximization}
	
\begin{figure} [!htbp]
		\centering
		\subfigure[Training Set]{ 
			\includegraphics[width=315pt]{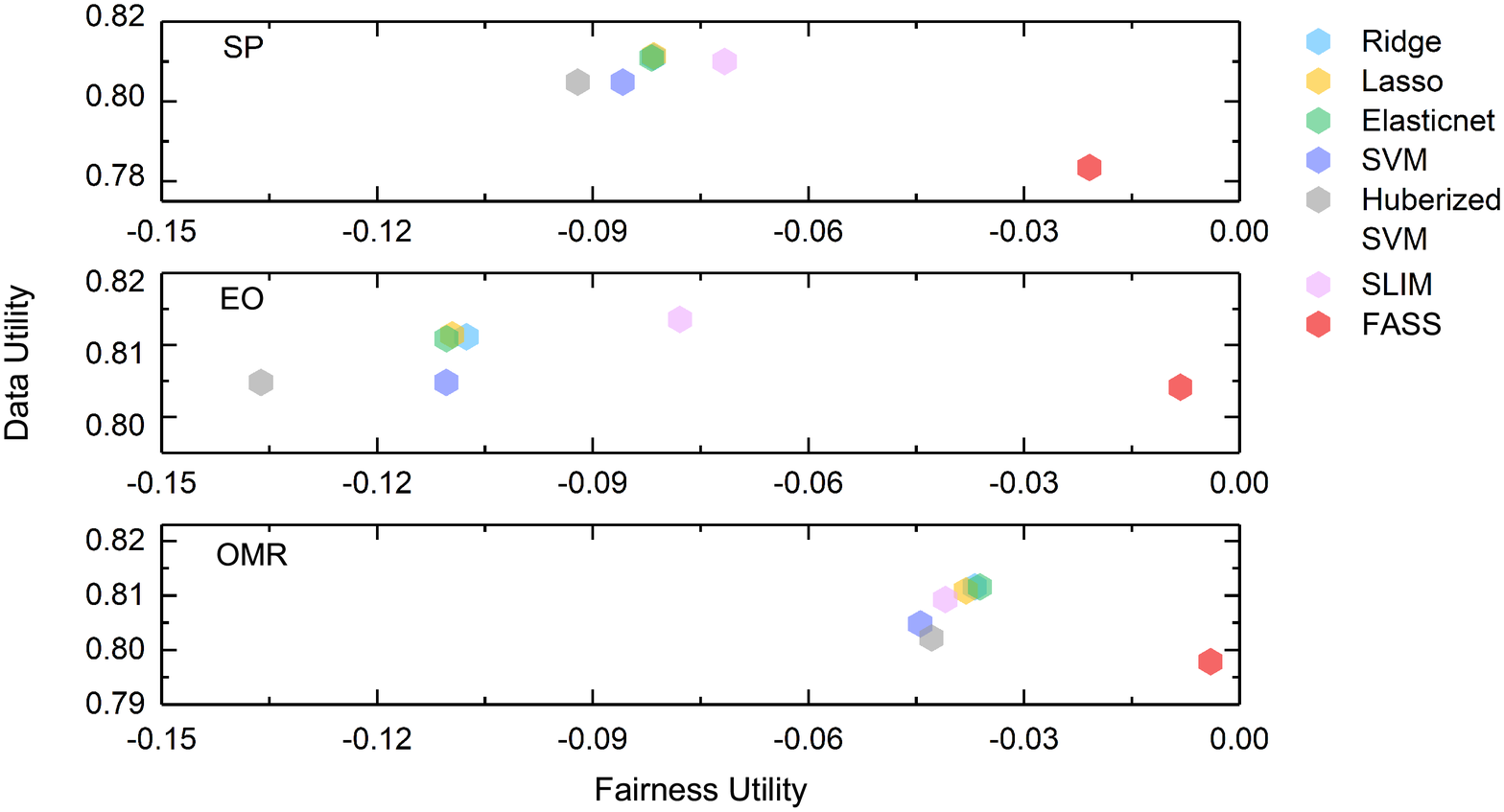}}
		\subfigure[Test Set]{ 
			\includegraphics[width=315pt]{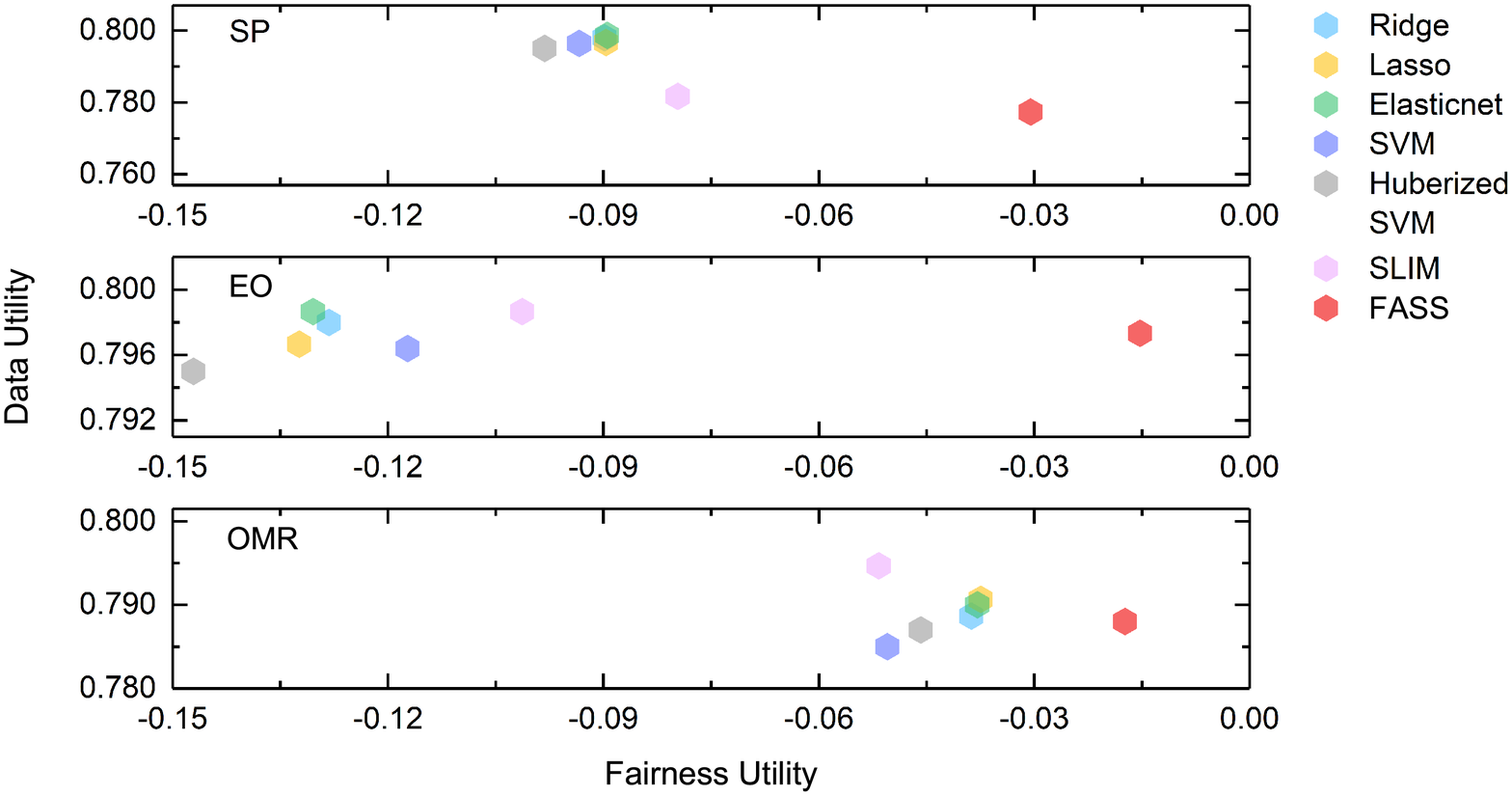}}
		\caption{Data utility and fairness utility with different fairness measures on \textit{Adult} data set.}
		\label{fig:Adult}
	\end{figure}

\begin{figure} [!htbp]
		\centering
		\subfigure[Training Set]{ 
			\includegraphics[width=315pt]{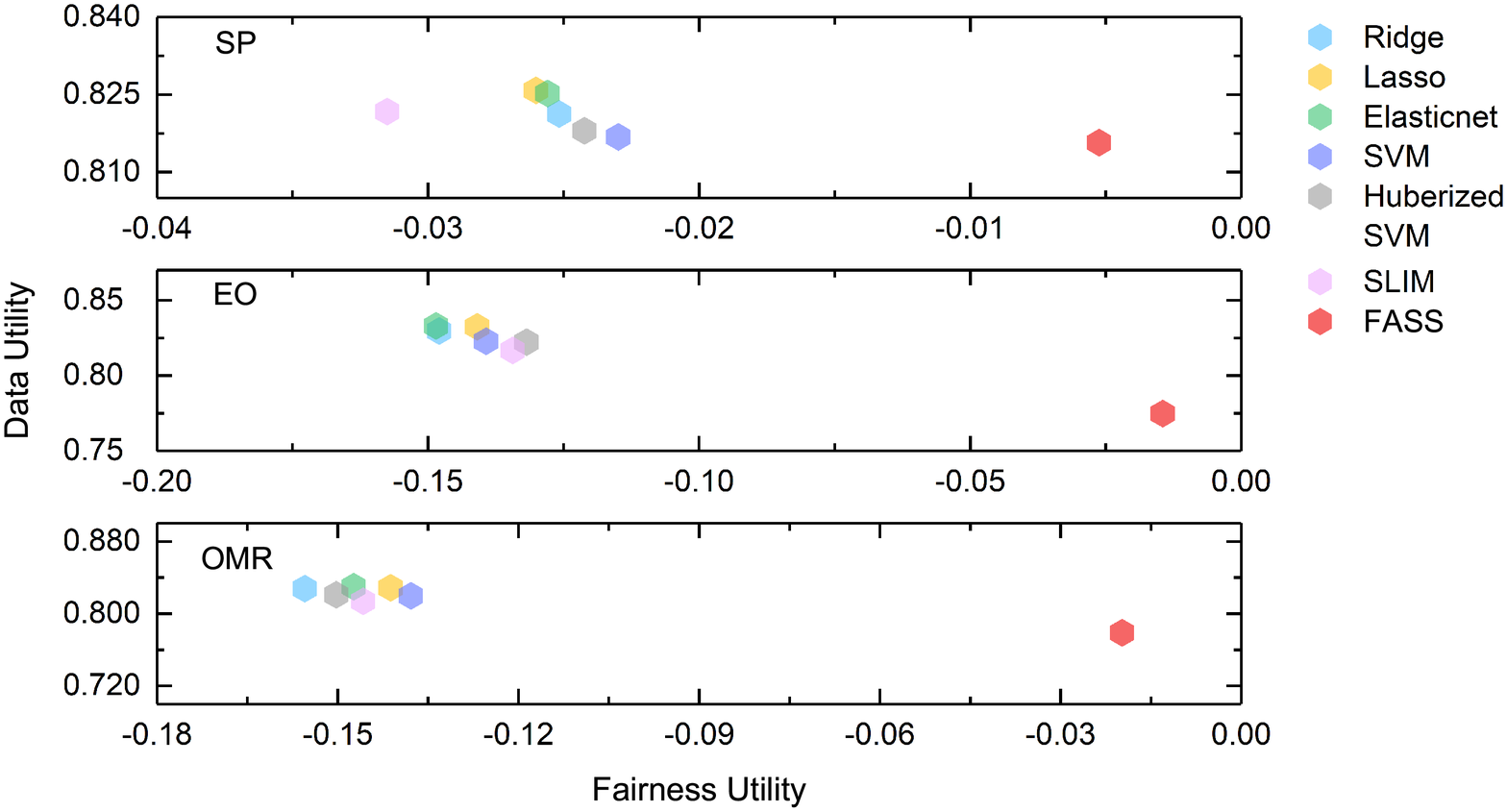}}
		\subfigure[Test Set]{ 
			\includegraphics[width=315pt]{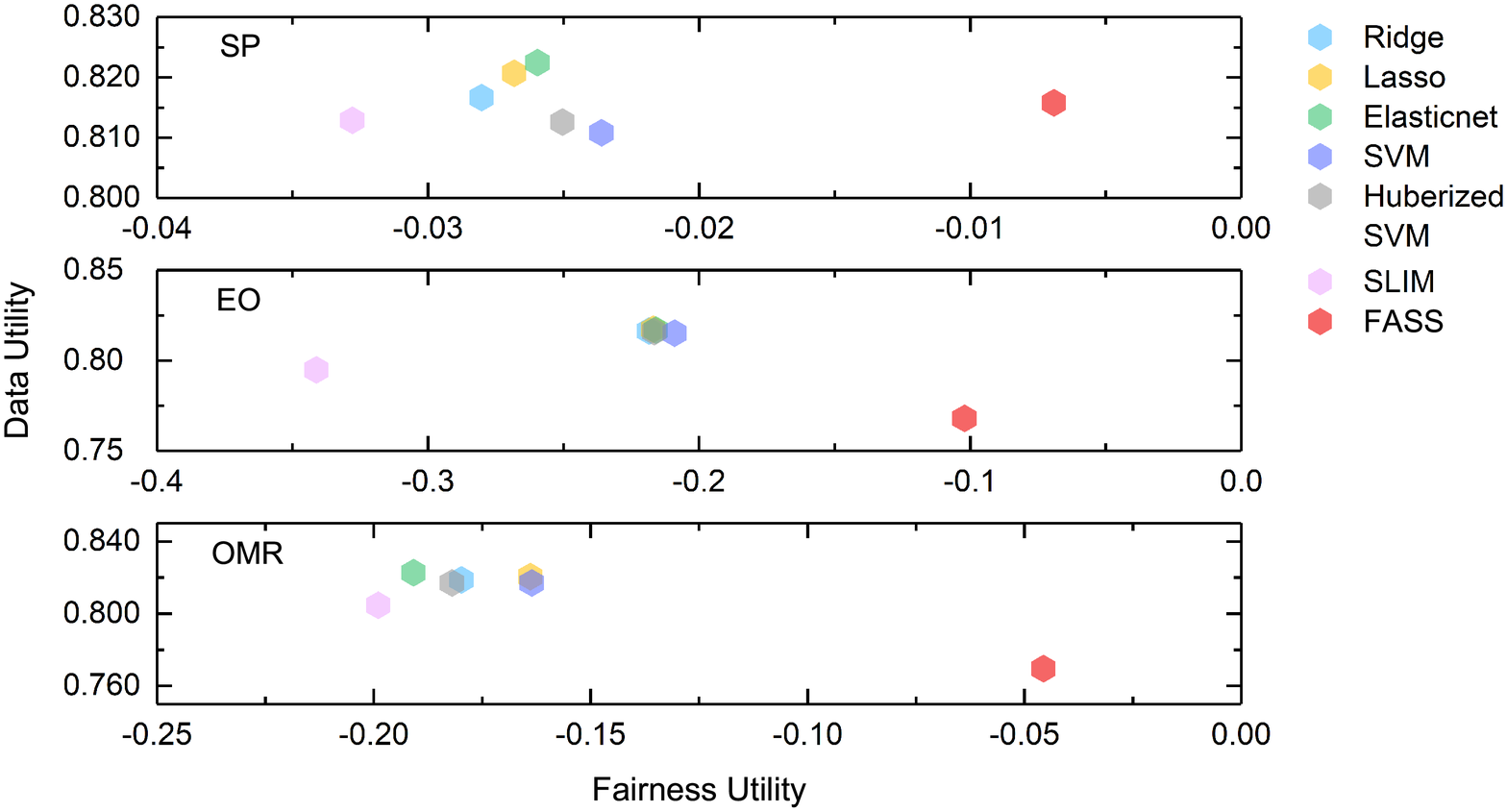}}
		\caption{Data utility and fairness utility with different fairness measures on \textit{German} data set.}
		\label{fig:German}
	\end{figure}

	We provide the total social welfare of all methods on training and test sets in Table \ref{TabSM_UCI1}. The complete results of the data utility and fairness utility on \textit{Adult} and \textit{German} data sets are displayed in Figures \ref{fig:Adult} and \ref{fig:German}, respectively.  These results verify again the effectiveness of the proposed method. 
	
	\begin{table} [!htbp]
		\caption{The average values of total social welfare for all methods on UCI data sets.}
		\label{TabSM_UCI1}
		\footnotesize
		\centering
		\begin{threeparttable}
			\begin{tabular}{cccccccccc}
				\toprule
				\multirow{2}{*}{Dataset}&Fairness&\multirow{2}{*}{$\bar{\rho}$}& \multicolumn{6}{c}{Baselines}  & \multicolumn{1}{c}{Ours} \\
				\cmidrule(l){4-9} \cmidrule(l){10-10} 
				&Notions& &Ridge& Lasso&Elasticnet&SVM& Huberized SVM& SLIM& FASS\\
				\midrule
				\multirow{8}{*}{\textit{Adult}}&\multicolumn{9}{l}{Training set}\\\cmidrule(l){2-10} 
				&SP & 0.2 & 0.7296&0.7299&0.7290&0.7190&0.7127&0.7384&\textbf{0.7626} \\
				&EO  & 0.5   & 0.7036&0.7018&0.7005&0.6944&0.6686&0.7357&\textbf{0.7959}\\
				&OMR  &  0.5 & 0.7747&0.7728&0.7754&0.7604&0.7594&0.7684&\textbf{0.7938}\\
				\cmidrule(l){2-10} 
				&\multicolumn{9}{l}{Test set}\\\cmidrule(l){2-10} 
				
				&SP  &0.2 & 0.7082&0.7070&0.7092&0.7030&0.6968&0.7020&\textbf{0.7468}\\
				&EO  &0.5 & 0.6698&0.6643&0.6682&0.6791&0.6479&0.6973&\textbf{0.7821}\\
				&OMR  &0.5 & 0.7499&0.7532&0.7521&0.7345&0.7412&0.7430&\textbf{0.7707}\\ 
				\midrule
				
				\multirow{8}{*}{\textit{German}}&\multicolumn{9}{l}{Training set}\\\cmidrule(l){2-10} 
				&SP & 0.2 &0.7961&0.7998&0.7996&0.7938&0.7938&0.7902&\textbf{0.8105}  \\
				&EO  & 5  & 0.6818&0.6915&0.6845&0.6835&0.6904&0.6824&\textbf{0.7604}\\
				&OMR  &  5 & 0.6721&0.6875&0.6825&0.6818&0.6705&0.6680&\textbf{0.7591}\\
				\cmidrule(l){2-10} 
				&\multicolumn{9}{l}{Test set}\\\cmidrule(l){2-10} 
				
				&SP  &0.2 & 0.7886&0.7938&0.7965&0.7872&0.7876&0.7801&\textbf{0.8089}\\
				&EO  &5& 0.5980&0.6004&0.6009&0.6061&0.5994&0.4537&\textbf{0.6659} \\
				&OMR  &5& 0.6389&0.6565&0.6318&0.6534&0.6350&0.6057&\textbf{0.7240}\\ 
				
				\bottomrule
			\end{tabular}
			\begin{tablenotes}
				\item Note: \textit{statistical parity} is denoted by SP, \textit{equality of opportunity} by EO, and \textit{equal overall misclassification rate} by OMR. The optimal values are highted in bold.
			\end{tablenotes}		
		\end{threeparttable}
	\end{table}

	\clearpage
	
		\clearpage
	
{In the previous experiments, we set $a_{i}=b_{i}=1$ for all $i=1,\dots,n$ for simplicity. This implicitly assumes all types of misclassification errors cost equally since $b_{i}$ is same for all individuals. However, in some real applications such as credit evaluation, the costs of different types of classification errors may be not equal. In these situations, decision-makers may would like to take the varying costs associated with misclassifying examples into considerations. 
	Therefore, in the following experiments, we take the \textit{German} credit data set as an example to verify the performance of the proposed method with the cost-sensitive setting. Here, we focus on the case where the misclassification cost depends only on the class of an individual.	For simplicity, we assume the cost of false negative is higher than false positive, and the ratio between two costs is set arbitrarily as $C_{FN}:C_{FP} =4:1$. In other words, for $\forall i \in I^{+}=\big\{i\in\{1,2,\dots,n\}\left| y_{i}=1\right.\big\}$ and $\forall j \in I^{-}=\big\{i\in\{1,2,\dots,n\}\left| y_{i}=0\right.\big\}$, we have $b_{i}:b_{j}=4:1$. We restrict ourselves to $b_{i}+b_{j}=2$ without loss of generality. In addition, several cost-sensitive baseline classifiers\footnotemark[1] are also introduced for comparison. \footnotetext[1]{We extend the baselines used in Subsection \ref{sec_CIbaseline} to their cost-sensitive versions by applying the sample-based weighting method proposed in \citet{Yang2021}.} 
	
	Tables \ref{TabSM_UCIcost} and \ref{TabSM_UCIcost2} shows the average values of total social welfare for all methods on \textit{German} data set while considering cost-sensitivity. These results verify again that the proposed method still achieves the optimal social welfare for cost-sensitive classification.} 	
	
		\begin{table} [!htbp]
		\caption{The average values of total social welfare for all methods on \textit{German} data set while considering cost-sensitivity.}
		\label{TabSM_UCIcost}
		\footnotesize
		\centering
		\begin{threeparttable}
			\begin{tabular}{ccccccccc}
				\toprule
				Fairness&\multirow{2}{*}{$\bar{\rho}$}& \multicolumn{6}{c}{Baselines}  & \multicolumn{1}{c}{Ours} \\
				\cmidrule(l){3-8} \cmidrule(l){9-9} 
				Notions& &Ridge& Lasso&Elasticnet&SVM& Huberized SVM& SLIM& FASS\\
				\midrule
				\multicolumn{9}{l}{Training set}\\\cmidrule(l){1-9} 
				SP & 0.2 & 0.8228&0.8180&0.8192&0.8393& 0.8226&0.8047&\textbf{0.8868}\\
		        EO & 5 & 0.6988&0.7108&0.7038&0.7194&0.6557&0.7023&\textbf{0.8780}\\
		        OMR & 5 & 0.7177&0.7036&0.7295&0.7561& 0.6864&0.7155&\textbf{0.8121}\\
				\cmidrule(l){1-9} 
				\multicolumn{9}{l}{Test set}\\\cmidrule(l){1-9} 
				
				SP & 0.2 &0.8133&0.8116&0.8252&0.8328&0.8271&0.7952& \textbf{0.8715}\\
		        EO & 5 &0.6265&0.6103&0.6103&0.6327&0.6690&0.4850& \textbf{0.8483}\\
		        OMR & 5 &0.6760&0.7654&0.6380&0.6728&0.7175&0.6137& \textbf{0.7964}\\
			\bottomrule
			\end{tabular}
			\begin{tablenotes}
				\item Note: \textit{statistical parity} is denoted by SP, \textit{equality of opportunity} by EO, and \textit{equal overall misclassification rate} by OMR. The optimal values are highted in bold.
			\end{tablenotes}		
		\end{threeparttable}
	\end{table}

		\begin{table} [!htbp]
		\caption{The average values of total social welfare for all cost-sensitive classifiers on \textit{German} data set while considering cost-sensitivity.}
		\label{TabSM_UCIcost2}
		\footnotesize
		\centering
		\begin{threeparttable}
			\begin{tabular}{cccccccc}
				\toprule
				Fairness&\multirow{2}{*}{$\bar{\rho}$}& \multicolumn{5}{c}{Baselines}  & \multicolumn{1}{c}{Ours} \\
				\cmidrule(l){3-7} \cmidrule(l){8-8} 
				Notions& &$\text{Ridge}_{\text{CS}}$&$\text{Lasso}_{\text{CS}}$&$\text{Elasticnet}_{\text{CS}}$&$\text{SVM}_{\text{CS}}$&$\text{Huberized}$ $\text{SVM}_{\text{CS}}$& FASS\\
				\midrule
				\multicolumn{8}{l}{Training set}\\\cmidrule(l){1-8} 
				SP & 0.2 & 0.8640&0.8641&0.8656&0.8522&0.8387&\textbf{0.8868}\\
		        EO & 5 &0.7996&0.8084&0.8020&0.7789&0.7518&\textbf{0.8780}\\
		        OMR & 5 &0.8034&0.7809&0.7740&0.7972& 0.7808 &\textbf{0.8121}\\
				\cmidrule(l){1-8} 
				\multicolumn{8}{l}{Test set}\\\cmidrule(l){1-8} 
				SP & 0.2 & 0.8621&0.8608&0.8625&0.8521&0.8316 & \textbf{0.8715}\\
		        EO & 5 & 0.7750&0.7785&0.7749&0.7290&0.7201& \textbf{0.8483}\\
		        OMR & 5 &0.7181&0.7340&0.7024&0.7554&0.7387 & \textbf{0.7964}\\
			\bottomrule
			\end{tabular}
			\begin{tablenotes}
				\item Note: \textit{statistical parity} is denoted by SP, \textit{equality of opportunity} by EO, and \textit{equal overall misclassification rate} by OMR. The optimal values are highted in bold.
			\end{tablenotes}		
		\end{threeparttable}
	\end{table}

	\subsubsection{Impact of Average Preference for Fairness}
	
\begin{figure} [!htbp]
		\begin{centering}
			\centering
			\subfigure[Training Set]{ 
				\includegraphics[width=440pt]{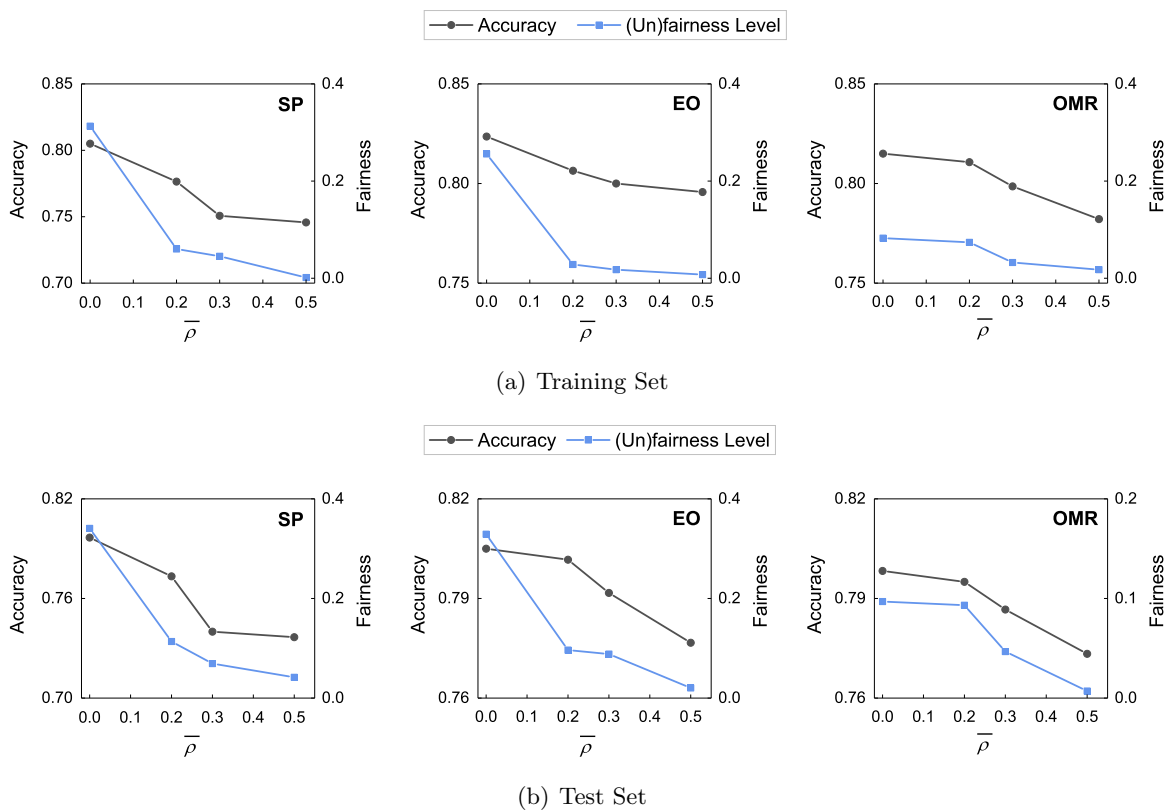}}\\
			
			\subfigure[Test Set]{ 
				\includegraphics[width=440pt]{Fig//Adult_rhobar_Te.eps}}
			\caption{Trade-offs between accuracy and fairness on \textit{Adult} data set in a randomly selected run.}
			\label{fig:AveragePreference}
		\end{centering}
	\end{figure}

\begin{figure} [!hbtp]
		\begin{centering}
			\centering
			\subfigure[Training Set]{ 
				\includegraphics[width=440pt]{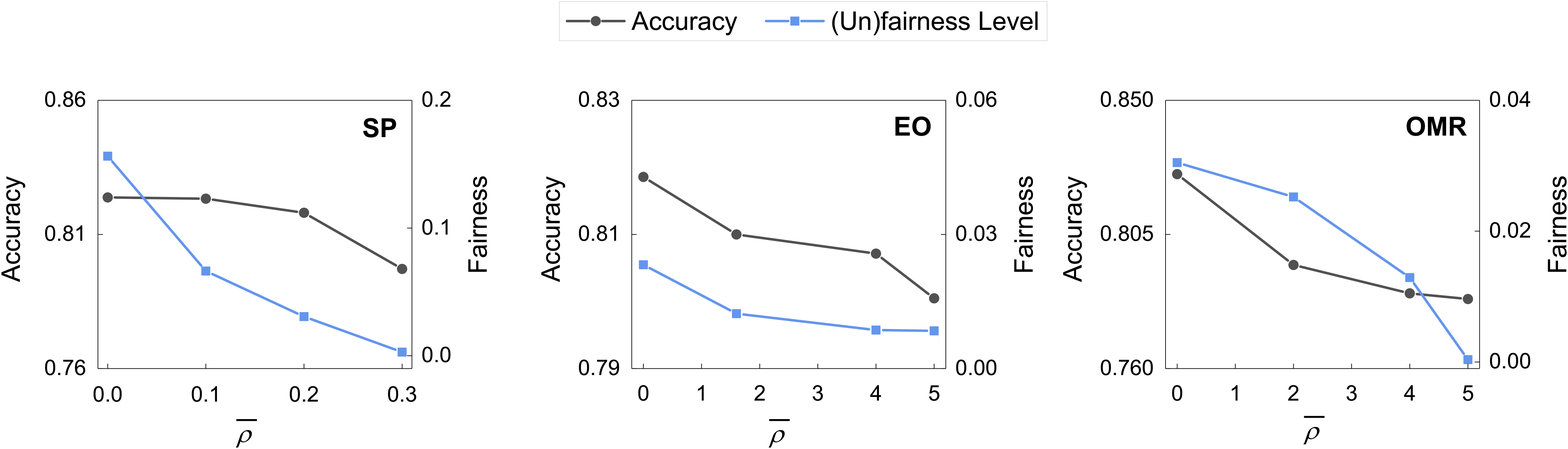}}		
			\subfigure[Test Set]{ 
				\includegraphics[width=440pt]{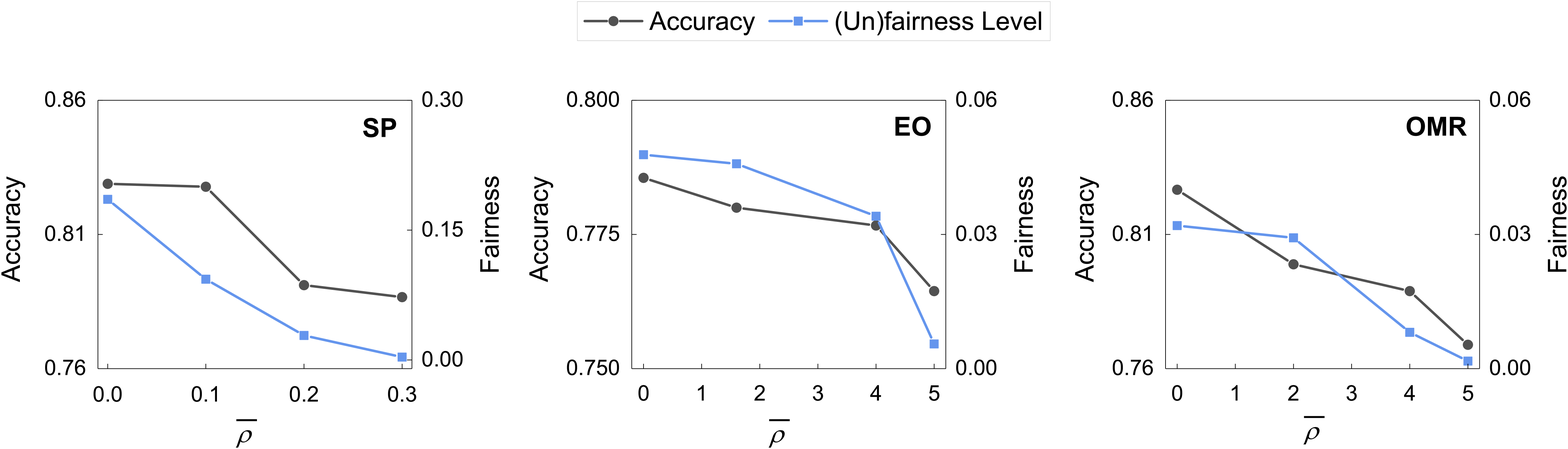}}
			\caption{Trade-offs between accuracy and fairness on \textit{German} data set in a randomly selected run.}
			\label{fig:GGAveragePreference}
		\end{centering}
	\end{figure}

	The trade-offs between accuracy and fairness with different average preference for fairness on \textit{Adult} and \textit{German} data sets are represented in Figures \ref{fig:AveragePreference} and \ref{fig:GGAveragePreference}, respectively.

	\subsubsection{Classification Model for Specified $\delta$}
		
		Finally, we consider a simplified case where the fairness level $\delta^{s}$ is specified in advance, as discussed previously in Section \ref{degdelta}. In this situation, the data utility maximizing scoring system which satisfies the given fairness level is developed. Tables \ref{tab: sepciAdult} and \ref{tab: sepciGerman} provide the data utility and (un)fairness level on training and test sets when the maximal unfairness level is pre-specified on two data sets.

			\begin{table} [!htbp]
		\captionsetup{font={footnotesize}}
		\caption{The average values of data utility and fairness level for all methods on \textit{Adult} data set with $\delta^{s}=0.05$.}
		\label{tab: sepciAdult}
		\footnotesize
		\centering
		\begin{threeparttable}
			\begin{tabular}{@{}lccccccccc@{}}
				\toprule
				& \multicolumn{6}{c}{Baselines}  & \multicolumn{3}{c}{Ours}   \\ \cmidrule(l){2-7} \cmidrule(l){8-10} 
				Metric &  Ridge  &  Lasso  &  Elasticnet  &  SVM  &  Huberized SVM  &  SLIM  &  FASS-EO  &  FASS-OMR  & FASS-SP\\
				\midrule
				\multicolumn{10}{l}{Training set}\\
				\hline
				Data utility &  \textbf{0.8149} & 0.8129 & 0.8141 & 0.8020 & 0.8040 & 0.8114 & 0.7986 & 0.7996 & 0.7681   \\
				EO level    &0.1976 & 0.2139 & 0.2042 & 0.2106 & 0.2458 & 0.1444 &\textbf{0.0389} & 	      & \\
				OMR level  &	0.0715 & 0.0684 & 0.0703 & 0.0925 & 0.0815 & 0.0740 & 	& \textbf{0.0440}& \\
				SP level & 0.4033 & 0.4098 & 0.4065 & 0.4548 & 0.4615 & 0.3724 &  &  &\textbf{0.0414} \\
				\hline
				\multicolumn{10}{l}{Test set}\\
				\hline
				Data utility & \textbf{0.8067} & 0.8050 & 0.8047 & 0.7866 & 0.7876 & 0.7930 & 0.7927 & 0.7880 & 0.7593 \\
				EO level      &0.1897 & 0.1897 & 0.1941 & 0.1956 & 0.2180 & 0.1210 & \textbf{0.0495} &  & \\
				OMR level    & 0.0828 & 0.0817 & 0.0822 & 0.1086 & 0.1073 & 0.0966 &  &\textbf{0.0599} & \\
				SP level      & 0.4259 & 0.4258 & 0.4278 & 0.4771 & 0.4872 & 0.3959 &  &  &\textbf{0.0783}\\ \bottomrule
			\end{tabular}
			\begin{tablenotes}
				\item Note: \textit{Statistical parity} is denoted by SP, \textit{equality of opportunity} by EO, and \textit{equal overall misclassification rate} by OMR. The optimal values are highted in bold.
			\end{tablenotes}		
		\end{threeparttable}
		
	\end{table}
	
	\begin{table} [!htbp]
		\captionsetup{font={footnotesize}}
		\caption{The average values of data utility and fairness level for all methods on \textit{German} data set with $\delta^{s}=0.01$.}
		\label{tab: sepciGerman}
		\footnotesize
		\centering
		\begin{threeparttable}
			\begin{tabular}{@{}lccccccccc@{}}
				\toprule
				& \multicolumn{6}{c}{Baselines}  & \multicolumn{3}{c}{Ours}   \\ \cmidrule(l){2-7} \cmidrule(l){8-10} 
				Metric &  Ridge  &  Lasso  &  Elasticnet  &  SVM  &  Huberized SVM  &  SLIM  &  FASS-EO  &  FASS-OMR  & FASS-SP\\
				\midrule
				\multicolumn{10}{l}{Training set}\\
				\hline
				Data utility & \textbf{0.8261}&0.8251&0.8257&0.8232&0.8178&0.8161&0.7694&0.8035&0.7969\\
				EO level      &0.0227 &0.0249 &0.0203 &0.0242 &0.0215 &0.0371 &\textbf{0.0035} && \\
				OMR level   &0.0202  &0.0210 &0.0193  &0.0208  &0.0221  &0.0171 &   & \textbf{0.0096}  &   \\
				SP level      &0.1494  &0.1294 &0.1280 &0.1269 &0.1272 &0.1115 &&&\textbf{0.0076}\\
				\hline
				\multicolumn{10}{l}{Test set}\\
				\hline
				Data utility &0.8187&0.8196&\textbf{0.8207}&0.8144&0.8114&0.8076&0.7522&0.8002&0.7993\\
				EO level      &0.0325 &0.0372 &0.0366 &0.0341 &0.0523 &0.0480 &\textbf{0.0188} &&   \\
				OMR level     & 0.0293 & 0.0346 & 0.0347  & 0.0321  & 0.0297 &0.0427  &  & \textbf{0.0282}  &   \\
				SP level      &0.1624  & 0.1369 & 0.1384& 0.1455 & 0.1488  &0.1115   &    &    & \textbf{0.0334} \\ \bottomrule
			\end{tabular}
			\begin{tablenotes}
				\item Note: \textit{Statistical parity} is denoted by SP, \textit{equality of opportunity} by EO, and \textit{equal overall misclassification rate} by OMR. The optimal values are highted in bold.
			\end{tablenotes}		
		\end{threeparttable}
	\end{table}

		In the \textit{Adult} data set, we first experiment with baseline classifiers. As shown in Table \ref{tab: sepciAdult}, all baselines lead to similar performance on overall data utility (around $0.8$) on test set. 
		However, these classifiers result in highly disparate impact and disparate mistreatment for male and female groups. Specifically, the disparate impact is exceptionally high since the minimum SP level between the two groups, achieved by SLIM, is $0.3959$ on test set. Moreover, the disparate mistreatment based on EO is also significant since the minimum EO fairness level, again by SLIM, is $0.1210$ on test set. In comparison, the disparate mistreatment based on OMR is relatively milder, where the best level is $0.0817$ using Lasso. To harness these disparities, we set $\delta^{s}=0.05$ and develop FASS scoring systems with different fairness metrics. Results show that our models only sacrifice a small degree of data utility to remedy the disparities. As shown in Table \ref{tab: sepciAdult}, our systems strictly limit the unfairness level to less than $0.05$ on the training set and significantly reduce the disparity levels on the test set. For example,  FASS-EO decreases disparate mistreatment based on EO to $0.0495$ on test set while maintaining high accuracy close to the baselines (it even outperforms two SVMs on accuracy, besides decreasing unfairness level).

		For the \textit{German} data, the phenomenon of disparity is less severe compared to the \textit{Adult} data, thus $\delta^{s}$ is set as $0.01$ in this scenario. As shown in Table \ref{tab: sepciGerman}, the disparate impact, again, is the most significant since the minimum SP rate difference of baselines is $0.1115$ by SLIM on test sets. However, our approach can limit this value to $0.0076 < 0.01$ on the training set and reduce it to $0.0334$ on the test set while guaranteeing a competitive accuracy. In contrast to the disparate impact, the disparate mistreatment issue is less of a concern on \textit{German} data, and our framework still outperforms the baselines on fairness levels for both EO and OMR metrics.

\begin{figure} [htbp]
		\centering
		\subfigure[Training Set]{ 
			\includegraphics[width=470pt]{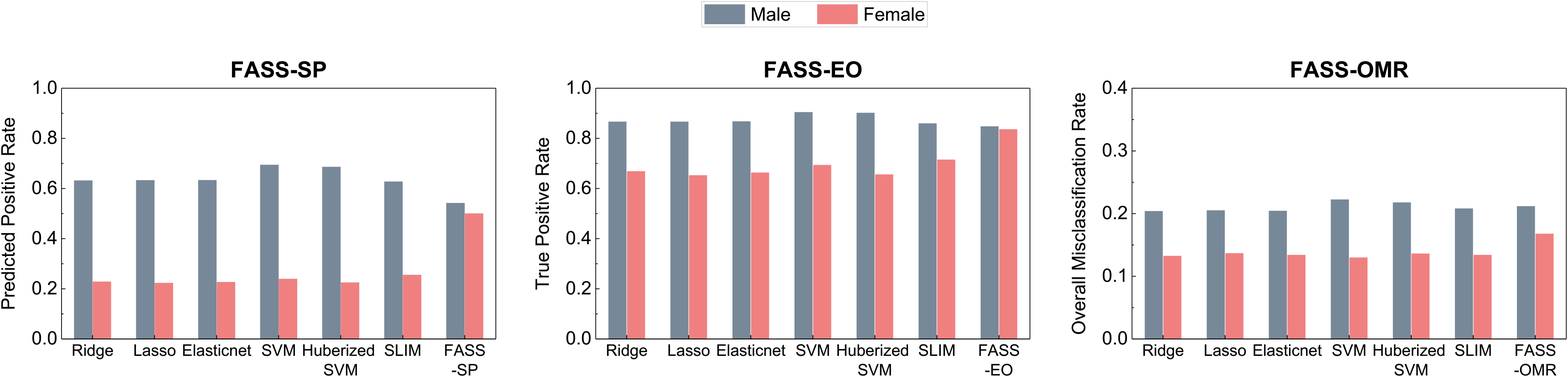}}		
		\subfigure[Test Set]{ 
			\includegraphics[width=470pt]{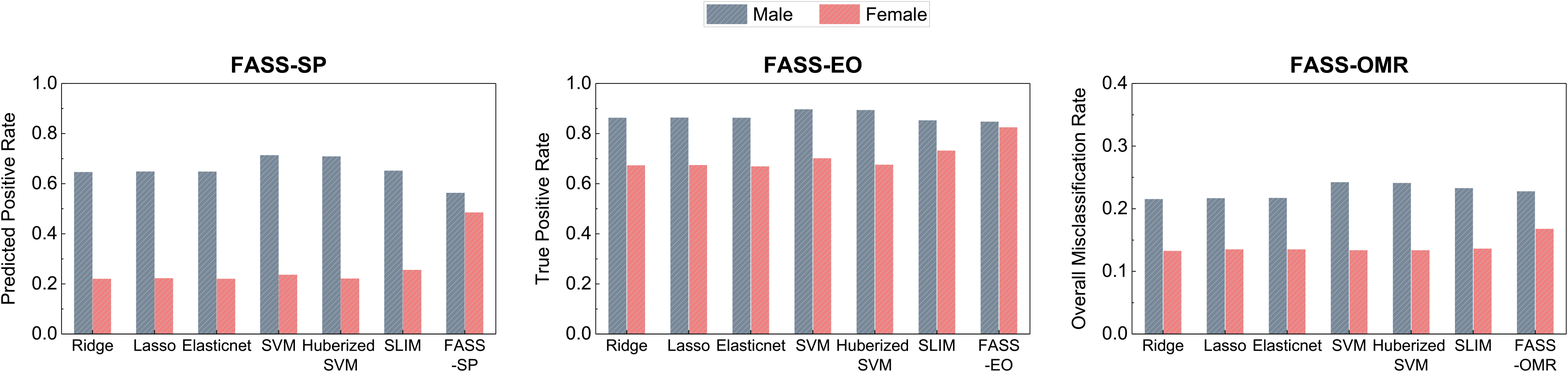}}
		\caption{Illustration of disparities for male and female subgroups on \textit{Adult} data set. }
		\label{fig:Zhuadult}
	\end{figure}

\begin{figure} [htbp]
		\centering
		\subfigure[Training Set]{ 
			\includegraphics[width=470pt]{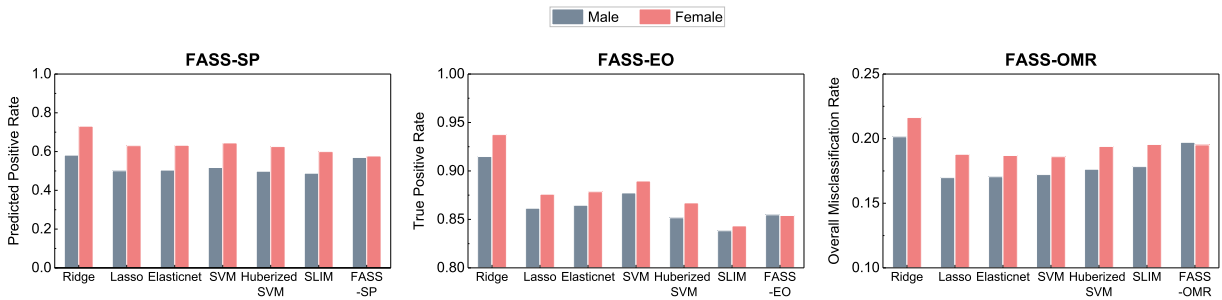}}		
		\subfigure[Test Set]{ 
			\includegraphics[width=470pt]{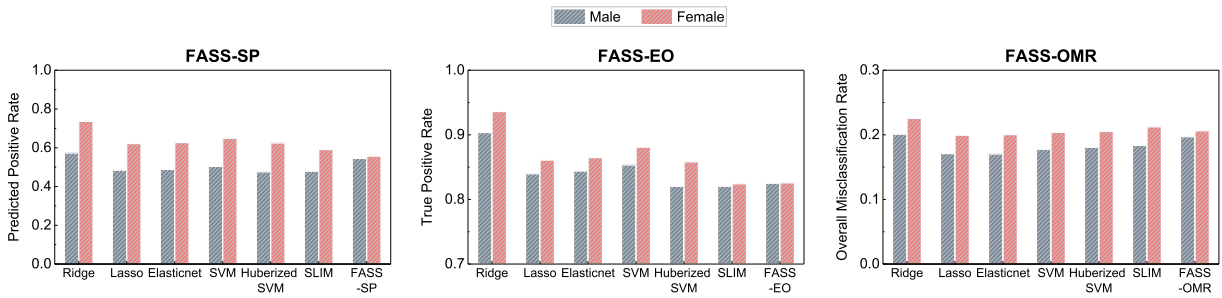}}
		\caption{Illustration of disparities for male and female subgroups on \textit{German} data set. }
		\label{fig:Zhugerman}
	\end{figure}

		The details regarding disparities between the two groups on \textit{Adult} and \textit{German} data sets are presented in Figures \ref{fig:Zhuadult} and \ref{fig:Zhugerman}, respectively. As can be seen from these figures, the rate gap between the two groups is substantially narrowed in our framework compared to baselines, and thus our scoring system achieves a better fairness level with all fairness measures.
		
		Overall, the experimental results on UCI data sets demonstrate the effectiveness of the proposed methods when the maximal fairness level is specified in advance. In this case, our approach can achieve a better fairness level for all fairness metrics compared to baselines, with only a moderate loss in data utility (accuracy).

	\clearpage

	%
	%
	%

\end{document}